%% file: acl_latex.tex
\pdfoutput=1

\documentclass[11pt]{article}

\usepackage[final]{acl}

\usepackage{times}
\usepackage{latexsym}

\usepackage[T1]{fontenc}

\usepackage[utf8]{inputenc}

\usepackage{microtype}

\usepackage{inconsolata}

\input{preamble}
\title{MMMU-Pro: A More Robust  Multi-discipline Multimodal \\Understanding Benchmark}

\author{
\textmd{Xiang Yue\thanks{Equal contributions. Contact: xyue2@andrew.cmu.edu}~,  Tianyu Zheng\footnotemark[1], \, Yuansheng Ni\footnotemark[1], \;} \\ Yubo Wang, Kai Zhang, Shengbang Tong, Yuxuan Sun, Botao Yu, Ge Zhang,\\ Huan Sun, Yu Su, Wenhu Chen, Graham Neubig 
\AND \large MMMU Team 
\\[0.5em] \url{https://mmmu-benchmark.github.io/\#leaderboard}
}
\begin{document}
\maketitle
\input{sec/0_abstract}


\input{sec/1_introduction}

\input{sec/2_main}

\input{sec/3_experiment}
\input{sec/4_guide}
\input{appendix/X_related_work}
\input{sec/5_conclusion}

\input{bbl}
\input{sec/X_appendix}

\end{document}

%% file: preamble.tex
\usepackage{wrapfig}
\usepackage{xcolor}         
\usepackage{multirow}
\usepackage{graphicx}
\usepackage{subcaption}
\usepackage{enumitem}
\usepackage{tcolorbox}
\usepackage{color-edits}
\usepackage{tocloft}
\usepackage{booktabs}
\usepackage{adjustbox}
\usepackage{amsmath}
\newlistof{appfigures}{afg}{List of Comparision Figures in Different Settings}

\extrafloats{100}

\usepackage[normalem]{ulem}
\useunder{\uline}{\ul}{}

\addauthor{gn}{magenta}
\addauthor{ysu}{cyan}
\definecolor{darkblue}{RGB}{84, 112, 198}
\definecolor{lightgreen}{RGB}{145, 204, 117}
\definecolor{lightyellow}{RGB}{250, 200, 88}
\definecolor{lightred}{RGB}{238, 102, 102}
\definecolor{lightblue}{RGB}{115, 192, 222}

\newtcolorbox{promptbox}[2][Prompt]{
colback=black!5!white,
arc=5pt, 
boxrule=0.5pt,
fonttitle=\bfseries,
title=#1, 
before upper={\small}, fontupper=\fontfamily{ptm}\selectfont,
colframe=#2,
}

\usepackage{etoc}
\usepackage{titletoc}

\newcommand\DoToC{%
  \startcontents
  \printcontents{}{1}{\noindent \textbf{\Large{Table of Contents in Appendix}}\vskip3pt\vskip5pt}
  \vskip3pt\vskip5pt
}

%% file: sec/0_abstract.tex
\begin{abstract}

This paper introduces MMMU-Pro, a robust version of the Massive Multi-discipline Multimodal Understanding and Reasoning (MMMU) benchmark. 
MMMU-Pro rigorously assesses multimodal models' true understanding and reasoning capabilities through a three-step process based on MMMU: 
(1) filtering out questions answerable by text-only models, 
(2) augmenting candidate options, and 
(3) introducing a vision-only input setting where questions are embedded within images. 
This setting challenges AI to truly ``see" and ``read" simultaneously, testing \textit{a core human cognitive skill of seamlessly integrating visual and textual information}. Results show that model performance is substantially lower on MMMU-Pro than on MMMU, ranging from 16.8\% to 26.9\% across models. 
We explore the impact of OCR prompts and Chain of Thought (CoT) reasoning, finding that OCR prompts have minimal effect while CoT generally improves performance. 
MMMU-Pro provides a more rigorous evaluation tool, closely mimicking real-world scenarios and offering valuable directions for future multimodal research.

\end{abstract}

%% file: sec/1_introduction.tex
\section{Introduction}
Recent advances in multimodal large language models (MLLMs) have led to progress in tackling complex reasoning tasks that combine textual and visual information~\citep{yin2023survey, jin2024efficient}. Models like GPT-4o~\citep{gpt4o} have achieved impressive results, e.g., on the Massive Multi-discipline Multimodal Understanding and Reasoning (MMMU) benchmark~\citep{yue2024mmmu}, reaching an accuracy of 69.1\% on college-level questions that integrate text and images. 

While these achievements are significant, they raise a critical question: \textit{Do the current benchmark results truly reflect a deep, multifaceted understanding of diverse subjects, or are these models exploiting subtle shortcuts and statistical patterns to arrive at correct answers without genuine comprehension and reasoning?}

This question has profound implications for the development and deployment of AI systems in real-world applications. If models rely on superficial cues rather than true multimodal understanding~\citep{du2023shortcut, yuksekgonul2023and}, we risk overestimating their capabilities and potentially deploying systems that fail in unpredictable ways when faced with novel scenarios~\citep{wu2024v}. 

To address this concern and push the boundaries of multimodal AI evaluation, we introduce MMMU-Pro, a more robust and challenging version of the MMMU benchmark. MMMU-Pro is designed to more accurately and rigorously assess a model's true multimodal understanding and reasoning capabilities across a wide range of academic disciplines. 
The development of MMMU-Pro is motivated by key observations, including the text-only solvability of some benchmark questions, limited option space in multiple-choice formats~\citep{wang2024mmlu}, and the need to challenge models' ability to jointly understand different modalities in a more integrated way.

\begin{figure*}[!ht]
    \centering
    \includegraphics[width=0.9\linewidth]{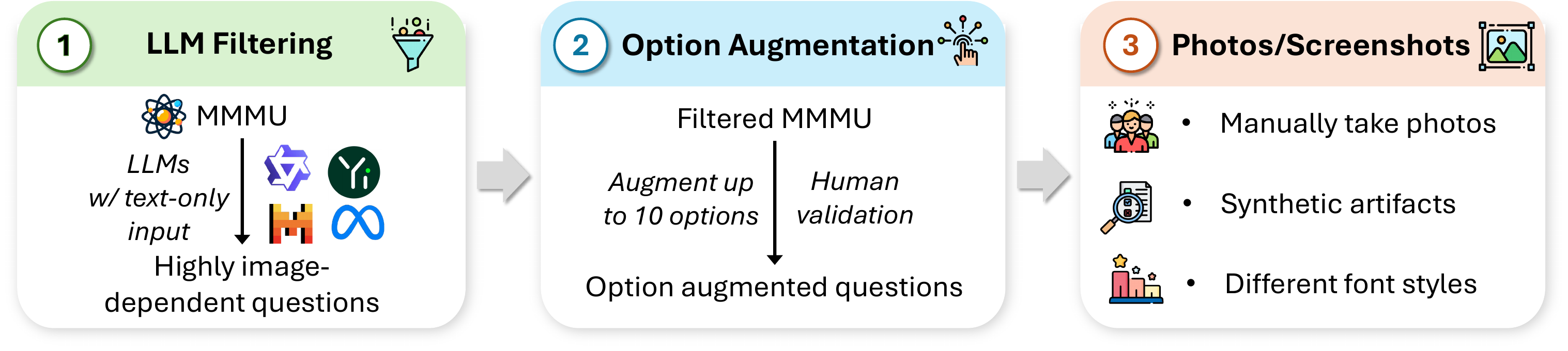}
    \caption{An overview of the construction process of MMMU-Pro.
}
\vspace{-15pt}
    \label{fig:method}
\end{figure*}

MMMU-Pro employs a rigorous three-step construction process (as shown in \autoref{fig:method}) that builds upon MMMU~\citep{yue2024mmmu}: (1) filtering out questions answerable by text-only language models, (2) augmenting candidate options to reduce the effectiveness of guessing based on the options, and (3) introducing a vision-only input setting (as shown in \autoref{fig:mmmu_pro_examples}) where models are presented with questions embedded in a screenshot or photo. 

The introduction of the vision-only input setting is particularly crucial, as it tests a fundamental human cognitive ability: \textit{the seamless integration and switching between visual and textual information}. This setting challenges models to develop the capability to truly ``see'' and ``read'' simultaneously, mirroring how humans effortlessly process complex scenes where text and images are intertwined.  This ability is crucial for tasks ranging from interpreting scientific diagrams~\citep{li2024mmsci} to navigating graphical user interfaces~\citep{liu2024visualwebbench,zheng2024gpt,koh2024visualwebarena}. Moreover, this approach aligns with how users naturally interact with AI systems, often sharing screenshots or photos rather than separating text and images. 

\begin{figure*}[!ht]
    \centering
    \includegraphics[width=\linewidth]{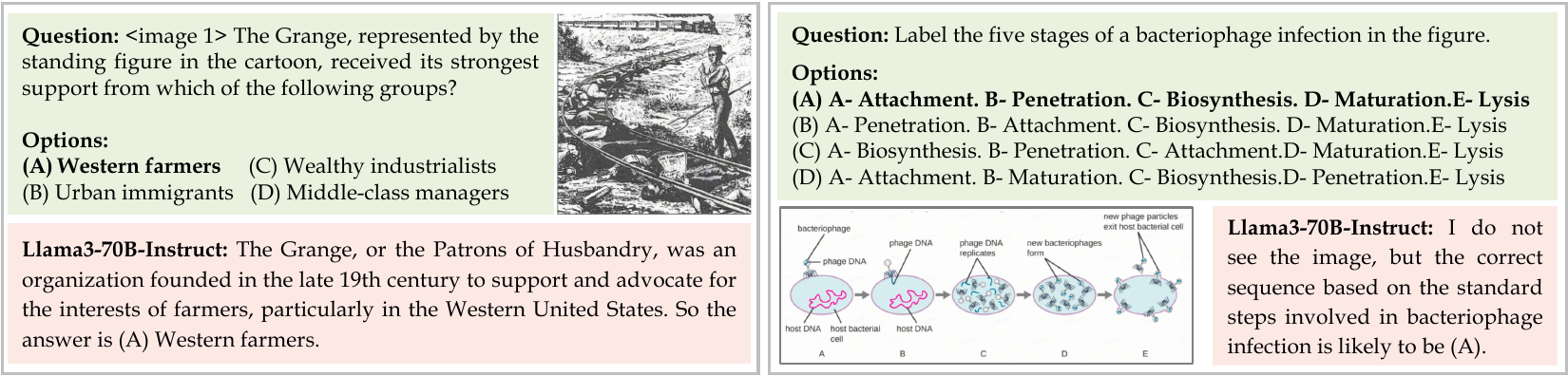}
    \vspace{-15pt}
    \caption{Two MMMU questions that are answered correctly by a text-only LLM Llama-3-70B Instruct. The model finds shortcuts or correlations in the text question and the candidate options. }
    \vspace{-15pt} 
    \label{fig:text_answerable_example}
\end{figure*}

Our experimental results demonstrate the effectiveness of MMMU-Pro in providing a more rigorous evaluation of multimodal models. We observe significant performance drops across all tested models when compared to the original MMMU benchmark, with decreases ranging from 16.8\% to 26.9\%. These results highlight the limitations of current state-of-the-art models in true multimodal understanding and reasoning. Furthermore, our analysis reveals that while CoT~\citep{wei2022chain} prompting generally improves performance, the benefits vary across models and settings.

Interestingly, we find that explicit OCR prompts do not significantly impact performance for most models, suggesting that advanced multimodal models have already developed robust text extraction capabilities from images. However, this result also underscores that simple OCR is insufficient for the challenges presented by MMMU-Pro's vision-only input setting. Our further qualitative analysis indicates that when text is embedded within images, it significantly increases the overall complexity of the visual input, requiring models to not only recognize text but also understand its context, relationship to visual elements, and relevance to the question. These findings not only provide a more accurate assessment of current multimodal AI capabilities but also highlight the need for more sophisticated multimodal reasoning abilities.

%% file: sec/2_main.tex
\section{MMMU-Pro: A More Robust Version of MMMU}
\subsection{Revisiting the MMMU Benchmark
}

The Massive Multi-discipline Multimodal Understanding and Reasoning (MMMU) benchmark~\citep{yue2024mmmu} is a comprehensive dataset designed to evaluate multimodal AI models on college-level tasks that require subject-specific knowledge and deliberate reasoning. MMMU consists of 11.5K carefully curated multimodal questions from college exams, quizzes, and textbooks, covering six core disciplines across 30 subjects and 183 subfields. Each question in MMMU is a multimodal image-text pair with 4 multiple-choice options, featuring 30 diverse image types such as charts, diagrams, maps, and chemical structures. 
MMMU has rapidly established itself as a standard evaluation framework for testing prominent multimodal models upon their release.~\citep{gpt4o,gpt4o-mini,claude-3.5-sonnet,reid2024gemini,li2024llava}.

However, we find that text-only LLMs can accurately answer some questions without requiring any visual input. We take a closer look at these questions and identify two main issues: 1) \textbf{Text-Only Dependency:} Certain questions are relatively independent or irrelevant to the corresponding images. 2) \textbf{Shortcut Exploitation:} Even when questions require images for humans to answer correctly, models often find shortcuts or correlations within the candidate options, leveraging their pre-existing knowledge (from pre-training) to arrive at the correct answer. Two examples that are answered correctly by Llama-3-70B Instruct~\citep{dubey2024llama} are shown in \autoref{fig:text_answerable_example}.

\subsection{Methods}

To address these issues and build a more robust benchmark, we implemented a three-step approach.

\noindent\textbf{Filtering Questions:} We begin by filtering out questions that can be answered by text-only LLMs. We select four strong open-source LLMs: Llama3-70B-Instruct~\citep{dubey2024llama}, Qwen2-72B-Instruct~\citep{yang2024qwen2}, Yi-1.5-34B-Chat~\citep{young2024yi}, and Mixtral-8$\times$22B-Instruct~\citep{mixtral-8x22}—and task them with answering the MMMU questions without access to images. The models are required to provide answers even when they indicate that visual input is necessary. We repeat this process ten times for each model, considering a question as ``answerable'' if a model correctly answers it more than five times. We then exclude any question where at least three out of the four models answer correctly across the majority of trials. We randomly sample 1800 questions from the remaining pool, evenly distributed across 30 subjects (60 questions per subject).

\noindent\textbf{Augmenting Candidate Options:} Despite the filtering, some questions can still be answered by text-only LLMs, often exploiting subtle hints within the candidate options. To counteract this, we increase the number of candidate options from four to ten, making it more challenging for models to rely on guessing. This augmentation is done by human experts with the help of GPT-4o, with additional validation steps to ensure the quality and diversity of the options. Specifically, GPT-4o generates and Claude 3.5 filters the options, followed by two rounds of human review to refine and verify the augmented options. This augmentation is done by human experts with the help of GPT-4o. During this process, experts also review the original annotated questions to ensure their relevance to the images and to eliminate any questions that lack a clear connection or coherence. This step filters out 70 questions, and we obtain 1730 questions in total.

\begin{figure}[ht]
    \centering
    \includegraphics[width=\linewidth]{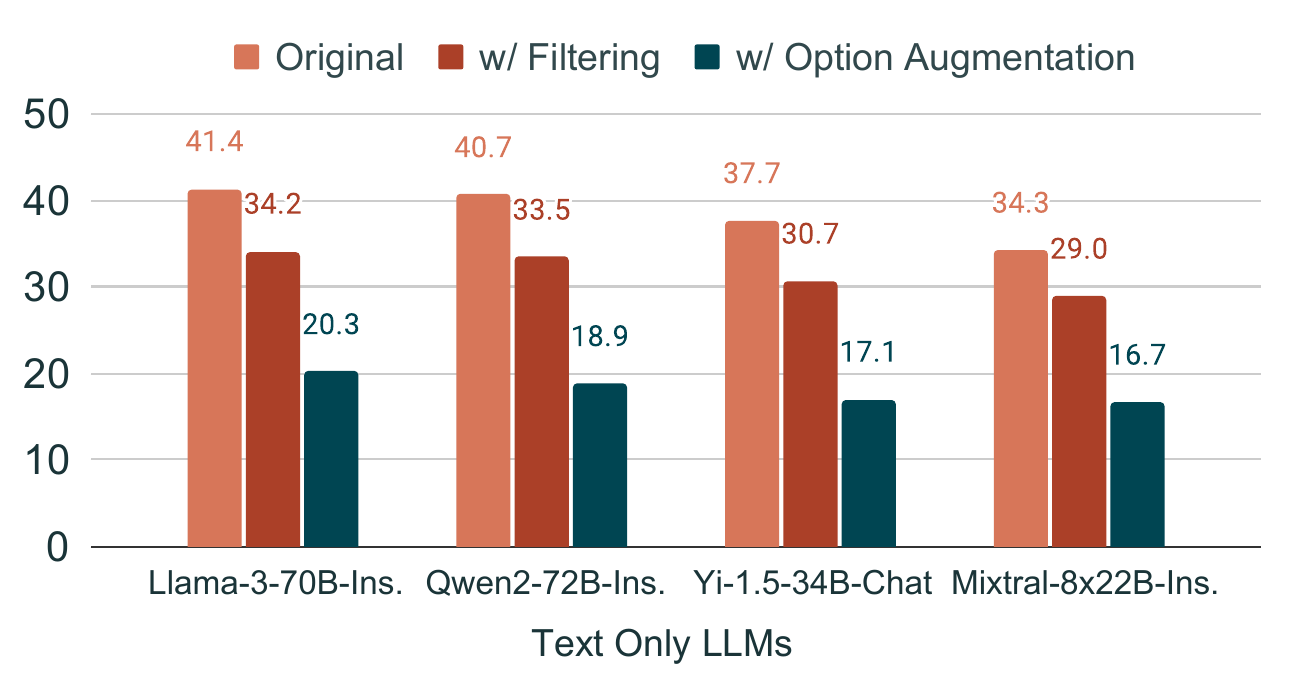}
    \caption{Accuracy of text-only LLMs in different sets of MMMU questions.}
    \label{fig:text_only}
\end{figure}
As illustrated in \autoref{fig:text_only}, these two steps significantly reduce the accuracy of text-only models attempting to guess the answers.

\begin{figure*}[!ht]
    \centering
    \includegraphics[width=\linewidth]{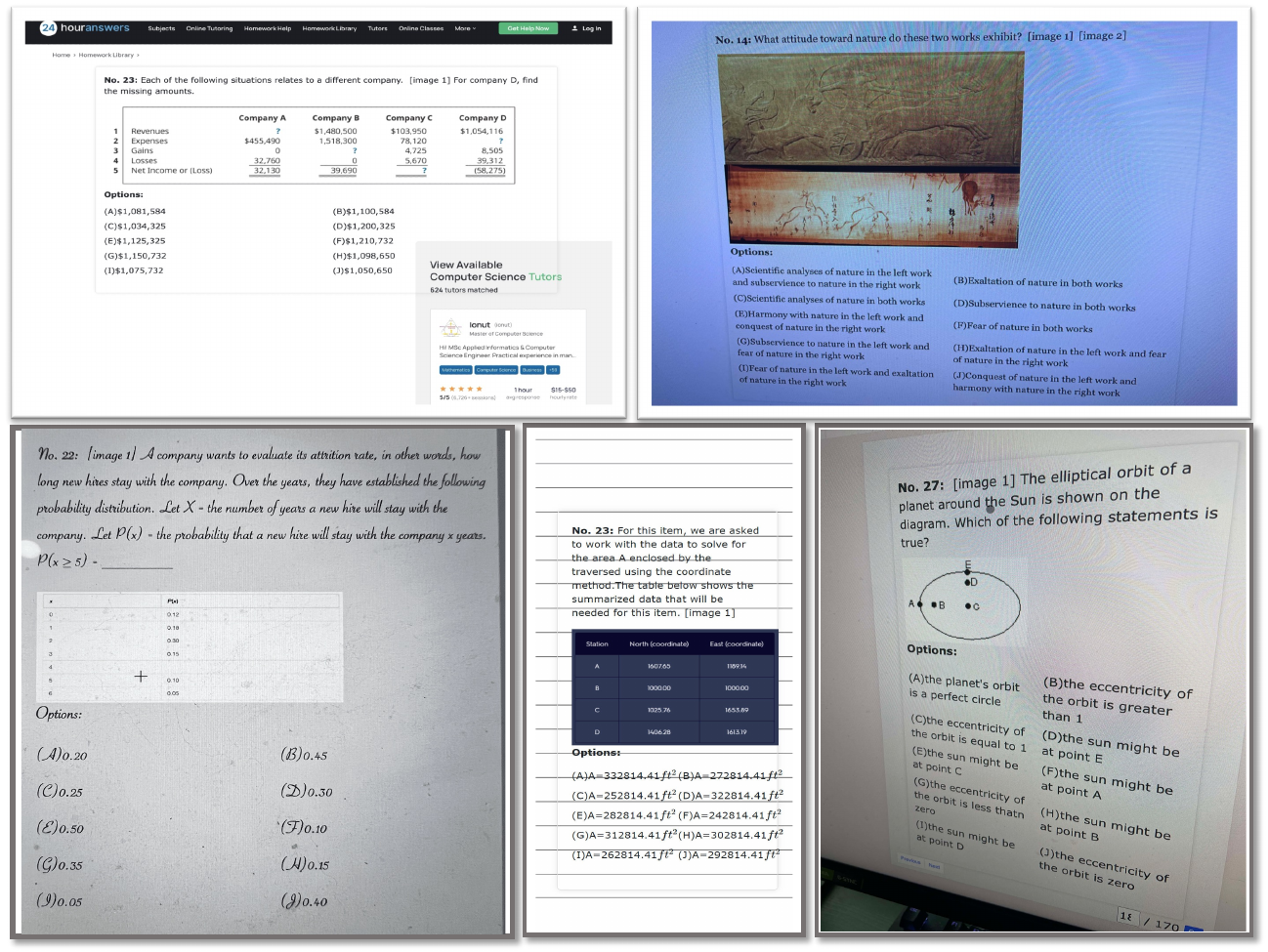}
    \caption{Sample questions from MMMU-Pro Vision. The model is required to answer a multiple-choice question with up to 10 options, each embedded within a screenshot or photo. The images were manually captured by annotators in diverse display environments to reflect real-world cases.}
    \vspace{-11pt}    
    \label{fig:mmmu_pro_examples}
\end{figure*}

\noindent\textbf{Enhancing Evaluation with a Vision-Only Setting:} To further challenge the multimodal understanding of models, we introduce a vision-only input setting in MMMU-Pro. In this setting, the model is presented with a question embedded within a screenshot or photo, without any text explicitly fed into the model. To implement this setting, we ask the human annotators to manually capture photos and screenshots over a simulated display environment. This process involves varying the backgrounds, font styles, and font sizes to replicate the diversity of real-world conditions. By using different combinations of these elements, we create a broad range of visual contexts, ensuring that the models are not only challenged by the integration of text and images but also by the variability in how this content is presented. Examples of the vision-only input setting are shown in \autoref{fig:mmmu_pro_examples}. 

The motivation for this setting comes from real-world usage and human cognition. Users often capture screenshots of questions with both text and images instead of inputting text separately, reflecting a natural tendency to process information holistically. Humans excel at understanding integrated visual-textual content, and this setting encourages models to develop similar comprehension. By mimicking this behavior, the vision-only input setting enhances realism and prepares models for real-world multimodal tasks. Ultimately, we obtain 3,460 questions—1,730 in standard format and 1,730 as screenshots or photos.

%% file: sec/3_experiment.tex
\section{Experiments}
\subsection{Experimental Setups}
\noindent\textbf{Baselines.} To establish a comprehensive understanding of MMMU-Pro's difficulty and to provide reference points for future research, we evaluate a diverse set of state-of-the-art multimodal models as baselines. These models represent a range of training approaches and capabilities in the field of multimodal AI. Our baseline models include:

\textit{Proprietary Models:} GPT-4o (0513)~\citep{gpt4o} and GPT-4o mini~\citep{gpt4o-mini}, Claude 3.5 Sonnet~\citep{claude-3.5-sonnet}, and Gemini 1.5 Pro (0801 and 0523 versions)~\citep{deepmind_gemini_report,reid2024gemini}. These models represent the cutting edge of multimodal AI capabilities.

\textit{Open-source models:} We evaluate a range of open-source models, including InternVL2 (8B, 40B, and Llama3-76B versions)~\citep{chen2024far}, LLaVA (OneVision-7B, OneVision-72B, and various NeXT versions)~\citep{li2024llava,liu2024llava}, VILA-1.5-40B~\citep{lin2024vila}, MiniCPM-V2.6~\citep{yao2024minicpm}, Phi-3.5-Vision~\citep{abdin2024phi}, and Idefics3-8B-Llama3~\citep{laurenccon2024building}. These models showcase the current state of publicly available multimodal AI systems.
We evaluate these models across three different settings: 1) Standard setting without augmented options (usually 4 options); 2) Standard setting with augmented options (usually 10 options); 3)Vision-only input setting.

The overall performance score for MMMU-Pro is calculated as the average of scores from settings (2) and (3). We include setting (1) and report the original MMMU validation set performance solely for comparison purposes, to highlight the increased difficulty of MMMU-Pro.

We evaluate the models with both \textit{Direct} and \textit{CoT} prompts (as shown in \autoref{ocr_prompt}), and report the higher ones in the overall results. We also discuss the influence of the CoT prompt in \ref{sec:cot_prompt}.

\begin{table*}[!t]
\centering
\small
\resizebox{\linewidth}{!}{%
\begin{tabular}{@{}lccccll@{}}
\toprule
 &
  \multicolumn{3}{c}{\textbf{MMMU-Pro}} &
  \multirow{4}{*}{\begin{tabular}[c]{@{}c@{}}\textbf{MMMU} \\ \textbf{(Val)}\end{tabular}} &
  \multirow{4}{*}{\textbf{$\Delta_1$}} &
  \multirow{4}{*}{\textbf{$\Delta_2$}} \\ \cmidrule(r){2-4}
 &
  \begin{tabular}[c]{@{}c@{}}\textbf{Standard}\\ \textbf{(4 Opts)}\end{tabular} &
  \begin{tabular}[c]{@{}c@{}}\textbf{Standard}\\ \textbf{(10 Opts)}\end{tabular} &
  \textbf{Vision} &
   &
   &
   \\ \midrule
{\color[HTML]{9B9B9B} Random Choice} & \color[HTML]{9B9B9B}24.9 &\color[HTML]{9B9B9B} 12.8 &\color[HTML]{9B9B9B} 12.4 &\color[HTML]{9B9B9B} 22.1 &\color[HTML]{9B9B9B} -9.3 &\color[HTML]{9B9B9B} -9.7 \\
{\color[HTML]{9B9B9B} Frequent Choice} &\color[HTML]{9B9B9B} 27.8 &\color[HTML]{9B9B9B} 12.1 &\color[HTML]{9B9B9B} 12.1  &\color[HTML]{9B9B9B} 26.8 &\color[HTML]{9B9B9B} -14.7 &\color[HTML]{9B9B9B} -14.7 \\
Human Expert (Low) &75.4  &73.0  &73.0  &76.2  &-3.2  &-3.2  \\
Human Expert (Medium) &82.1  &80.8  &80.8  & 82.6 &-1.8  &-1.8 \\
Human Expert (High) &88.6  &85.4  &85.4  &88.6  &-3.2  &-3.2 \\
    \midrule
GPT-4o (0513) \citep{gpt4o}& \textbf{64.7} & {\ul 54.0} & \textbf{49.7} & \textbf{69.1} & {\ul -15.1} {\color[HTML]{009901}($\uparrow$ 1)} & \textbf{-19.4} ( - ) \\
Claude 3.5 Sonnet \citep{claude-3.5-sonnet}& {\ul 63.7} & \textbf{55.0} & {\ul 48.0} & {\ul 68.3} & \textbf{-13.3} {\color[HTML]{FE0000}($\downarrow$ 1)}& {\ul -20.3} ( - )\\
Gemini 1.5 Pro (0801) \citep{reid2024gemini}& 60.6 & 49.4 & 44.4 & 65.8 & -16.4 ( - )& -21.4 ( - )\\
Gemini 1.5 Pro (0523) \citep{reid2024gemini}& 57.6 & 46.5 & 40.5 & 62.2 & -15.7 ( - )& -21.7 ( - )\\
GPT-4o mini \citep{gpt4o-mini}& 55.3 & 39.9 & 35.2 & 59.4 & -19.5 {\color[HTML]{009901}($\uparrow$ 1)}& -24.2 {\color[HTML]{009901}($\uparrow$ 1)}\\ 
\midrule
Qwen2-VL-72B \citep{Qwen2-VL}& \textbf{59.3} & \textbf{49.2} & \textbf{43.3} & \textbf{64.5} & \textbf{-15.3} ( - )& -21.2 ( - )\\
InternVL2-Llama3-76B \citep{chen2024far}& {\ul 55.0} & {\ul 41.9} & {\ul 38.0} & 58.3 & -16.4 {\color[HTML]{FE0000}($\downarrow$ 1)}& \textbf{-20.3} {\color[HTML]{FE0000}($\downarrow$ 1)}\\
InternVL2-40B \citep{chen2024far}& 47.4 & 36.3 & 32.1 & 55.2 & -18.9 ( - )& -23.1 {\color[HTML]{FE0000}($\downarrow$ 1)}\\
LLaVA-OneVision-72B \citep{li2024llava}& 52.3 & 38.0 & 24.0 & 56.8 & -18.8 ( - )& -32.8 {\color[HTML]{009901}($\uparrow$ 5)}\\
Qwen2-VL-7B \citep{Qwen2-VL}& 46.6 & 34.1 & 27.0 & 54.1 & -20.0 {\color[HTML]{009901}($\uparrow$ 1)}
& -27.1 {\color[HTML]{FE0000}($\downarrow$ 1)}\\
Pixtral-12B \citep{pixtral-12b}& 47.5 & 33.4 & 25.0 & 52.5 & -19.1 {\color[HTML]{009901}($\uparrow$ 1)}& -27.5 ( - )\\
InternVL2-8B \citep{chen2024far}& 42.6 & 32.5 & 25.4 & 51.2 & -18.7 ( - )& -25.8 {\color[HTML]{FE0000}($\downarrow$ 3)}\\
MiniCPM-V2.6 \citep{yao2024minicpm}& 40.6 & 30.2 & 24.2 & 49.8 & -19.6 {\color[HTML]{009901}($\uparrow$ 1)}& -25.6 {\color[HTML]{FE0000}($\downarrow$ 3)}\\
VILA-1.5-40B \citep{lin2024vila}& 46.8 & 35.9 & 14.1 & 51.9 & -16.0 {\color[HTML]{FE0000}($\downarrow$ 2)}& -37.8 {\color[HTML]{009901}($\uparrow$ 9)}\\
LLaVA-NEXT-72B \citep{liu2024llava}& 43.0 & 31.0 & 19.2 & 49.9 & -18.9 ( - )& -30.7 ( - )\\
LLaVA-OneVision-7B \citep{li2024llava}& 42.8 & 29.5 & 18.7 & 48.8 & -19.3 {\color[HTML]{009901}($\uparrow$ 2)}& -30.1 {\color[HTML]{FE0000}($\downarrow$ 1)}\\
LLaVA-NeXT-34B \citep{liu2024llava}& 44.5 & 30.3 & 17.2 & 48.1 & -17.8 {\color[HTML]{FE0000}($\downarrow$ 2)}& -30.9 {\color[HTML]{FE0000}($\downarrow$ 1)}\\
Idefics3-8B-Llama3 \citep{laurenccon2024building}& 40.8 & 30.1 & 15.6 & 46.6 & -16.5 {\color[HTML]{FE0000}($\downarrow$ 1)}& -31.0 ( - )\\
Qwen2-VL-2B \citep{Qwen2-VL}& 34.8 & 25.3 & 17.2 & 41.1 & {\ul -15.8} ( - )& -23.9 {\color[HTML]{FE0000}($\downarrow$ 3)}\\
Phi-3.5-Vision \citep{abdin2024phi}& 37.8 & 26.3 & 13.1 & 43.0 & -16.7 ( - )& -29.9 {\color[HTML]{009901}($\uparrow$ 3)}\\
LLaVA-NeXT-7B \citep{liu2024llava}& 33.7 & 19.4 & 14.6 & 35.3 & -15.9 ( - )& {\ul -20.7} {\color[HTML]{FE0000}($\downarrow$ 3)}\\
LLaVA-NeXT-13B \citep{liu2024llava}& 33.9 & 19.8 & 14.5 & 36.2 & -16.4 ( - )& -21.7 {\color[HTML]{FE0000}($\downarrow$ 1)}\\ \bottomrule
\end{tabular}%
}
\caption{Results of models on MMMU-Pro and MMMU (Val). $\Delta_1$: Standard (10 options) - MMMU (Val); $\Delta_2$: Vision - MMMU (Val). {\color[HTML]{FE0000}($\downarrow$)} represents a decrease in ranking, while {\color[HTML]{009901}($\uparrow$)} indicates an increase. The best-performing model in each category is \textbf{in-bold}, and the second best is {\ul{underlined}.}}
\vspace{-15pt} 
\label{tab:overall_results}
\end{table*}


\noindent\textbf{Approximating Human Expert Performance.} While rigorous human evaluation of MMMU-Pro provides valuable insights, conducting such an assessment is both time-consuming and costly. Instead, we develop an approach to approximate human expert performance based on the original MMMU human evaluation data. This approximation is justified by several key factors. Firstly, the core content and difficulty of the questions remain unchanged in MMMU-Pro, supporting the validity of using the original human performance data as a close approximation. Secondly, in the original MMMU evaluation, human experts are required to write out their problem-solving processes, significantly reducing the likelihood of random guessing. For questions without detailed solving processes, we randomly select one option from the augmented candidates and recalculate the accuracy. Finally, human experts, with their innate ability to seamlessly integrate visual and textual information, are expected to perform similarly in the vision-only input setting as they do in the original format. Based on these considerations, we posit that human expert performance on MMMU-Pro closely aligns with the original MMMU results, allowing us to maintain a human performance benchmark without incurring the substantial costs of a new expert evaluation. More details of the human estimation performance can be found in \autoref{sec:apdx:human}.
\subsection{Overall Results}

We presented the overall results of MMMU-Pro of different models in \autoref{tab:overall_results}. 

\noindent\textbf{Effect of Increased Candidate Options:} The shift from 4 to 10 candidate options ($\Delta_1$) reveals a significant drop in performance for all models. GPT-4o (0513) experienced a decrease of 10.7\%, from 64.7\% to 54.0\%. This indicates that increasing the number of options effectively reduces the likelihood of models guessing the correct answer, forcing them to engage more deeply with the multimodal content.

\noindent\textbf{Impact of Vision-Only Setting:} The introduction of the vision-only input setting further challenges models, as evidenced by the additional drop in performance when comparing the vision-only results to the 10-option standard ($\Delta_2$). For instance, GPT-4o (0513) dropped another 4.3\% in accuracy when evaluated in the vision-only setting, and LLaVA-OneVision-72B saw a dramatic 14.0\% decrease. This suggests that the vision-only setting successfully tests the models' ability to integrate visual and textual information, highlighting their limitations when the text is not explicitly provided.
\begin{figure*}[!t]
    \centering
    \includegraphics[width=0.49\linewidth]{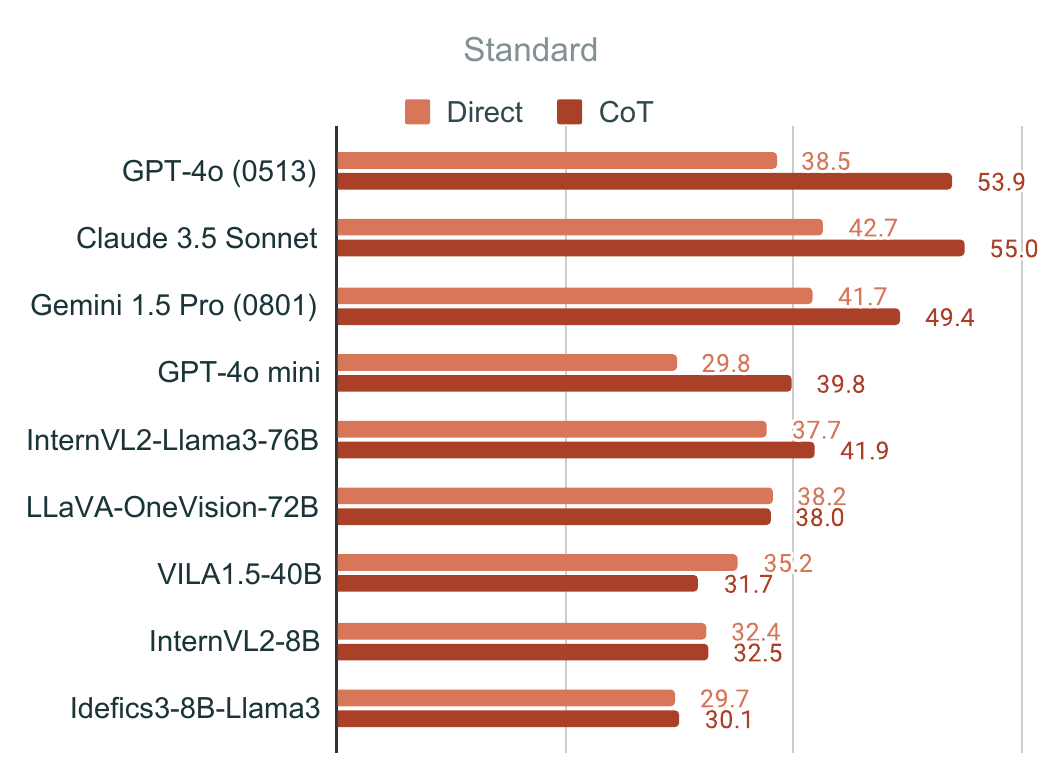}
    \includegraphics[width=0.49\linewidth]{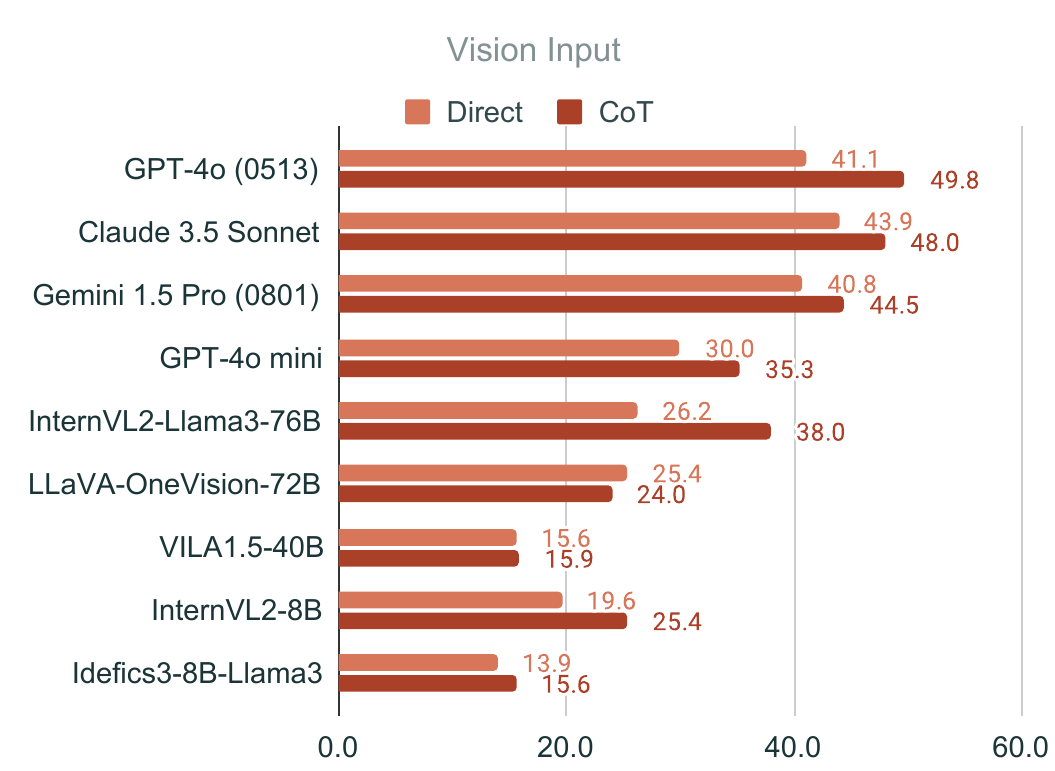}
    \caption{Impact of CoT prompting of different models in the two settings of MMMU-Pro. }
    \label{fig:cot_results}
\end{figure*}

\noindent\textbf{Combined Effects on MMMU-Pro:} The overall $\Delta_3$, representing the difference between MMMU-Pro and MMMU (Val), shows a significant decrease across the board. For instance, models like Gemini 1.5 Pro (0801) and Claude 3.5 Sonnet exhibited declines of 18.9\% and 16.8\%, respectively, while more drastic drops were seen in models like VILA-1.5-40B with a 26.9\% decrease.


This significant reduction in accuracy across the board suggests that MMMU-Pro successfully mitigates the shortcuts and guessing strategies that models could exploit in the original benchmark.

\subsection{Impact of CoT Prompting}
\label{sec:cot_prompt}
\autoref{fig:cot_results} examines the effectiveness of Chain of Thought (CoT) prompting on the MMMU-Pro benchmark, in both Standard and Vision Input settings.
Across both settings, CoT prompts generally improved performance, though the extent varied significantly. For instance, Claude 3.5 Sonnet saw a substantial increase in the Standard setting, rising from 42.7\% to 55.0\%, while models like LLaVA-OneVision-72B showed only minimal gains.

     

Interestingly, we observed a significant performance drop for some models, such as VILA1.5-40B. This decline might be attributed to challenges in instruction-following abilities. When a model struggles to follow instructions accurately, generating CoT explanations becomes more difficult. Additionally, these models may face issues with maintaining the correct response format, leading to what is known as ``boiled response format'' problems.
These findings highlight the potential of CoT to enhance model performance in complex, real-world tasks that require nuanced reasoning and integration of multiple information sources. However, they also underscore the importance of robust instruction-following capabilities as a prerequisite for effective CoT implementation.

The effectiveness of CoT prompting across disciplines is summarized in \autoref{tab:accuracy_comparison} and \autoref{fig:sub_cot}, comparing CoT and direct accuracy for GPT-4o and LLaVA-OneVision 72B. CoT shows significant improvements in reasoning-intensive fields like \textit{Tech and Engineering} (e.g., a 14.49\% gain for GPT-4o) and \textit{Science} (8.22\% gain). Smaller yet consistent gains are observed for LLaVA-OneVision 72B, such as 2.33\% in \textit{Tech and Engineering}. However, CoT's benefits are limited or negative in fields like \textit{Art and Design}, where GPT-4o gains only 1.58\%, and LLaVA-OneVision 72B sees a 17.12\% decline. These results underscore CoT's strengths in structured reasoning tasks but its reduced effectiveness in domains requiring subjective interpretation.


\subsection{Does OCR Help in the Vision Setting?}



In the Vision Input setting, one natural question is whether Optical Character Recognition (OCR) helps improve model performance on MMMU-Pro. We answer this question by first calculating the OCR accuracy of different models. Specifically, we ask the model to extract the full text of the question and answer choices. Then the OCR accuracy is calculated by comparing the text extracted with the original text using Levenshtein distance, which measures the difference between the two strings. The similarity between the extracted and original text is computed as:

\scalebox{0.8}{%
\(\displaystyle \text{OCR Accuracy} = 1 - \frac{\text{Levenshtein.distance}(\text{text1}, \text{text2})}{\max(\text{len(text1)}, \text{len(text2)})}\)%
}

\autoref{tab:model_results} shows that although most of the models demonstrate strong OCR capabilities, as indicated by high similarity scores. Based on the result, we then explore whether explicitly asking the model to first extract the question and then solve it (with an OCR prompt shown in \autoref{ocr_prompt}) could help in improving performance within the Vision Input setting of MMMU-Pro. Across the models evaluated, the inclusion of OCR prompts did not significantly alter performance. These minimal differences suggest that strong capable models are already proficient at extracting and understanding textual information from images, even without explicit OCR prompts.

\begin{table}[!h]
    \centering
    \small
    \resizebox{0.93\linewidth}{!}{%
    \begin{tabular}{@{}lccc@{}}
    \toprule
    \multirow{4}{*}{\begin{tabular}[c]{@{}c@{}}\textbf{Model}\end{tabular}}     & \multirow{4}{*}{\begin{tabular}[c]{@{}c@{}}\textbf{OCR}\\ \textbf{Acc.}\end{tabular}} &  \multicolumn{2}{c}{\textbf{Vision Setting Acc.}} \\\cmidrule(r){3-4} 
         &    & \begin{tabular}[c]{@{}c@{}}\textbf{w/ OCR}\\ \textbf{Prompt}\end{tabular} & \begin{tabular}[c]{@{}c@{}}\textbf{w/o OCR}\\ \textbf{Prompt}\end{tabular}  \\ \midrule
    GPT-4o                  & 92.3   & \textbf{49.7}         & 49.4 \\
    Gemini 1.5 Pro(0801)     & 89.7   & \textbf{44.4}         & 43.6 \\
    GPT-4o mini             & 89.6   & 35.2         & \textbf{35.6} \\
    InternVL2-Llama3-76B     & 88.1   & \textbf{38.0}         & 37.9 \\
    InternVL2-Llama3-40B     & 85.5   & \textbf{32.1}        & 28.9 \\
    Pixtral-12B              & 83.1   & \textbf{25.0}         & 24.1 \\
    LLaVA-OneVision-72B      & 87.8   & \textbf{24.0}         & 23.8 \\
    InternVL2-8B             & 85.2   & \textbf{25.4}         & 24.6 \\
    MiniCPM-V2.6             & 67.0   & \textbf{24.2}         & 21.1 \\
    LLaVA-NEXT-72B           & 62.0   & 19.2         & \textbf{20.0} \\
    Idefics3-8B-Llama3       & 68.5   & \textbf{15.6}         & 14.1 \\
    LLaVA-NeXT-7B            & 36.6   & \textbf{14.6}         & 14.3 \\
    LLaVA-NeXT-13B           & 51.1   & \textbf{14.5}        & 12.8 \\
    \bottomrule
    \end{tabular}%
    }
    \caption{Model performance in the Vision Input setting, comparing OCR accuracy with/without OCR prompts.} 
    \label{tab:model_results}
\end{table}

\begin{figure}[ht]
    \centering
    \includegraphics[width=0.9\columnwidth]{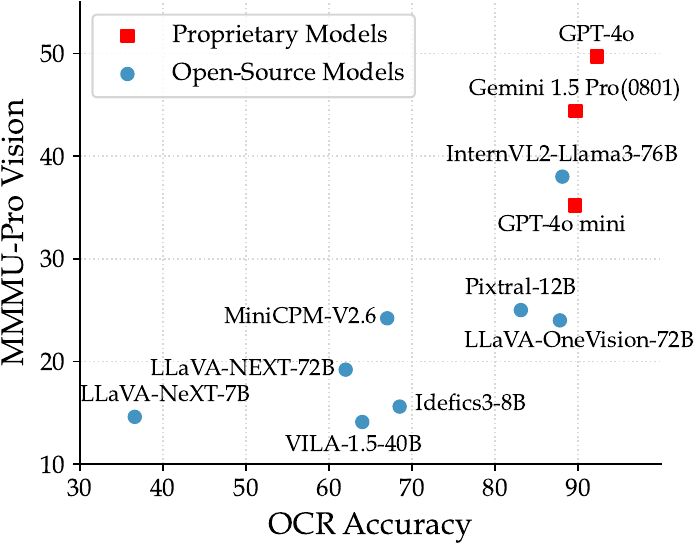}
    \caption{Correlation between OCR accuracy and MMMU-Pro Vision performance.}
    \label{fig:ocr_corr}
\end{figure}%


Interestingly, \autoref{fig:ocr_corr} shows that high OCR accuracy doesn’t always translate to strong multimodal reasoning. For example, LLaVA-OneVision-72B matches InternVL2-Llama3-76B and GPT-4o mini in OCR accuracy but lags significantly in MMMU-Pro Vision performance, indicating that OCR accuracy alone is insufficient for robust reasoning. Conversely, top-performing models like GPT-4o consistently excel in both areas. Despite GPT-4o’s high OCR accuracy, its MMMU-Pro Vision performance drops notably compared to MMMU (Val), revealing that even advanced models struggle to fully integrate and reason over multimodal inputs in the vision-only setting.

\subsection{Qualitative Analysis}

To gain deeper insights into model performance beyond quantitative metrics, we conducted a thorough qualitative analysis of MMMU-Pro results, focusing on two key scenarios: 1) Correct answers with four options but failure with ten options in the standard setting; 2) Success in the standard ten-option setting but failure in the vision input setting. Our analysis revealed several critical factors affecting model performance:

\noindent\textbf{Challenges with Increased Options.} Models often select the closest answer rather than arriving at a definitive choice, leading to increased errors with more options, as shown in \autoref{fig:case_study1}. Conceptually similar options, particularly in nuanced questions, can cause confusion. For instance, in conceptual questions, models struggled to differentiate subtle distinctions within a subject area, revealing limitations in fine-grained understanding.


\noindent\textbf{Increased Cognitive Load in Vision-Text Integration.} Processing visual and textual inputs simultaneously increases the cognitive load on models. An example is shown in \autoref{fig:case_study2}. The model perfectly extracted the text from the image but still failed to answer the question correctly. Another case is shown in \autoref{fig:manage}. The graph's similar lines and overlapping data points may distract the model from distinguishing between the two unemployment categories, leading to the error.

\noindent\textbf{Overemphasis on Visual Cues in Multimodal Reasoning.} When visual cues dominate over textual reasoning, models may incorrectly prioritize less relevant information from the images. In the \autoref{fig:history} example, the Vision Setting incorrectly chose the League of Nations by focusing on the World War I image, missing the broader context of World War II and the United Nations. A proper balance between visual and textual information is essential to avoid such mistakes.

\noindent\textbf{Impact of Context Switching.} Rapid transitions between visual and textual information can cause models to lose focus or misinterpret key data. For example, in \autoref{fig:math}, the model initially correctly defined both the objective function and the algebraic constraints. However, due to context switching between the textual description and the geometric figure, it misinterpreted the feasible region.


\subsection{Error Analysis}

\begin{figure}[ht]
    \centering
    \includegraphics[width=0.7\columnwidth]{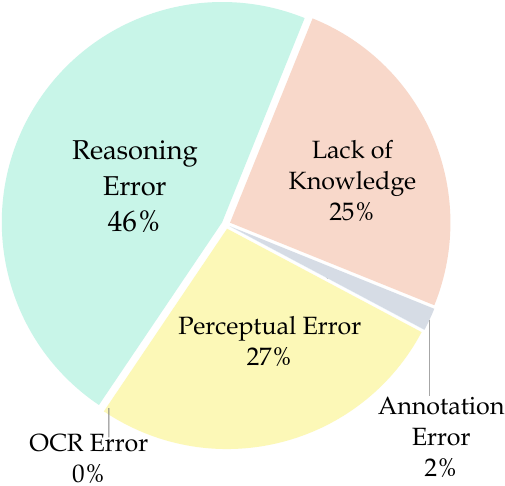}
    \caption{Distribution of 60 annotated GPT-4o errors.}
    \label{fig:error_type}
\end{figure}

Following the MMMU error analysis, we analyze 60 error cases from GPT-4o in the Vision setting to better understand the error reasons (\autoref{fig:error_type}). Consistent with MMMU findings, the errors are broadly categorized into three main types: perception errors, knowledge errors, and reasoning errors. However, reasoning errors account for 46\% of cases, a significant increase from the original MMMU distribution (26\%). Within perception errors, text recognition and OCR do not prove to be the primary bottleneck. Instead, the main challenges lie in the integration and interpretation of visual and textual information. This shift in error distribution highlights the increased difficulty for models in transitioning from accurate perception to complex multimodal reasoning.

\subsection{Response Length Comparison}

\begin{figure}[ht]
    \centering
    \includegraphics[width=0.85\linewidth]{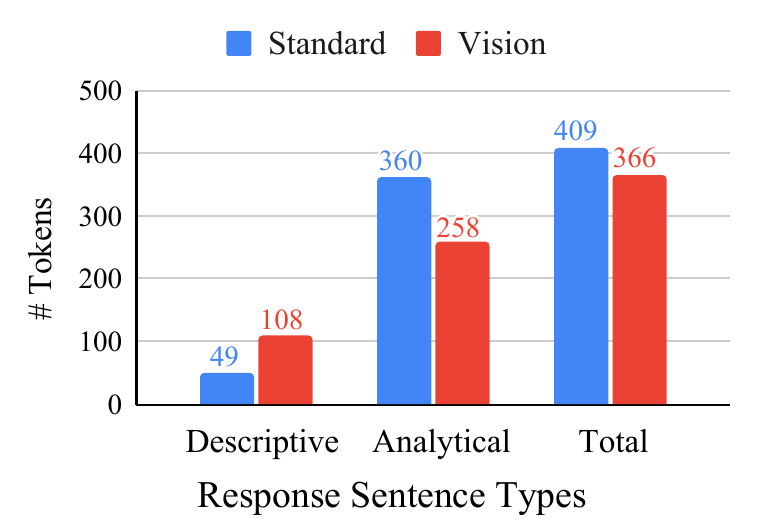}
    \caption{GPT-4o outputs' length comparison between the Standard and Vision settings.}
    \label{fig:model_standard_vision}
\end{figure}

One interesting observation we have from the previous qualitative examples is that responses (especially the reasoning sentences) of GPT-4o under the Vision Input setting seem to be shorter than the Standard setting. We quantify this phenomenon by asking another LLM (Qwen2-72B-Instruct~\citep{yang2024qwen2}) to classify the GPT-4o's responses into ``Descriptive'' sentences and ``Analytical'' sentences. As shown in \autoref{fig:model_standard_vision}, GPT-4o generates significantly shorter responses but uses more tokens for ``Descriptive'' rather than ``Analytical''. One possible reason is that the increased cognition workload of the vision inputs requires the model to focus more on visual processing, which distracts the model from generating extensive reasoning chains. 

%% file: sec/4_guide.tex
\section{Guide for Future Model Training}

The results of MMMU-Pro provide valuable insights into the challenges faced by current multimodal models and suggest several promising directions for future model development.

\noindent\textbf{Scaling of LLM Backbones.}
As demonstrated in \autoref{tab:overall_results}, increasing the scale of large language model (LLM) backbones consistently enhances both perception and reasoning capabilities. For example, larger models such as GPT-4o outperform their smaller counterparts like GPT-4o mini, while LlavaOneVision-72B achieves better results than LlavaOneVision-7B. Similarly, InternVL2-78B demonstrates superior performance compared to InternVL2-8B. This trend underscores the importance of scaling as a critical factor in improving multimodal understanding and reasoning.



\noindent\textbf{More Capable Vision Encoders that Highlights Visual Representation Learning.}
We train two Cambrian~\cite{tong2024cambrian} models on 1M Cambrian data with two different vision encoders to explore their impact (more details of the setup are in \autoref{Appendix:vision_encoder}). As shown in \autoref{tab:vision_encoders}, encoders such as Siglip ViT-SO400M-14~\citep{zhai2023sigmoid}, trained with extensive language supervision, perform well on MMMU (Val) but struggle on MMMU-Pro (Vision). In comparison, self-supervised encoders like DINOv2 ViT-G-14~\citep{oquab2023dinov2} achieve better results on the Vision input setting.  These findings suggest future work may focus on further enhancing visual feature learning while exploring the integration of language-based training objectives with self-supervised training objectives.

\noindent\textbf{Better Integration of Vision and Text Modalities.}
Integration of visual and textual information remains a key challenge for multimodal models. Current architectures often struggle with tasks requiring deep cross-modal understanding. Developing models with better cross-modal attention and effective feature fusion is critical to bridge this gap.

\noindent\textbf{CoT Data Generation.}
The CoT prompting technique shows significant benefits in reasoning-heavy domains within MMMU-Pro, as reflected in \autoref{fig:cot_results} and \autoref{tab:accuracy_comparison}. While domains like \textit{Tech and Engineering} and \textit{Business} see notable improvements, CoT performance remains weak or even detrimental in areas such as \textit{Art and Design}. To address these gaps, future efforts focus on synthesizing more diverse reasoning-intensive CoT data and tailoring strategies for domains where CoT impact is minimal. Leveraging inference-compute concepts~\citep{welleck2024decoding} further enhances CoT capabilities, enabling models to generalize more effectively across varied reasoning tasks.

\noindent\textbf{Text-Rich Image Generation in Reasoning Scenarios.}  
Our analysis shows that strong OCR accuracy and reasoning performance on traditional benchmarks do not always translate to success on MMMU-Pro Vision. A potential reason is the lack of training data with text-rich images in reasoning-intensive contexts. To address this, we developed a tool leveraging the MMMU-Pro Vision human annotation process. This tool processes a JSON file with questions and images and outputs screenshots embedding both. Such tools can further generate similar datasets at scale, enhancing models’ ability to integrate visual and textual information in real-world scenarios.


\begin{table}[!t]
\centering
\vspace{-5pt}
\resizebox{0.9\linewidth}{!}{
\begin{tabular}{@{}lcc@{}}
\toprule
\multirow{2}{*}{\textbf{Method}} & \multicolumn{1}{c}{\textbf{MMMU}} & \multicolumn{1}{c}{\textbf{MMMU-Pro}} \\ 
 & \textbf{(Val)} & \textbf{(Vision)} \\ \midrule
DINOv2 ViT-G-14 & 37.1 & 17.4 \\
Siglip ViT-SO400M-14 & 37.9 & 16.7 \\ \bottomrule
\end{tabular}
}
\caption{Performance of an MLLM with different vision encoders on MMMU and MMMU-Pro.}
\vspace{-15pt} 
\label{tab:vision_encoders}
\end{table}

%% file: appendix/X_related_work.tex
\section{Related Work}
\noindent\textbf{Multimodal Large Language Models.} Recent progress in multimodal AI has been marked by innovative training approaches~\citep{lu2019vilbert,chen2020uniter,zhou2020unified,zhang2021vinvl,li2020oscar,alayrac2022flamingo,awadalla2023openflamingo}. Inspired by the success of large language models, researchers have developed various models with improved instruction-following capabilities~\citep{liu2023llava,liu2023improved,liu2024llava,li2024llava,dai2023instructblip,zhu2023minigpt,zhang2023llama,gao2023llama,ye2023mplug,ye2023mplug2,zhao2023svit,li2023otter,monajatipoor2023metavl,zhao2023mmicl,li2024llavainterleave,lin2024vila,internlmxcomposer2_5}. Proprietary models such as GPT-4V~\citep{openai2023gpt4v}, GPT-4o ~\citep{gpt4o}, Gemini ~\citep{deepmind_gemini_report}, and Claude-3.5 ~\citep{claude-3.5-sonnet} have demonstrated strong performance across various vision-language tasks. However, a significant challenge remains in accurately evaluating the capabilities of these advanced LMMs, highlighting the need for more robust and comprehensive benchmarks. 

\noindent\textbf{MLLM Benchmarks.}
The rise of more advanced multimodal pre-training and instruction tuning has exposed the limitations of earlier benchmarks like VQA~\citep{VQA,goyal2017making}, OK-VQA~\citep{okvqa}, and MSCOCO~\citep{lin2014microsoft}, which no longer suffice to evaluate the full spectrum of LMMs capabilities. To address this, recent benchmarks such as LAMM~\citep{yin2023lamm}, LVLM-eHub~\citep{xu2023lvlm}, SEED~\citep{li2023seed}, MMBench~\citep{liu2023mmbench},CV-Bench~\citep{tong2024cambrian}, MM-Vet~\citep{yu2023mm}, Mantis~\citep{jiang2024mantis}, and BLINK~\citep{fu2024blink} have emerged, covering aspects from basic perception to hallucination detection~\citep{cui2023holistic,liu2023hallusionbench}. 
However, existing benchmarks often fall short in evaluating expert-level domain knowledge and complex reasoning~\citep{lu2023mathvista,zhang2024mathverse}. While MMMU~\citep{yue2024mmmu} made strides by incorporating multimodal, college-level questions, it still permits text-only models to find shortcuts~\citep{lu2023vim,zhang2024mathverse}. 
To address these limitations, we introduce MMMU-Pro, a more robust benchmark that removes text-only answerable questions, expands candidate options, and includes a vision-only input setting to better reflect real-world multimodal scenarios.

%% file: sec/5_conclusion.tex
\section{Conclusion}
MMMU-Pro offers a stronger multimodal understanding and reasoning benchmark than its predecessor MMMU. Our results show MMMU-Pro's effectiveness in exposing current state-of-the-art model limitations, with significant performance drops across all tested systems. MMMU-Pro highlights critical research directions: 1) Developing models with consistent performance across settings, particularly bridging standard and vision-only input gaps. 2) Enhancing vision-text integration for complex mixed-format inputs. 3) Advancing reasoning techniques to address MMMU-Pro's heightened question complexity.

\section*{Ethical Statement}
The MMMU-Pro benchmark is designed with ethical considerations to ensure fair and responsible AI evaluation. The dataset excludes sensitive content, and the assessment focuses on testing multimodal capabilities without introducing bias. We aim for transparency in reporting model limitations and encourage further research to address any societal impacts related to the use of these models in real-world applications.

\section*{Limitations}
While MMMU-Pro improves upon existing benchmarks by filtering out text-only solvable questions and introducing a vision-only setting, some limitations remain. The dataset may still contain subtle statistical shortcuts that models can exploit, and its scope is limited to predefined disciplines and question formats. Additionally, while the vision-only input setting increases difficulty, it does not fully capture the complexities of human perception. Lastly, our reliance on approximated human performance rather than direct evaluation introduces potential biases in reporting accurate human expert performance.

%% file: sec/X_appendix.tex
\appendix
\newpage
\clearpage
\onecolumn
\phantomsection
\label{list:list_of_appendix}
\DoToC
\clearpage
\input{appendix/X_evaluation_prompts}
\input{appendix/X_human_performance}

\input{appendix/X_quality_ensure}
\input{appendix/X_cot}

\input{appendix/X_vision_encoder}
\input{appendix/X_compare_two_setting}
\input{appendix/X_compare_across_disciplines}
\input{appendix/X_aug_comparison_example}
\input{appendix/X_token_length_examples}
\input{appendix/X_comparison_example}

%% file: appendix/X_evaluation_prompts.tex
\newpage
\section{Evaluation Prompts}

\begin{promptbox}[Evaluation Prompts: OCR Prompt]{darkblue}
\label{ocr_prompt}
\textbf{OCR Prompt:}\\
   "\textbf{Write out the multiple-choice question in the image} and then solve it. The last line of your response should be of the following format: 'Answer: \$LETTER' (without quotes) where LETTER is one of the options. Think step by step before answering."
\\
\\
\textbf{w/o OCR Prompt:}\\
"Answer the following multiple-choice question in the image. The last line of your response should be of the following format: 'Answer: \$LETTER' (without quotes) where LETTER is one of the options. Think step by step before answering."

\end{promptbox}

\begin{promptbox}[Evaluation Prompts: Direct vs CoT]{lightgreen}
\textbf{Direct:}\\
      "Answer directly with the option letter from the given choices."
\\\\
\textbf{CoT:}\\
"Answer the following multiple-choice question. The last line of your response should be of the following format: 'Answer: \$LETTER' (without quotes) where LETTER is one of the options. Think step by step before answering."
\end{promptbox}

\begin{promptbox}[Evaluation Prompts: OCR Task]{orange}
\textbf{OCR Task Prompt:}\\
"Extract and output the full text of the question, including any introductory descriptions, as well as the corresponding answer choices from the multiple-choice question in the image. Exclude any text from associated images or the question number. Perform OCR only; do not attempt to solve the question."
\end{promptbox}

\begin{promptbox}[Evaluation Prompts: Split Response Task]{gray}
\textbf{Split Response Task Prompt:}\\
Your task is to split the given answer into two distinct parts: the part that describes the question and the part that analyzes the answer. This is a splitting task, so ensure you do not omit any content or generate any additional content not present in the input. Follow these guidelines:\\

1. Description of the Question:\\
- Extract the portion of the answer that describes the question being addressed.\\
- Ensure that this part is clear and provides enough context to understand the question.\\

2. Analysis of the Answer:\\
- Extract the portion of the answer that provides the analysis or reasoning behind the answer.\\
- Ensure that this part is detailed and provides a complete explanation or solution.\\

Please split the following answer into the two parts described above and output them in JSON format:

Answer: \$LETTER

\{ \\
"description\_of\_question": "Extracted description of the question", \\
"analysis\_of\_answer": "Extracted analysis of the answer" \\
\}
\end{promptbox}

%% file: appendix/X_human_performance.tex
\newpage
\section{Approximating Human Expert Performance}
\label{sec:apdx:human}
Establishing a reliable benchmark for human performance on MMMU-Pro is crucial to evaluating the true capabilities of multimodal AI models. Conducting new and rigorous human evaluations, however, is both time-consuming and expensive. To address this issue, we developed an approximation method based on the existing human evaluation data from the original MMMU. The resulting estimates are presented in \autoref{tab:human_results}.

\begin{table*}[!h]
\centering
\small
\begin{tabular}{lccccccc}
\toprule
 & \textbf{Overall} & \begin{tabular}[c]{@{}c@{}}\textbf{Art} \& \\ \textbf{Design}\end{tabular} & \textbf{Business} & \textbf{Science} & \begin{tabular}[c]{@{}c@{}}\textbf{Health} \& \\ \textbf{Medicine}\end{tabular} & \begin{tabular}[c]{@{}c@{}}\textbf{Human} \& \\ \textbf{Social Sci.}\end{tabular} & \begin{tabular}[c]{@{}c@{}}\textbf{Tech} \& \\ \textbf{Eng.}\end{tabular} \\ \midrule
Low & 73.0 & 77.4 & 77.9 & 78.5 & 65.2 & 63.6 & 73.5 \\
Medium & 80.8 & 83.3 & 88.4 & 84.9 & 72.8 & 75.8 & 78.2 \\
High & 85.4 & 85.7 & 89.5 & 86.0 & 84.8 & 81.8 & 84.4 \\ 
\bottomrule
\end{tabular}%
\caption{Estimated human performance on MMMU-Pro across different disciplines, based on the original MMMU evaluation data. The table presents low, medium, and high performance estimates in terms of overall accuracy and discipline-specific breakdowns.}
\label{tab:human_results}
\end{table*}

The validity of using this approximation method relies on several key factors. Firstly, the core content and difficulty of the questions in MMMU-Pro remain unchanged from those in the original MMMU, supporting the use of the original human performance data as a valid proxy. Secondly, in the initial MMMU evaluation, human experts were required to document their problem-solving processes, which significantly reduced the likelihood of random guessing. For questions lacking detailed solution processes, we simulated random selection from expanded candidate options and recalculated the accuracy. Finally, human experts inherently excel at seamlessly integrating visual and textual information, suggesting that their performance in a purely visual input setting would be analogous to their performance in the original format.

Given that the 577 questions in MMMU-Pro are sourced from the MMMU validation set, we extracted the corresponding data from the evaluations of the 90 human experts involved in the original MMMU assessment. We categorized and counted these questions based on whether they included a detailed solution process (\textbf{w/ Solution}) or were subjected to guessing due to the lack of a detailed solution process (\textbf{w/o Solution}). We then counted the correct and incorrect answers in each category, as summarized in \autoref{tab:human_overall_results}. Specifically, the categorization is defined in \autoref{eq:human1}:

\begin{equation}
\begin{split}
\text{Num}_{\text{total}} & = \text{Num}_{\text{w/o Solution}} + \text{Num}_{\text{w/ Solution}} \\
& = \text{Num}_{\text{w/o Solution(wrong)}} + \text{Num}_{\text{w/o Solution(correct)}} \\
& \quad + \text{Num}_{\text{w/ Solution(wrong)}} + \text{Num}_{\text{w/ Solution(correct)}}
\end{split}
\label{eq:human1}
\end{equation}

Using these counts, we can estimate the lower bound of human performance on MMMU-Pro with \autoref{eq:human2}:

\begin{equation}
\text{Num}_{\text{Estimate(correct)}} = \text{Num}_{\text{w/ Solution(correct)}} + \left\lfloor \left( \frac{\text{Num}_{\text{w/o Solution}}}{\text{Num}_{\text{total}}} \right) \times \text{Num}_{\text{w/o Solution}} \right\rceil
\label{eq:human2}
\end{equation}


This formula considers the number of correctly solved questions with detailed solution processes and the proportion of correctly guessed questions without detailed solution processes, ensuring a conservative estimate.

\begin{table*}
\centering
\small
\begin{adjustbox}{scale=0.75}
\begin{tabular}{lcccc|cccc|cccc}
\toprule
\multirow{2}{*}{} & \multicolumn{4}{c}{\textbf{Low}} & \multicolumn{4}{c}{\textbf{Medium}} & \multicolumn{4}{c}{\textbf{High}} \\ \cmidrule(r){2-5} \cmidrule(r){6-9} \cmidrule(r){10-13} 
 & \multicolumn{1}{c}{\textbf{\begin{tabular}[c]{@{}c@{}}w/o Sol.\\ (w/c)\end{tabular}}} & \multicolumn{1}{c}{\textbf{\begin{tabular}[c]{@{}c@{}}w/ Sol.\\ (w/c)\end{tabular}}} & \multicolumn{1}{c}{\textbf{\begin{tabular}[c]{@{}c@{}}Est.\\ (w/c)\end{tabular}}} & \textbf{Acc} & \multicolumn{1}{c}{\textbf{\begin{tabular}[c]{@{}c@{}}w/o Sol.\\ (w/c)\end{tabular}}} & \multicolumn{1}{c}{\textbf{\begin{tabular}[c]{@{}c@{}}w/ Sol.\\ (w/c)\end{tabular}}} & \multicolumn{1}{c}{\textbf{\begin{tabular}[c]{@{}c@{}}Est.\\ (w/c)\end{tabular}}} & \textbf{Acc} & \multicolumn{1}{c}{\textbf{\begin{tabular}[c]{@{}c@{}}w/o Sol.\\ (w/c)\end{tabular}}} & \multicolumn{1}{c}{\textbf{\begin{tabular}[c]{@{}c@{}}w/ Sol.\\ (w/c)\end{tabular}}} & \multicolumn{1}{c}{\textbf{\begin{tabular}[c]{@{}c@{}}Est.\\ (w/c)\end{tabular}}} & \textbf{Acc} \\ \midrule \addlinespace[0.5em] 
\textbf{Art \& Design} & 4/11 & 11/64 & 19/65 & \textbf{77.4} & 5/1 & 8/70 & 14/70 & \textbf{83.3} & 4/2 & 6/72 & 12/72 & \textbf{85.7} \\
\quad Art & 2/2 & 2/14 & 4/14 & 77.8 & 1/0 & 1/16 & 2/16 & 88.9 & 0/1 & 0/17 & 1/17 & 94.4 \\
\quad Art Theory & 1/2 & 2/18 & 5/18 & 78.3 & 1/1 & 2/19 & 4/19 & 82.6 & 1/1 & 3/18 & 5/18 & 78.3 \\
\quad Design & 1/4 & 4/10 & 5/10 & 66.7 & 1/0 & 2/12 & 3/12 & 80.0 & 1/0 & 1/13 & 2/13 & 86.7 \\
\quad Music & 0/3 & 3/22 & 6/22 & 78.6 & 2/0 & 3/23 & 5/23 & 82.1 & 2/0 & 2/24 & 4/24 & 85.7 \\\midrule
\textbf{Business} & 4/11 & 11/73 & 21/74 & \textbf{77.9} & 4/1 & 6/84 & 11/84 & \textbf{88.4} & 2/3 & 5/85 & 10/85 & \textbf{89.5} \\
\quad Accounting & 0/3 & 3/19 & 6/19 & 76.0 & 2/0 & 1/22 & 3/22 & 88.0 & 0/2 & 1/22 & 3/22 & 88.0 \\
\quad Economics & 0/4 & 4/13 & 5/13 & 72.2 & 1/0 & 1/16 & 2/16 & 88.9 & 1/0 & 0/17 & 1/17 & 94.4 \\
\quad Finance & 1/2 & 2/15 & 4/15 & 78.9 & 0/0 & 1/18 & 1/18 & 94.7 & 0/0 & 2/17 & 2/17 & 89.5 \\
\quad Manage & 2/2 & 2/8 & 4/9 & 69.2 & 1/1 & 2/9 & 4/9 & 69.2 & 1/1 & 2/9 & 4/9 & 69.2 \\
\quad Marketing & 1/0 & 0/18 & 2/18 & 90.0 & 0/0 & 1/19 & 1/19 & 95.0 & 0/0 & 0/20 & 0/20 & 100.0 \\\midrule
\textbf{Science} & 3/12 & 12/72 & 20/73 & \textbf{78.5} & 3/1 & 10/79 & 14/79 & \textbf{84.9} & 3/1 & 9/80 & 13/80 & \textbf{86.0} \\
\quad Biology & 0/5 & 5/13 & 7/13 & 65.0 & 2/0 & 5/13 & 7/13 & 65.0 & 1/1 & 5/13 & 7/13 & 65.0 \\
\quad Chemistry & 0/3 & 3/14 & 4/14 & 77.8 & 0/1 & 2/15 & 3/15 & 83.3 & 1/0 & 2/15 & 3/15 & 83.3 \\
\quad Geography & 2/0 & 0/8 & 2/8 & 80.0 & 0/0 & 1/9 & 1/9 & 90.0 & 0/0 & 1/9 & 1/9 & 90.0 \\
\quad Math & 1/4 & 4/14 & 7/14 & 66.7 & 1/0 & 1/19 & 2/19 & 90.5 & 1/0 & 1/19 & 2/19 & 90.5 \\
\quad Physics & 0/0 & 0/23 & 1/23 & 95.8 & 0/0 & 1/23 & 1/23 & 95.8 & 0/0 & 0/24 & 0/24 & 100.0 \\\midrule
\textbf{Health \& Med.} & 3/22 & 22/58 & 32/60 & \textbf{65.2} & 9/0 & 17/66 & 25/67 & \textbf{72.8} & 5/4 & 6/77 & 14/78 & \textbf{84.8} \\
\quad Basic Med. & 2/2 & 2/9 & 4/10 & 71.4 & 1/0 & 2/11 & 3/11 & 78.6 & 1/0 & 1/12 & 2/12 & 85.7 \\
\quad Clinical Med. & 1/6 & 6/8 & 9/9 & 50.0 & 3/0 & 5/10 & 7/11 & 61.1 & 2/1 & 1/14 & 3/15 & 83.3 \\
\quad Diagnostics & 0/6 & 6/14 & 9/14 & 60.9 & 3/0 & 4/16 & 7/16 & 69.6 & 2/1 & 2/18 & 5/18 & 78.3 \\
\quad Pharmacy & 0/3 & 3/13 & 4/13 & 76.5 & 1/0 & 3/13 & 4/13 & 76.5 & 0/1 & 1/15 & 2/15 & 88.2 \\
\quad Public Health & 0/5 & 5/14 & 6/14 & 70.0 & 1/0 & 3/16 & 4/16 & 80.0 & 0/1 & 1/18 & 2/18 & 90.0 \\\midrule
\textbf{Humani. \& Soc.} & 5/14 & 14/40 & 24/42 & \textbf{63.6} & 3/5 & 9/49 & 16/50 & \textbf{75.8} & 5/3 & 5/53 & 12/54 & \textbf{81.8} \\
\quad History & 1/4 & 4/4 & 6/4 & 40.0 & 1/0 & 1/8 & 2/8 & 80.0 & 0/1 & 1/8 & 2/8 & 80.0 \\
\quad Literature & 2/2 & 2/15 & 5/16 & 76.2 & 1/2 & 2/16 & 5/16 & 76.2 & 2/1 & 0/18 & 3/18 & 85.7 \\
\quad Sociology & 0/5 & 5/8 & 7/9 & 56.3 & 1/2 & 4/9 & 6/10 & 62.5 & 2/1 & 2/11 & 4/12 & 75.0 \\
\quad Psychology & 2/3 & 3/13 & 6/13 & 68.4 & 0/1 & 2/16 & 3/16 & 84.2 & 1/0 & 2/16 & 3/16 & 84.2 \\\midrule
\textbf{Tech \& Eng.} & 3/25 & 25/106 & 39/108 & \textbf{73.5} & 9/4 & 20/114 & 32/115 & \textbf{78.2} & 6/7 & 10/124 & 23/124 & \textbf{84.4} \\
\quad Agriculture & 0/6 & 6/10 & 9/10 & 52.6 & 1/2 & 5/11 & 8/11 & 57.9 & 2/1 & 2/14 & 5/14 & 73.7 \\
\quad Archi. Eng. & 2/2 & 2/17 & 5/17 & 77.3 & 1/1 & 2/18 & 4/18 & 81.8 & 1/1 & 0/20 & 2/20 & 90.9 \\
\quad Computer Sci. & 0/0 & 0/17 & 2/17 & 89.5 & 1/0 & 1/17 & 2/17 & 89.5 & 0/1 & 2/16 & 3/16 & 84.2 \\
\quad Electronics & 0/0 & 0/8 & 1/8 & 88.9 & 0/0 & 0/9 & 0/9 & 100.0 & 0/0 & 0/9 & 0/9 & 100.0 \\
\quad Energy Power & 0/4 & 4/20 & 6/20 & 76.9 & 2/0 & 4/20 & 6/20 & 76.9 & 1/1 & 1/23 & 3/23 & 88.5 \\
\quad Materials & 0/3 & 3/22 & 5/22 & 81.5 & 1/1 & 3/22 & 5/22 & 81.5 & 1/1 & 2/23 & 4/23 & 85.2 \\
\quad Mechanical Eng. & 1/10 & 10/12 & 13/12 & 48.0 & 3/0 & 5/17 & 8/17 & 68.0 & 1/2 & 3/19 & 6/19 & 76.0 \\\midrule
\textbf{Overall} & 22/95 & 95/413 & 156/421 & \textbf{73.0} & 33/12 & 70/462 & 111/466 & \textbf{80.8} & 25/20 & 41/491 & 84/493 & \textbf{85.4} \\
\bottomrule
\end{tabular}%
\end{adjustbox}
\caption{Detailed breakdown of estimated human performance on MMMU-Pro for low, medium, and high performance levels across various disciplines. Abbreviations: "\textbf{w/o Sol.}" (without Solution), "\textbf{w/ Sol.}" (with Solution), "\textbf{Est.}" (Estimate), and "\textbf{w/c}" (number of wrong/correct answers).}
\label{tab:human_overall_results}
\end{table*}

In summary, by leveraging the original MMMU human evaluation data and applying our estimation method, we provide a reasonable approximation of human performance on MMMU-Pro. This approach maintains the human performance benchmark without incurring the substantial costs associated with new expert evaluations.

%% file: appendix/X_quality_ensure.tex
\newpage
\section{Ensuring Quality and Diversity of Expanded Options}

Expanding the number of answer options naturally increases the difficulty of the benchmark, but its effectiveness relies heavily on the quality, diversity, and contextual relevance of these additional options. To ensure this, we implemented a rigorous multi-stage validation process, combining automated and human efforts to produce high-quality results.

\textbf{Initial Model-Based Option Augmentation and Filtering.}
We began by leveraging large language models (LLMs) to automate the initial generation and filtering of expanded options. Specifically, GPT-4o was used to generate additional options, while Claude 3.5 acted as a preliminary filter to remove options that were contextually irrelevant or logically inconsistent. This step significantly reduced the workload for human reviewers by pre-screening the candidates.

\textbf{Two Rounds of Human Review.}
To further enhance quality and eliminate potential issues, we conducted two rounds of meticulous human validation:

\begin{itemize}[leftmargin=*]
    \item \textbf{First Round of Review:} Individual reviewers assessed the expanded options for each question. They ensured that the options were diverse, logically distinct, and free from ambiguity. If any flaws were identified, reviewers were instructed to correct the issues or create new options to maintain the integrity of the question.
    \item \textbf{Second Round of Review:} A double-check process followed, involving two additional human experts who cross-validated each question and its options. This iterative step eliminated any residual inconsistencies or errors and provided an additional layer of assurance.
\end{itemize}

By combining automated methods with multi-stage human validation, we ensured that each expanded option met high standards of quality, robustness, and alignment with the intended challenges of the benchmark. This approach not only addressed potential weaknesses in automated generation but also significantly improved the reliability of the dataset.

%% file: appendix/X_cot.tex
\newpage
\clearpage
\section{Analysis of CoT's Impact}
\label{sec:apdx:cot}

\begin{figure}[ht]
    \centering
    \begin{subfigure}[b]{0.41\textwidth}
        \centering
        \includegraphics[width=\textwidth]{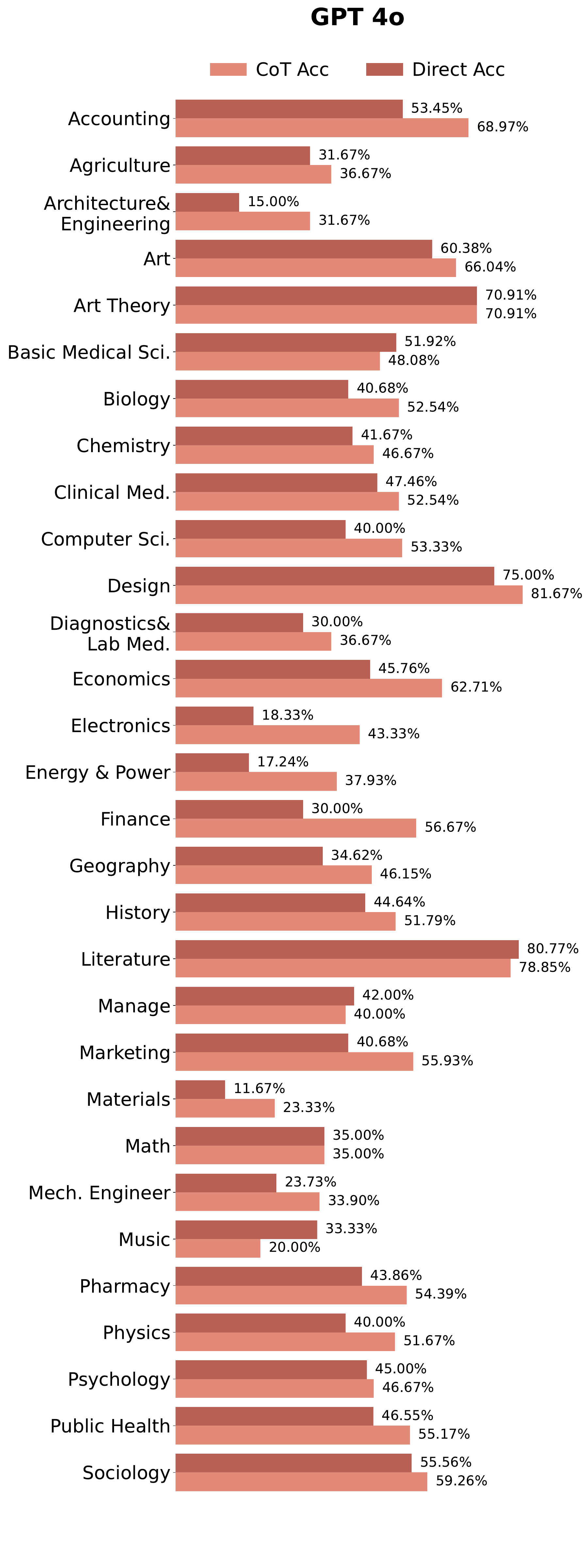} 
        \label{fig:left_image}
    \end{subfigure}
    \hfill
    \begin{subfigure}[b]{0.41\textwidth}
        \centering
        \includegraphics[width=\textwidth]{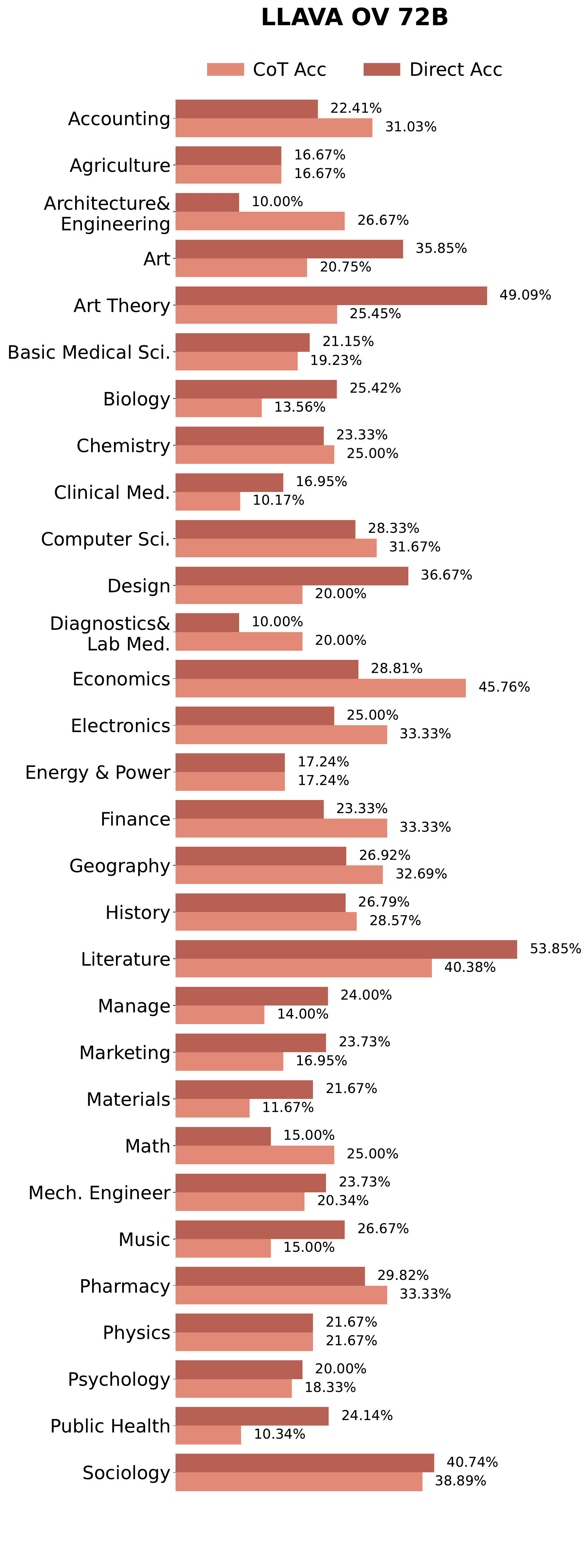}
        \label{fig:right_image}
    \end{subfigure}
    \caption{\centering Comparison of CoT and Direct Accuracy across subcategories within major domains for GPT-4o and LLaVA-OneVision 72B. }
    \label{fig:sub_cot}
\end{figure}

%% file: appendix/X_vision_encoder.tex
\newpage
\section{Experimental Setup of Vision Encoder Impact}
\label{Appendix:vision_encoder}
To evaluate the influence of vision encoders on model performance, we conduct experiments using the open-source architecture Cambrian-1. These experiments fix both the training data (Cambrian-1 1M SFT data) and the large language model (Llama 3.1 8B) to isolate the impact of different vision encoders. Inspired by Cambrian-1: A Fully Open, Vision-Centric Exploration of Multimodal LLMs~\citep{tong2024cambrian}, we follow their methodology by interpolating visual features to a fixed number of tokens (576) and concatenating them along the feature dimension.

%% file: appendix/X_compare_two_setting.tex
\section{Comparison of GPT-4o's responses between Standard and Vision Input settings}
\label{Appendix:compare_two_setting}

\begin{figure*}[!h]
    \centering
     \includegraphics[width=1\linewidth]{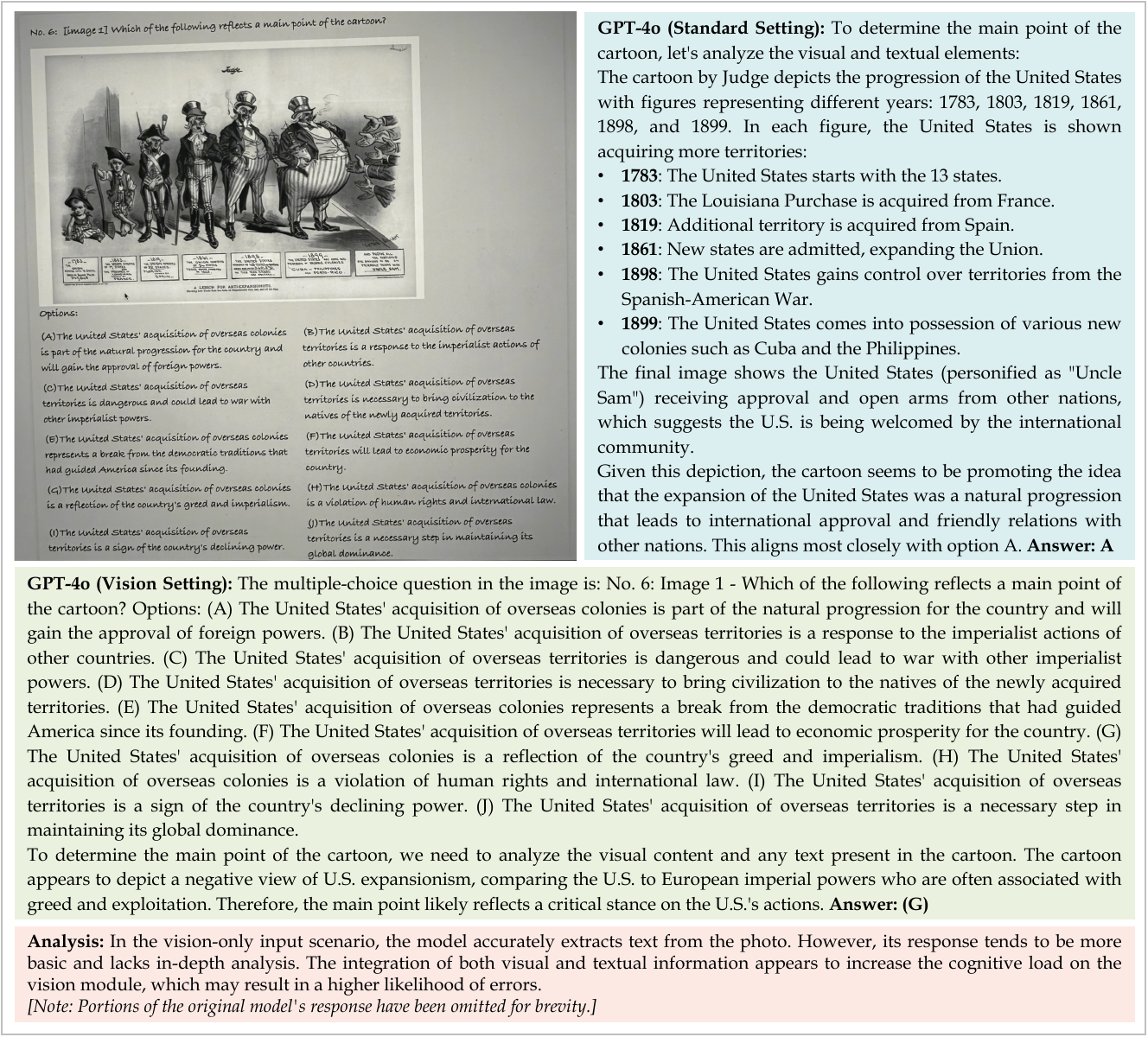}
    \caption{Comparison of GPT-4o's responses between Standard and Vision Input settings.}
    \label{fig:case_study2}
    \vspace{-15pt} 
\end{figure*}

%% file: appendix/X_compare_across_disciplines.tex
\clearpage
\section{CoT vs. Direct Acc: Model Differences Across Disciplines}
\label{Appendix:compare_across_disciplines}

\begin{table*}[!h]
\centering
\resizebox{0.9\linewidth}{!}{
\begin{tabular}{@{}l|cccccc@{}}
\toprule
\multirow{2}{*}{\textbf{Discipline}} & \multicolumn{3}{c}{\textbf{LLaVA-OneVision-72B}} & \multicolumn{3}{c}{\textbf{GPT4o}} \\
 \cmidrule(r){2-4} \cmidrule(l){5-7}
 & \textbf{CoT Acc} & \textbf{Direct Acc} & \textbf{Difference}  & \textbf{CoT Acc} & \textbf{DIRECT Acc} & \textbf{Difference} \\
\midrule
\textbf{Art and Design} & 20.42\% & 37.53\% & -17.12\%  & 63.14\% & 61.55\% & 1.58\% \\
\textbf{Science} & 23.89\% & 22.61\% & 1.28\%  & 46.67\% & 38.46\% & 8.22\% \\
\textbf{Business} & 29.26\% & 24.50\% & 4.76\% & 57.45\% & 42.79\% & 14.66\% \\
\textbf{Humanities and Social Science} & 32.14\% & 36.60\% & -4.46\%  & 60.08\% & 57.87\% & 2.21\% \\
\textbf{Health and Medicine} & 19.22\% & 20.78\% & -1.56\%  & 49.68\% & 44.34\% & 5.34\% \\
\textbf{Tech and Engineering} & 22.98\% & 20.65\% & 2.33\% & 37.72\% & 23.23\% & 14.49\% \\
\bottomrule
\end{tabular}}
\caption{Comparison of CoT and direct accuracy of two representative models across disciplines in the Vision Input setting. Difference = CoT Acc. - Direct Acc.}
\vspace{-5pt}
\label{tab:accuracy_comparison}
\end{table*}

%% file: appendix/X_aug_comparison_example.tex
\section{Comparison With and Without Augmented Options}
\label{Appendix:Aug_Comparison}

\begin{figure}[!h]
    \centering
    \includegraphics[width=1.0\linewidth]{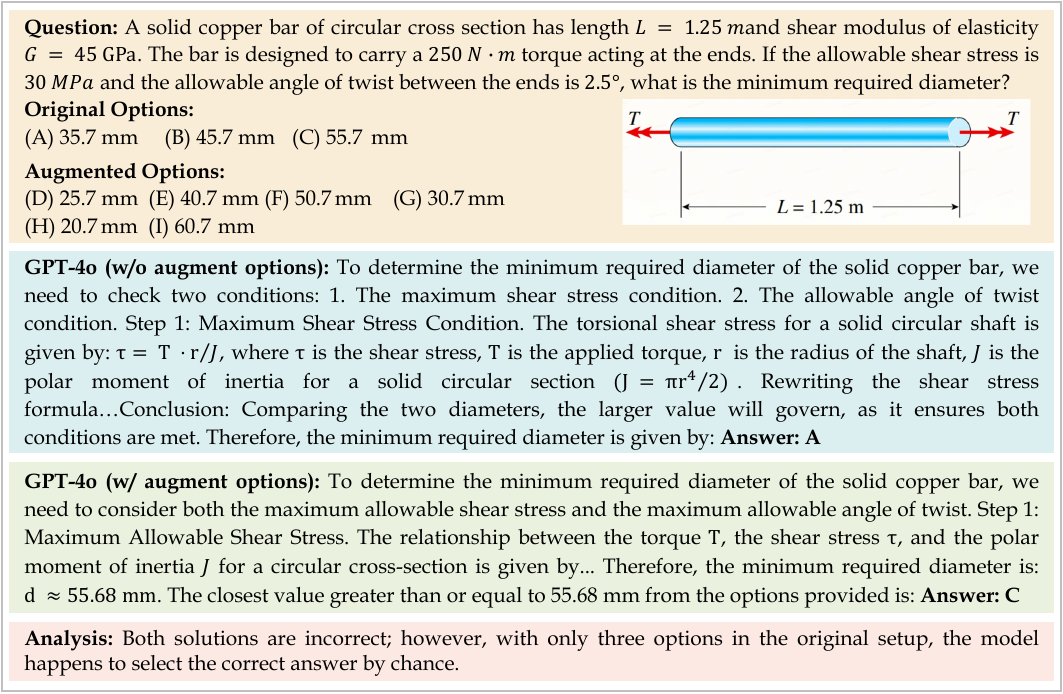}
    \caption{Comparison of GPT-4o's responses with and without augmented options.}
    \label{fig:case_study1}
\end{figure}

%% file: appendix/X_token_length_examples.tex
 \clearpage
\section{Comparison of Model Outputs Across Different Input Modes}
\label{Appendix:Output_Comparison}

\begin{figure*}[!htbp]
    \centering
\includegraphics[width=1\linewidth]{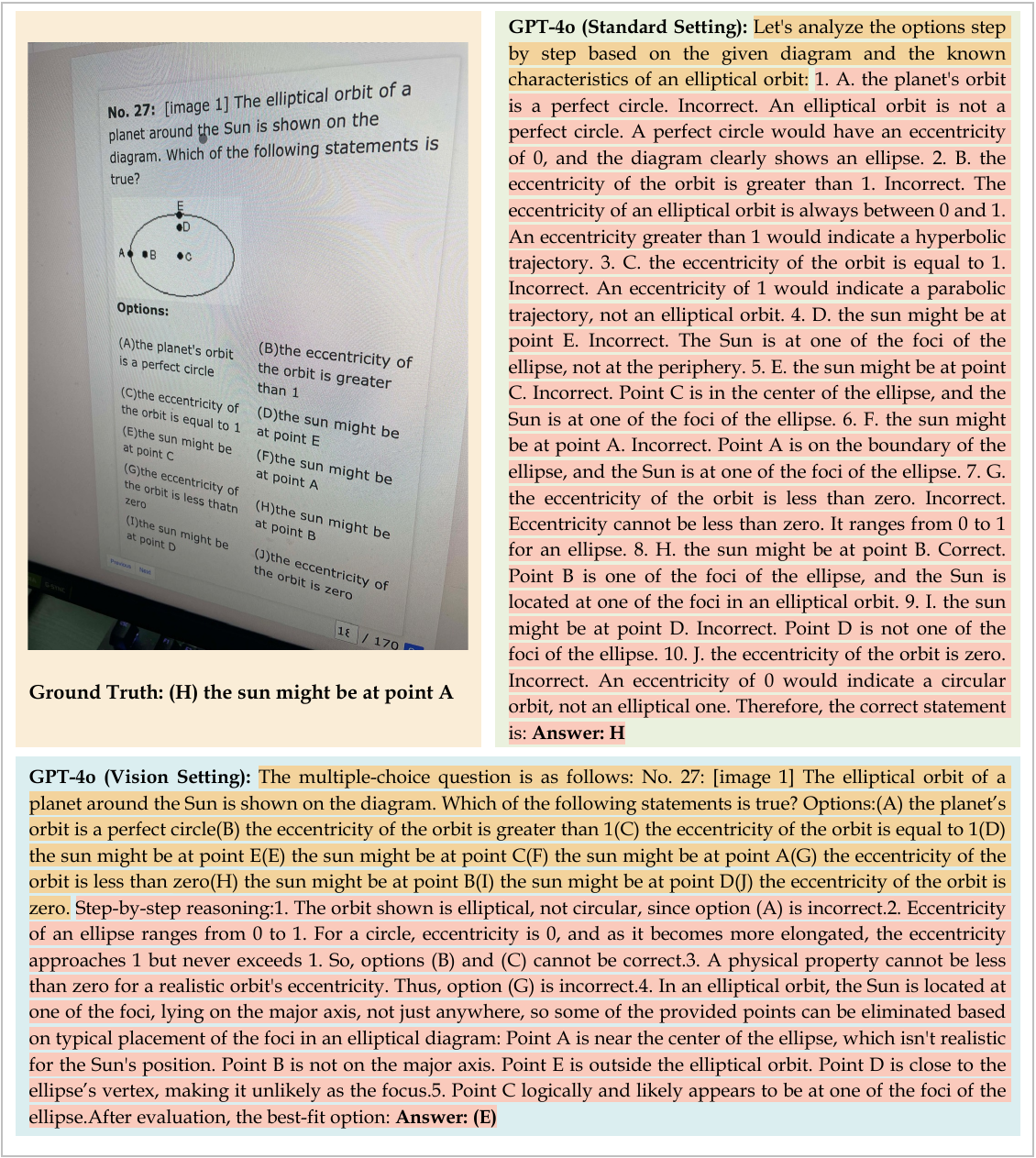}
    \caption{Example of GPT-4o output comparison in different settings. \textbf{Descriptions} are highlighted in orange, and \textbf{analyses} are highlighted in light red.}
\label{fig:token_1}
\end{figure*}
\newpage

\begin{figure*}[!htbp]
    \centering
\includegraphics[width=1\linewidth]{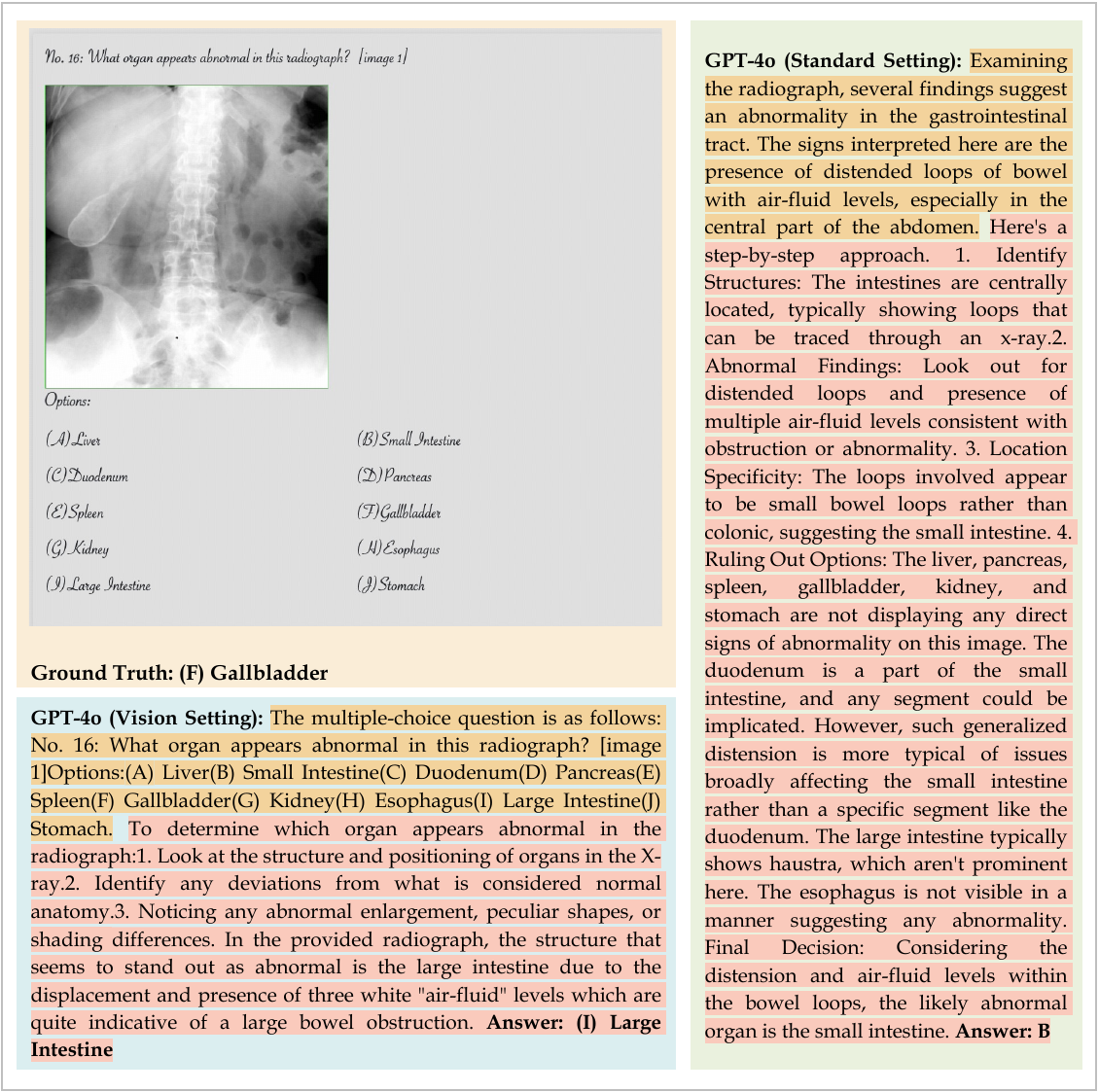}
    \caption{Example of GPT-4o output comparison in different settings. \textbf{Descriptions} are highlighted in orange, and \textbf{analyses} are highlighted in light red.}
\label{fig:token_2}
\end{figure*}

%% file: appendix/X_comparison_example.tex
\clearpage
\section{Qualitative Examples}
\label{Appendix:Setting_Comparison}


\renewcommand{\thefigure}{\arabic{figure}}

\subsection{Art and Design: Art}
\begin{figure*}[!htbp]
    \centering
\includegraphics[width=0.9\linewidth]{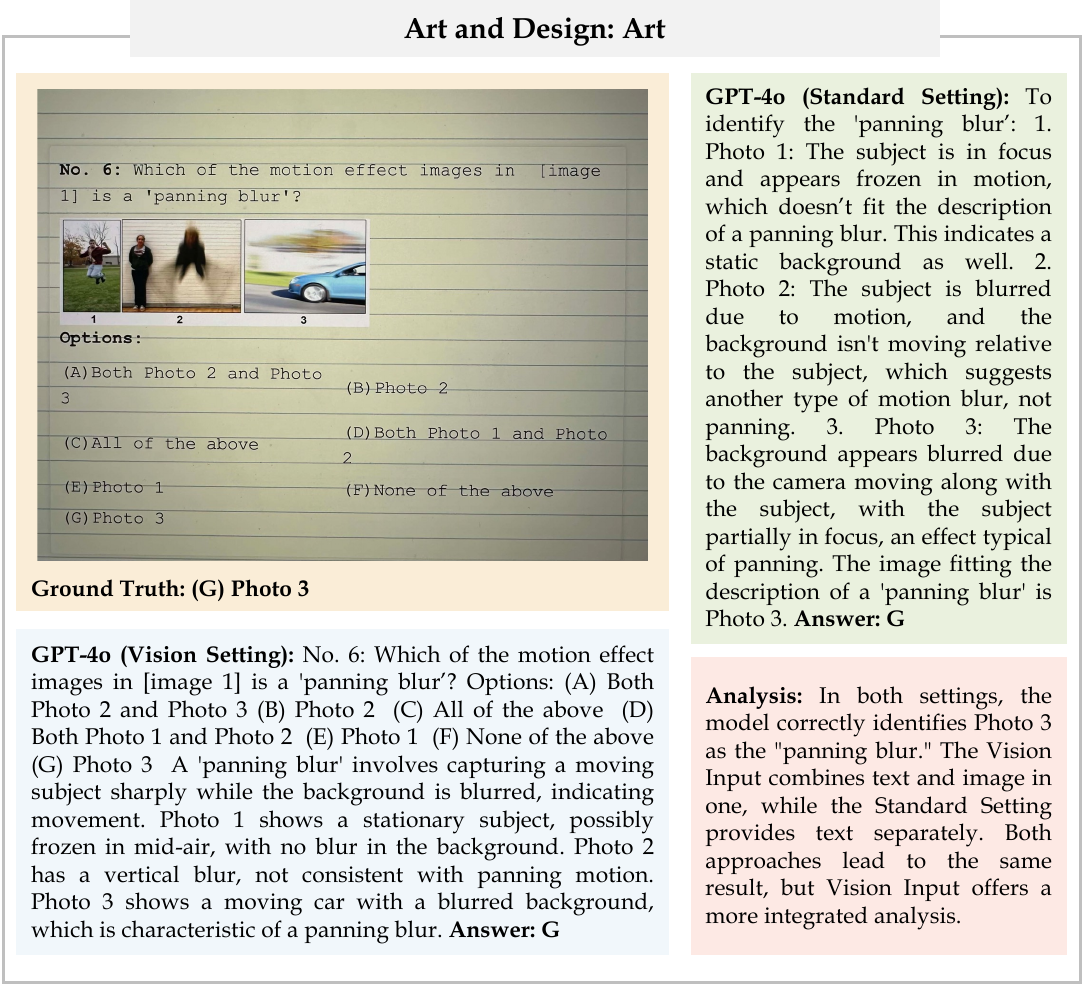}
    \caption{Example of a different input setting in Art and Design (subfield: Art). \quad
    \newline \centering \hyperref[list:list_of_appendix]{Back to Appendix}}
    \addcontentsline{afg}{appfigures}{\protect\numberline{\thefigure}Art and Design: Art}
\label{fig:art}
\end{figure*}

\newpage
\subsection{Art and Design: Art Theory}
\begin{figure*}[!htbp]
    \centering
\includegraphics[width=0.9\linewidth]{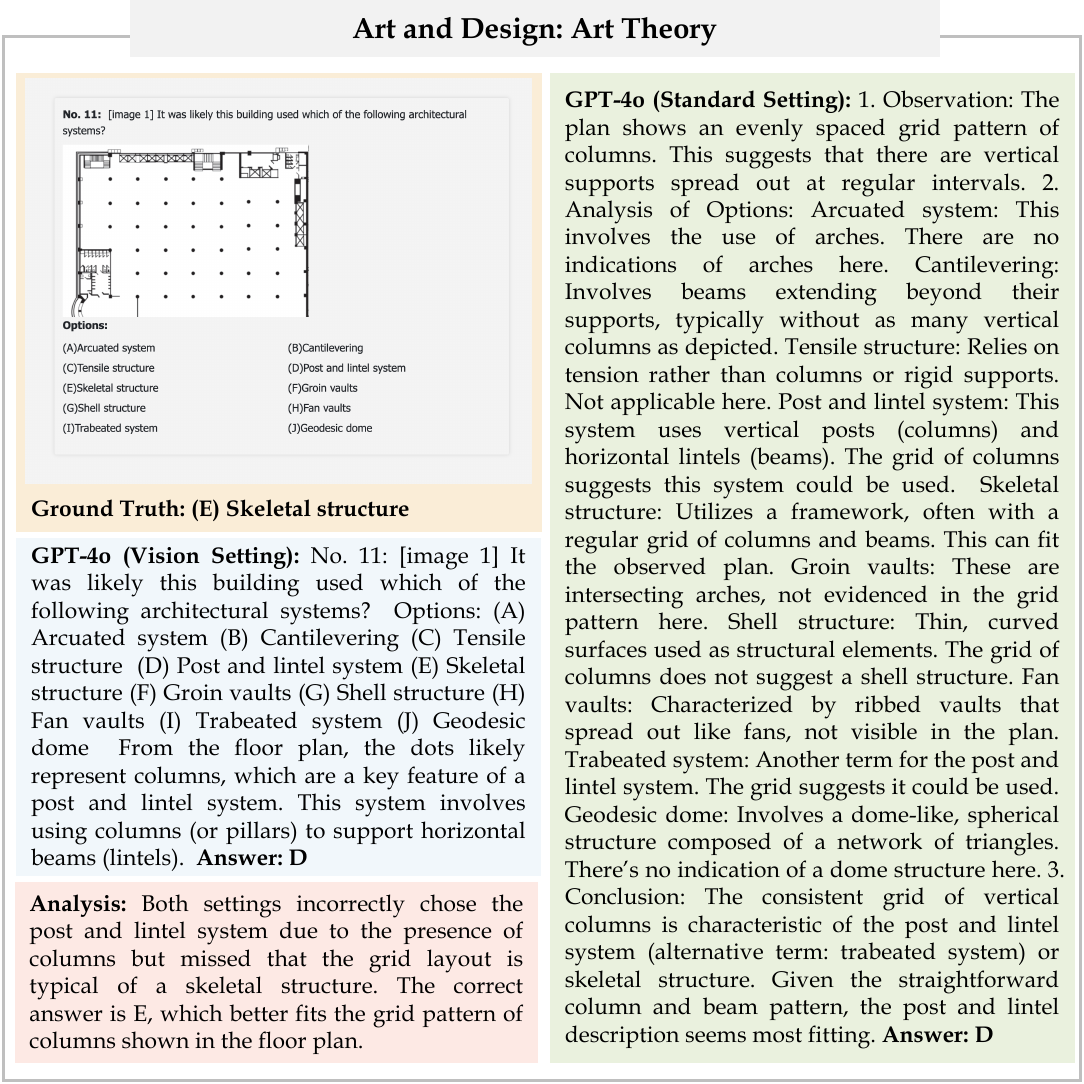}
    \caption{Example of a different input setting in Art and Design (subfield: Art Theory).\newline \centering \hyperref[list:list_of_appendix]{Back to Appendix}}
    \addcontentsline{afg}{appfigures}{\protect\numberline{\thefigure}Art and Design: Art Theory}
\label{fig:art_theory}
\end{figure*}

\newpage
\subsection{Art and Design: Design}
\begin{figure*}[!htbp]
    \centering
\includegraphics[width=0.9\linewidth]{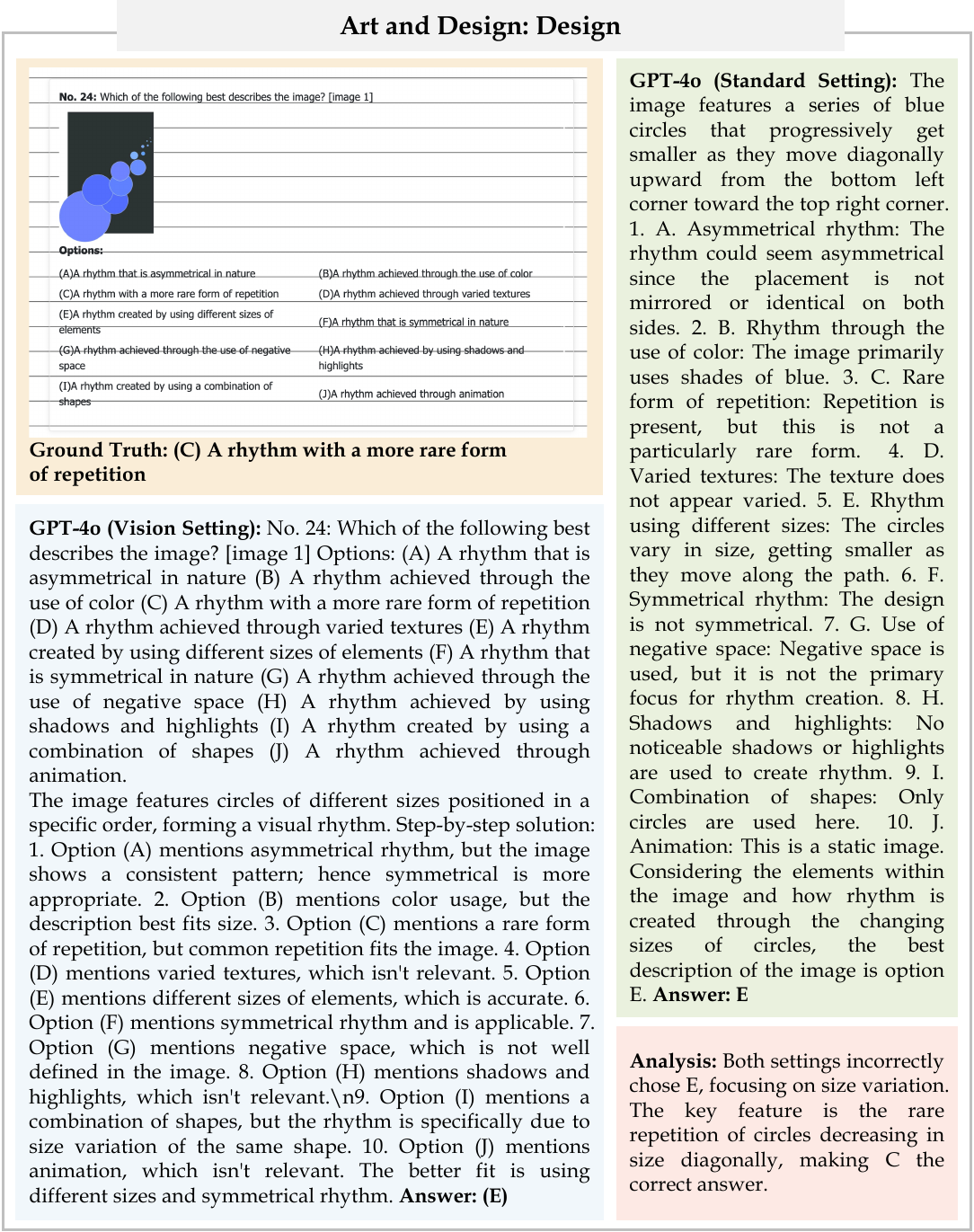}
    \caption{Example of a different input setting in Art and Design (subfield: Design).\newline \centering \hyperref[list:list_of_appendix]{Back to Appendix}}
    \addcontentsline{afg}{appfigures}{\protect\numberline{\thefigure}Art and Design: Design}
\label{fig:design}
\end{figure*}

\newpage
\subsection{Art and Design: Music}
\begin{figure*}[!htbp]
    \centering
\includegraphics[width=0.9\linewidth]{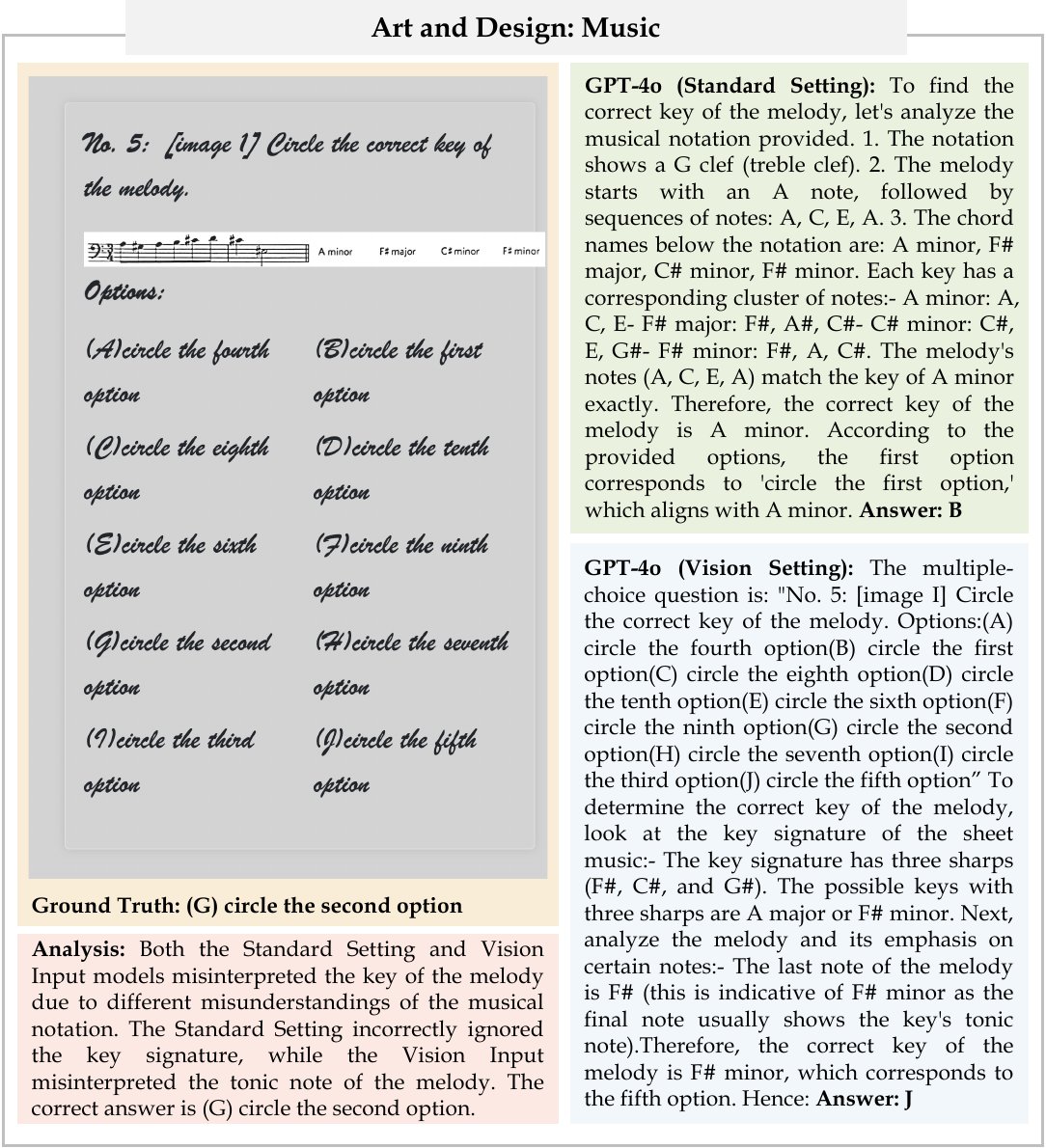}
    \caption{Example of a different input setting in Art and Design (subfield: Music).\newline \centering \hyperref[list:list_of_appendix]{Back to Appendix}}
    \addcontentsline{afg}{appfigures}{\protect\numberline{\thefigure}Art and Design: Music}
\label{fig:music}
\end{figure*}

\newpage
\subsection{Business: Accounting}
\begin{figure*}[!htbp]
    \centering
\includegraphics[width=0.9\linewidth]{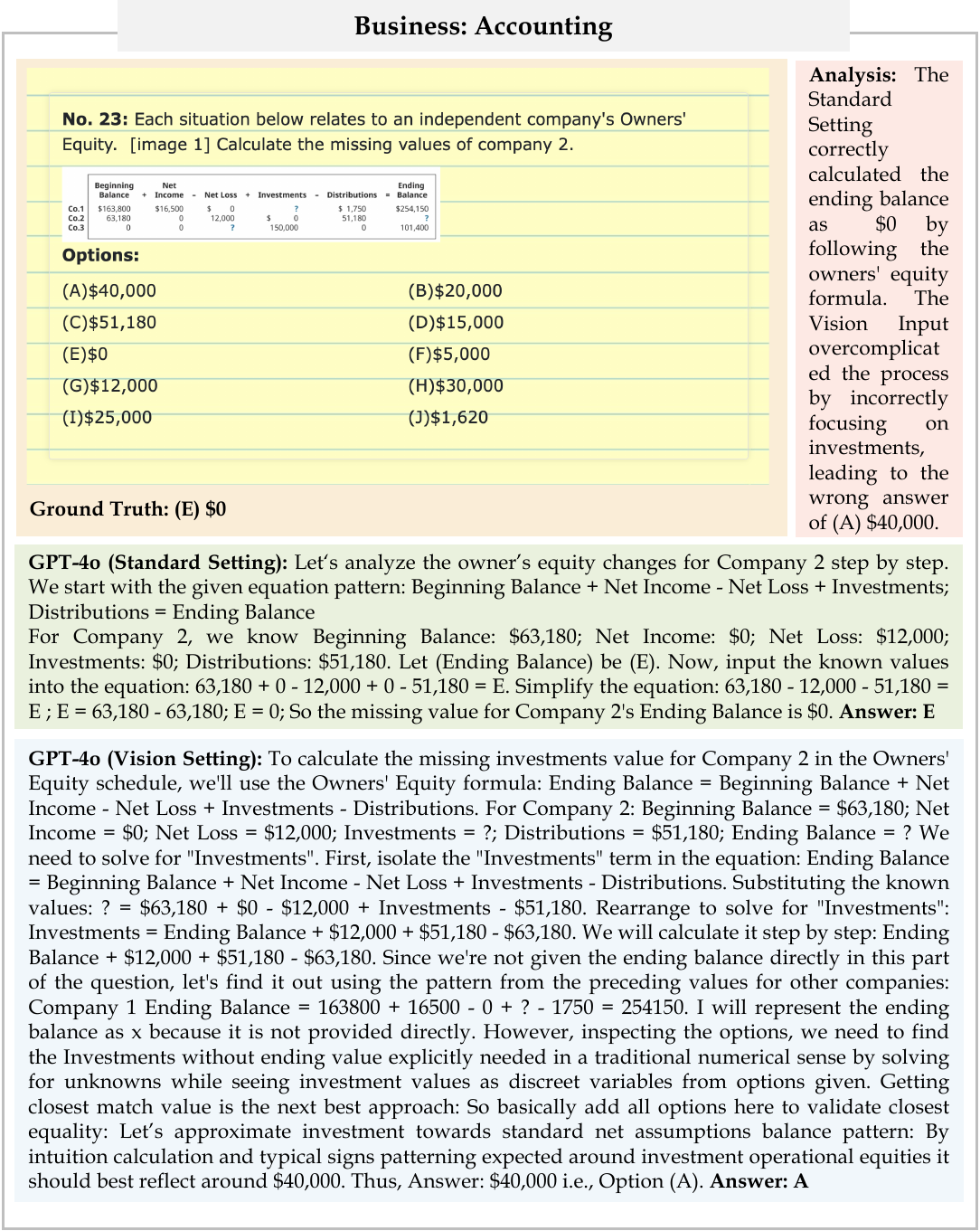}
    \caption{Example of a different input setting in Business (subfield: Accounting).\newline \centering \hyperref[list:list_of_appendix]{Back to Appendix}}
    \addcontentsline{afg}{appfigures}{\protect\numberline{\thefigure}Business: Accounting}
\label{fig:accounting}
\end{figure*}
\newpage

\newpage
\subsection{Business: Economics}
\begin{figure*}[!htbp]
    \centering
\includegraphics[width=0.9\linewidth]{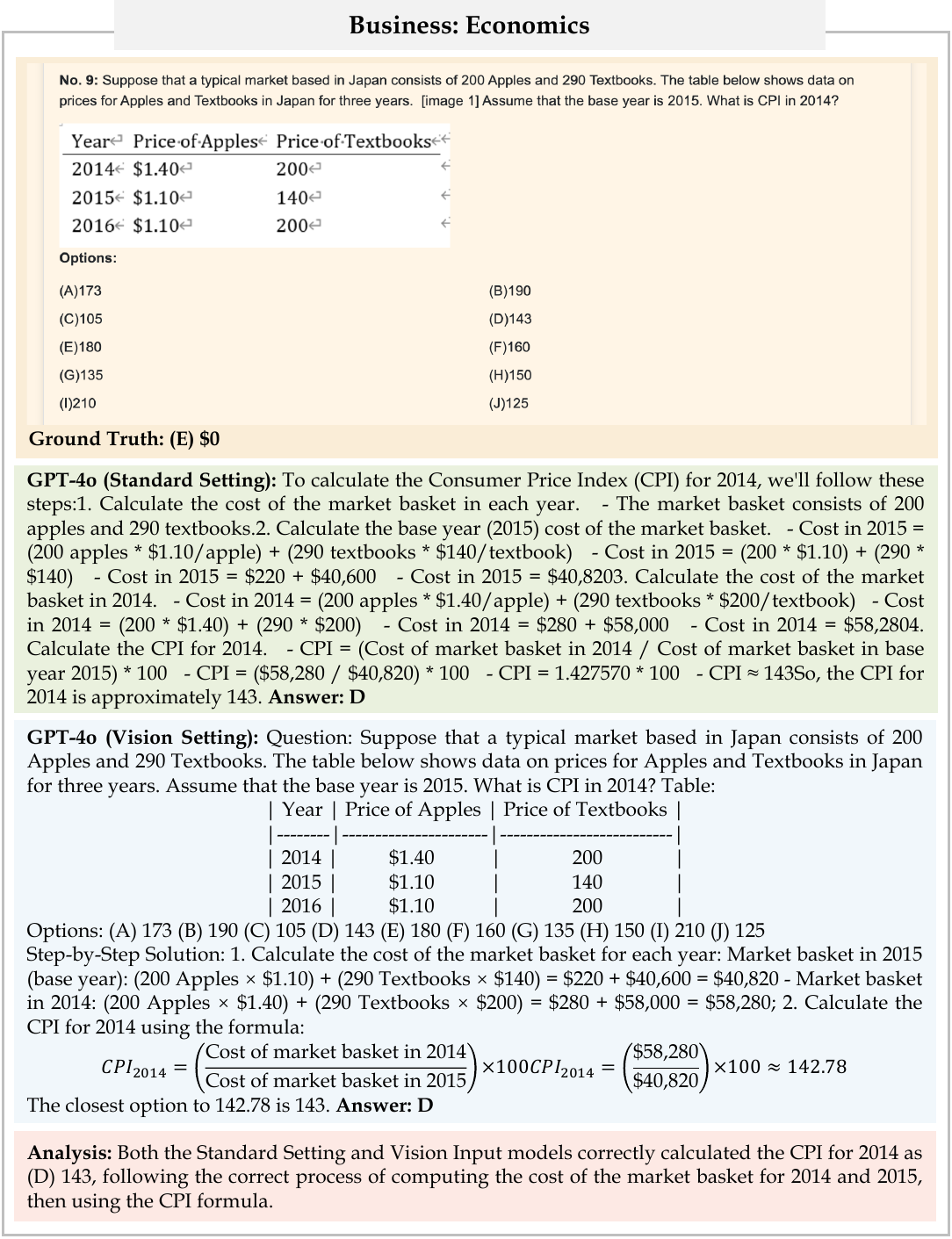}
    \caption{Example of a different input setting in Business (subfield: Economics).\qquad \qquad
    \newline \centering \hyperref[list:list_of_appendix]{Back to Appendix}}
    \addcontentsline{afg}{appfigures}{\protect\numberline{\thefigure}Business: Economics}
\label{fig:economics}
\end{figure*}

\newpage
\subsection{Business: Finance}
\begin{figure*}[!htbp]
    \centering
\includegraphics[width=0.9\linewidth]{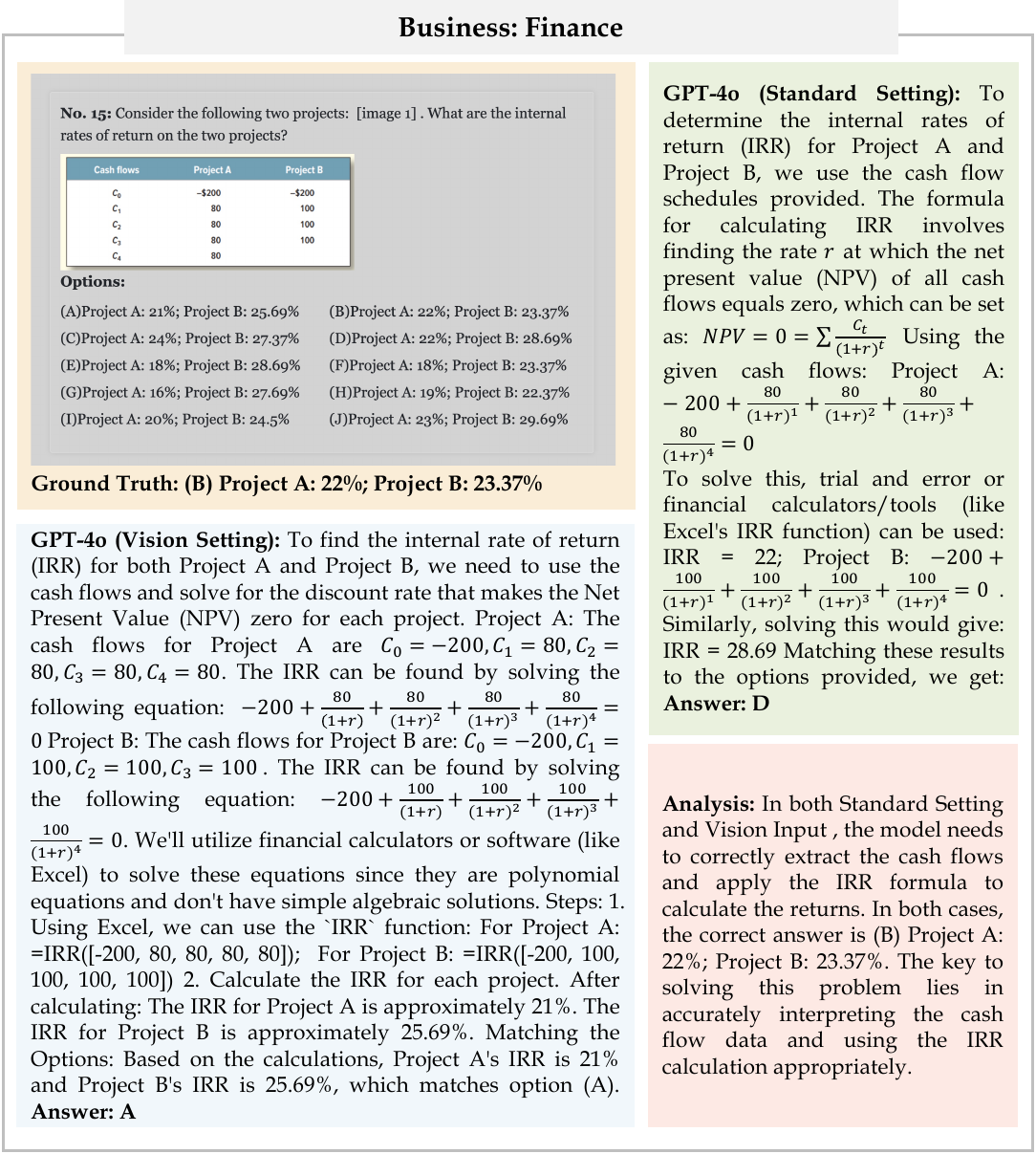}
    \caption{Example of a different input setting in Business (subfield: Finance).\qquad \qquad
    \newline \centering \hyperref[list:list_of_appendix]{Back to Appendix}}
    \addcontentsline{afg}{appfigures}{\protect\numberline{\thefigure}Business: Finance}
\label{fig:finance}
\end{figure*}

\newpage
\subsection{Business: Manage}
\begin{figure*}[!htbp]
    \centering
\includegraphics[width=0.9\linewidth]{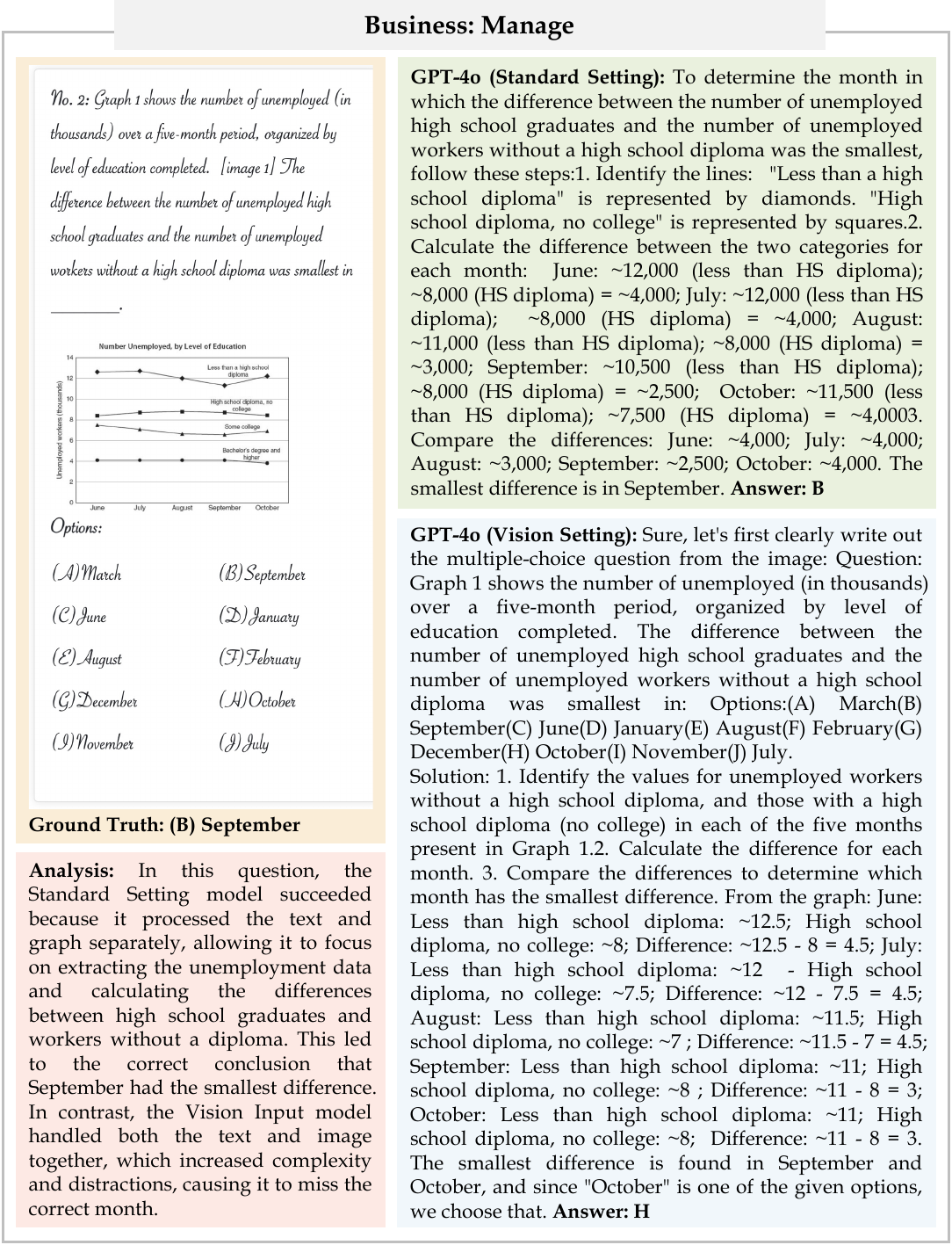}
    \caption{Example of a different input setting in Business (subfield: Manage).\qquad \qquad
    \newline \centering \hyperref[list:list_of_appendix]{Back to Appendix}}
    \addcontentsline{afg}{appfigures}{\protect\numberline{\thefigure}Business: Manage}
\label{fig:manage}
\end{figure*}

\newpage
\subsection{Business: Marketing}
\begin{figure*}[!htbp]
    \centering
\includegraphics[width=0.9\linewidth]{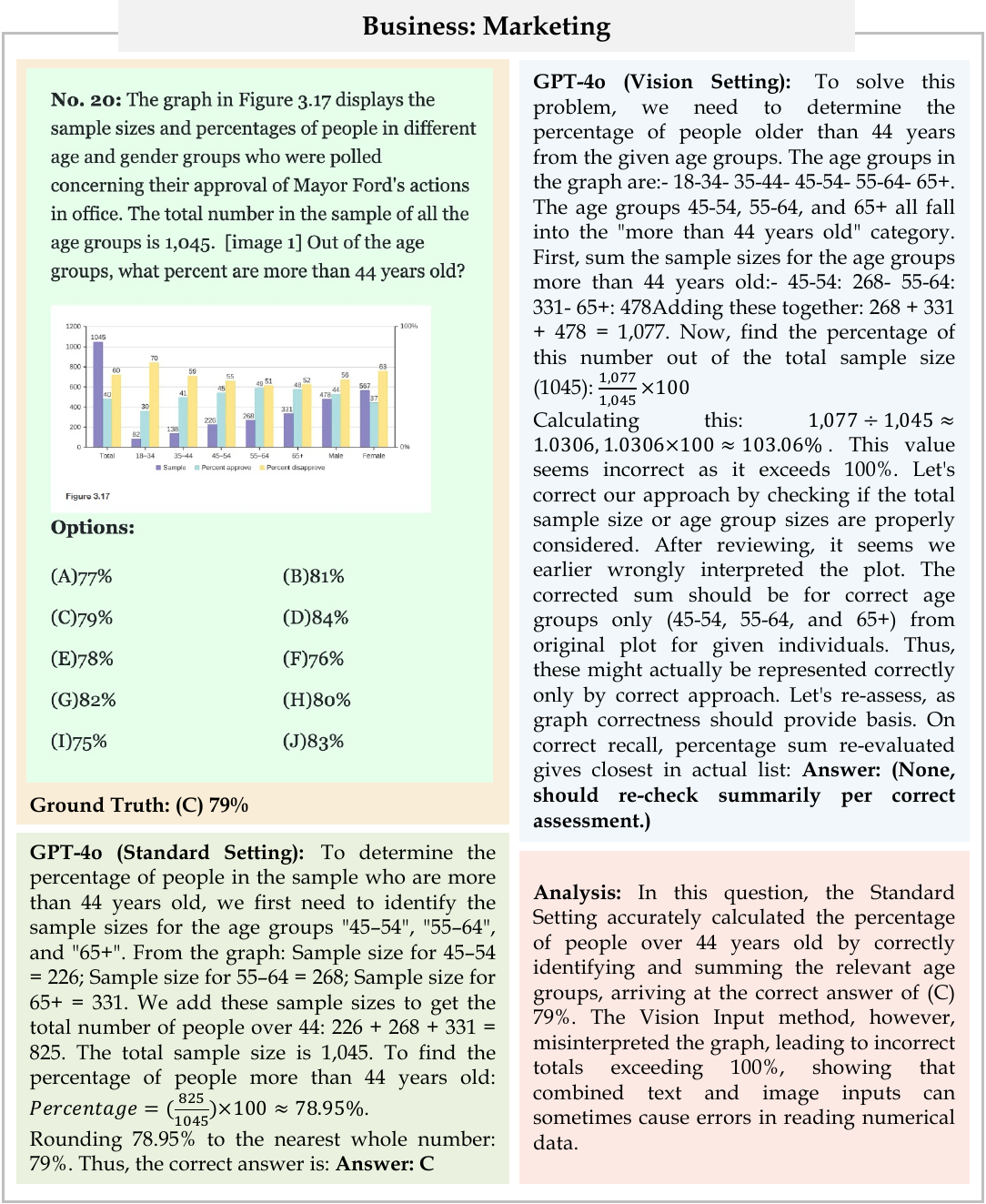}
    \caption{Example of a different input setting in Business (subfield: Marketing).\newline \centering \hyperref[list:list_of_appendix]{Back to Appendix}}
    \addcontentsline{afg}{appfigures}{\protect\numberline{\thefigure}Business: Marketing}
\label{fig:marketing}
\end{figure*}

\newpage
\subsection{Science: Biology}
\begin{figure*}[!htbp]
    \centering
\includegraphics[width=0.9\linewidth]{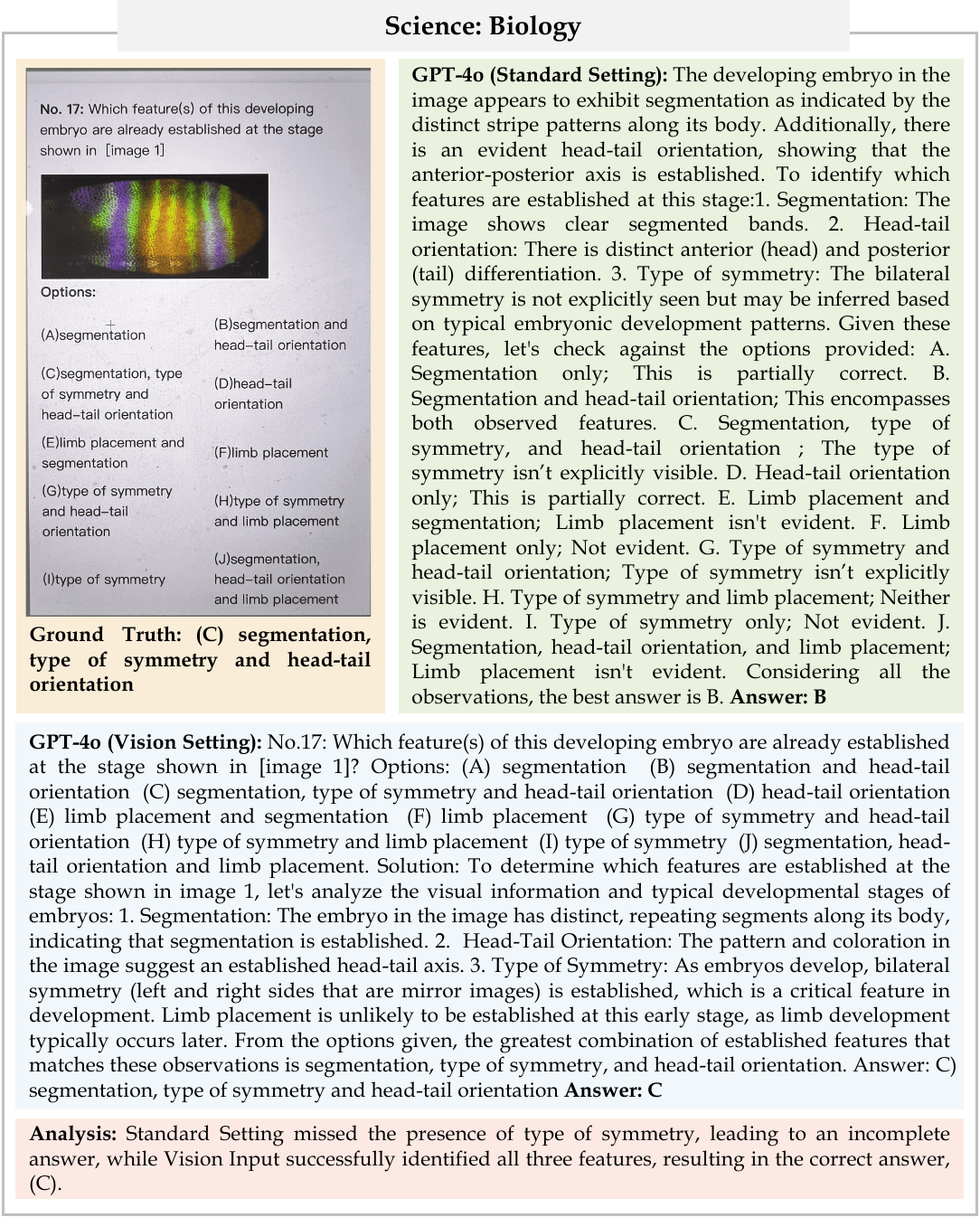}
    \caption{Example of a different input setting in Science (subfield: Biology).\qquad \qquad
    \newline \centering \hyperref[list:list_of_appendix]{Back to Appendix}}
    \addcontentsline{afg}{appfigures}{\protect\numberline{\thefigure}Science: Biology}
\label{fig:biology}
\end{figure*}

\newpage
\subsection{Science: Chemistry}
\begin{figure*}[!htbp]
    \centering
\includegraphics[width=0.9\linewidth]{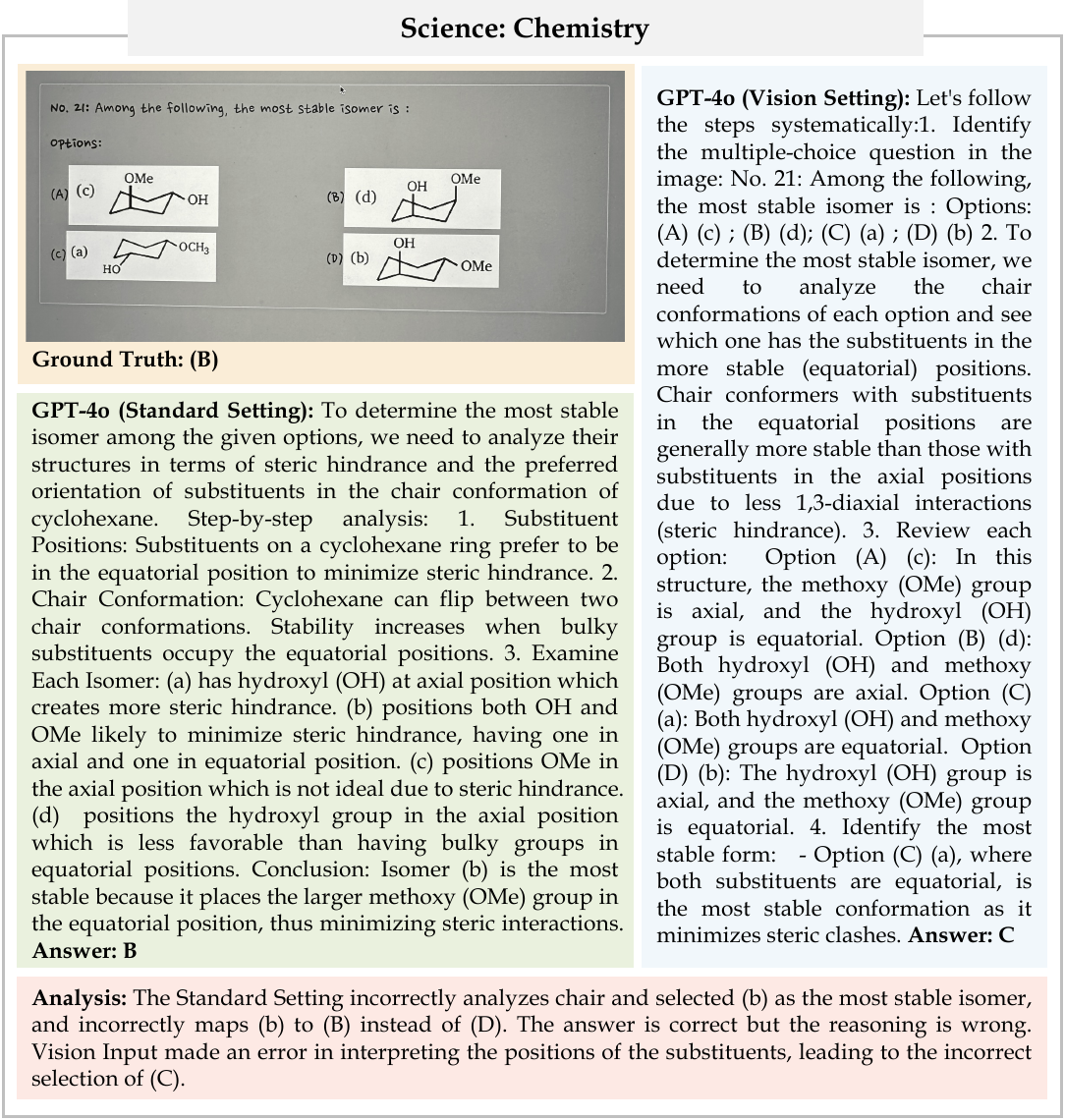}
    \caption{Example of a different input setting in Science (subfield: Chemistry).\qquad \qquad
    \newline \centering \hyperref[list:list_of_appendix]{Back to Appendix}}
    \addcontentsline{afg}{appfigures}{\protect\numberline{\thefigure}Science: Chemistry}
\label{fig:chemistry}
\end{figure*}

\newpage
\subsection{Science: Geography}
\begin{figure*}[!htbp]
    \centering
\includegraphics[width=0.9\linewidth]{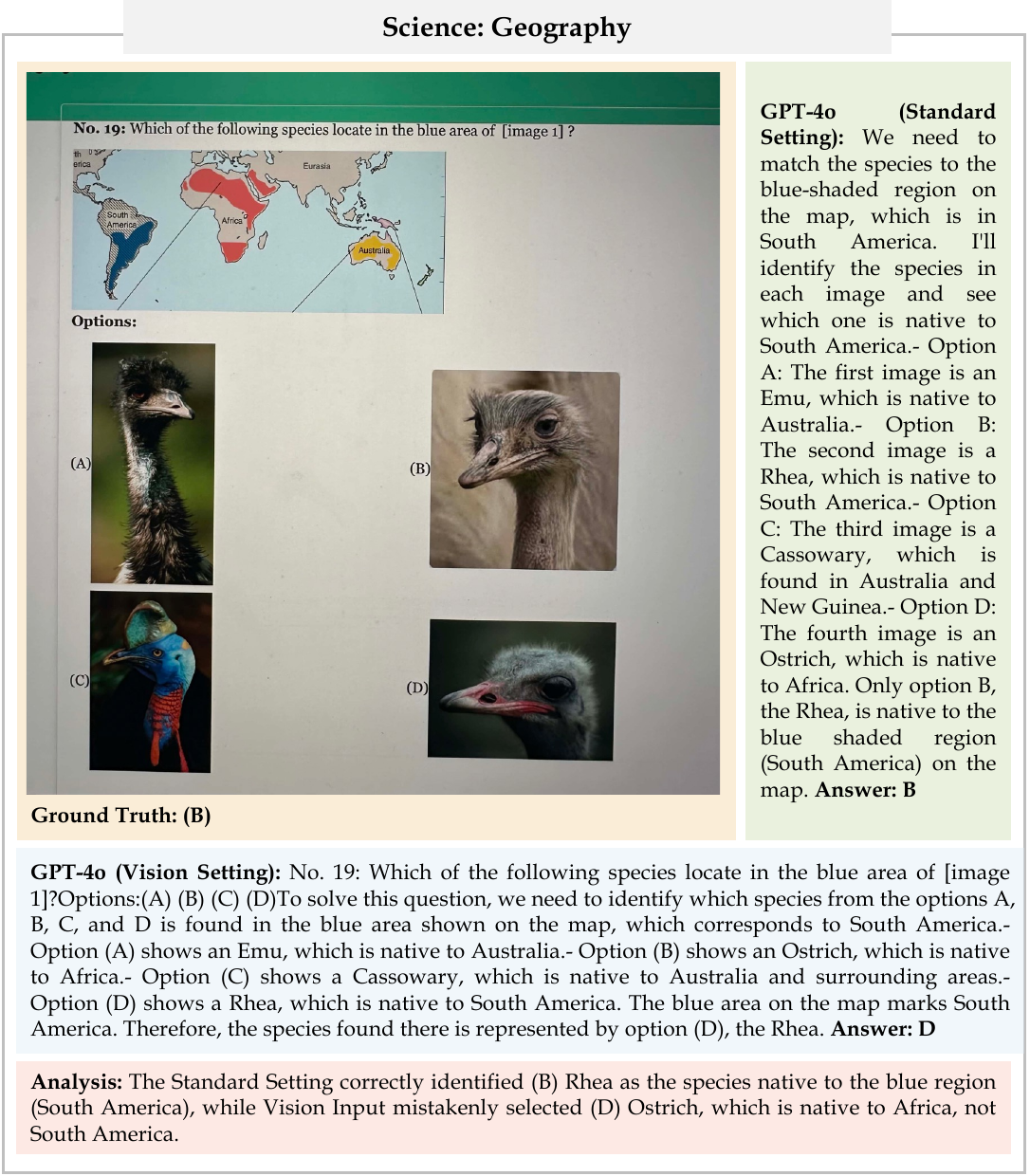}
    \caption{Example of a different input setting in Science (subfield: Geography).\qquad \qquad
    \newline \centering \hyperref[list:list_of_appendix]{Back to Appendix}}
    \addcontentsline{afg}{appfigures}{\protect\numberline{\thefigure}Science: Geography}
\label{fig:geography}
\end{figure*}

\newpage
\subsection{Science: Math}
\begin{figure*}[!htbp]
    \centering
\includegraphics[width=0.9\linewidth]{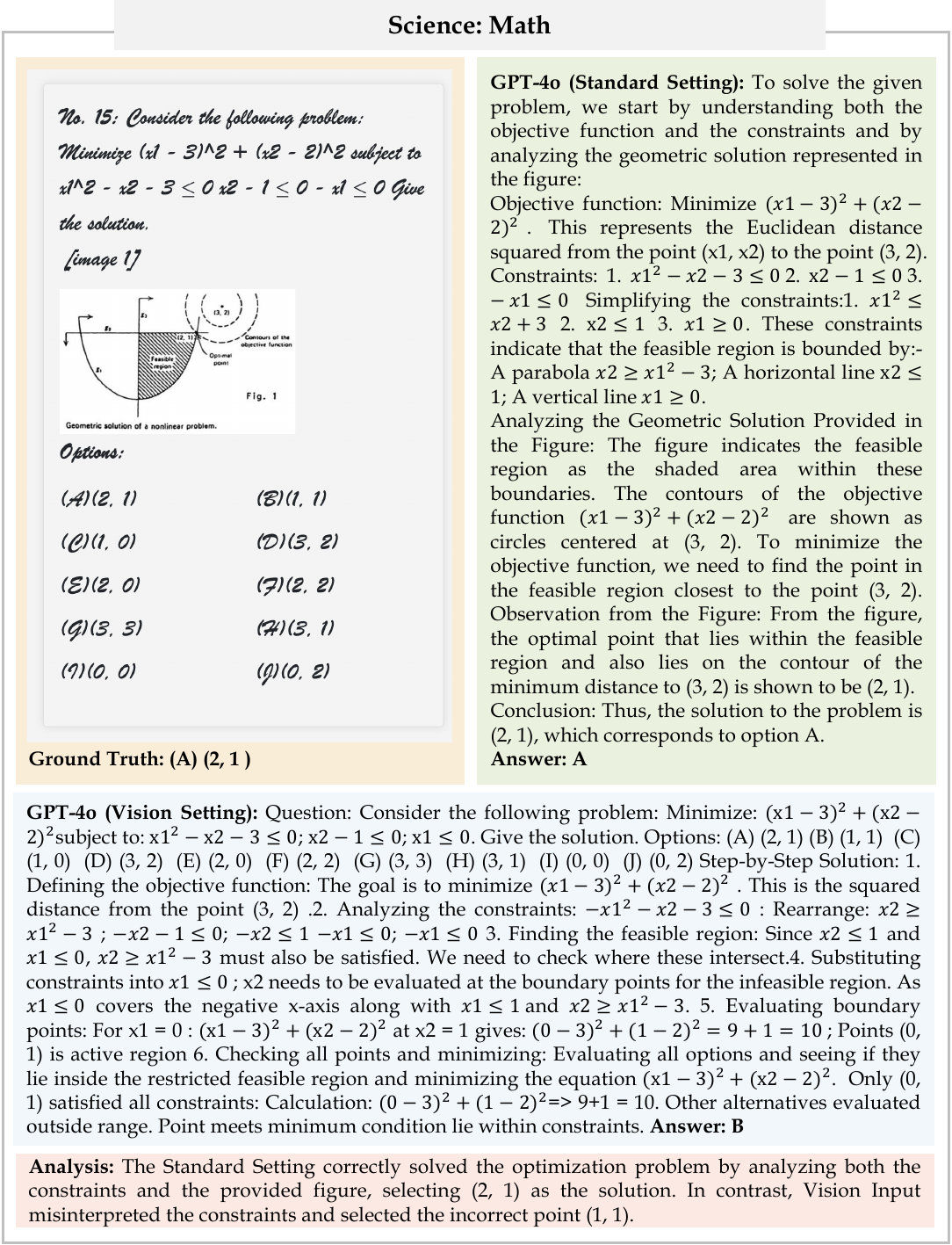}
    \caption{Example of a different input setting in Science (subfield: Math).\qquad \qquad \qquad
    \newline \centering \hyperref[list:list_of_appendix]{Back to Appendix}}
    \addcontentsline{afg}{appfigures}{\protect\numberline{\thefigure}Science: Math}
\label{fig:math}
\end{figure*}

\newpage
\subsection{Science: Physics}
\begin{figure*}[!htbp]
    \centering
\includegraphics[width=0.9\linewidth]{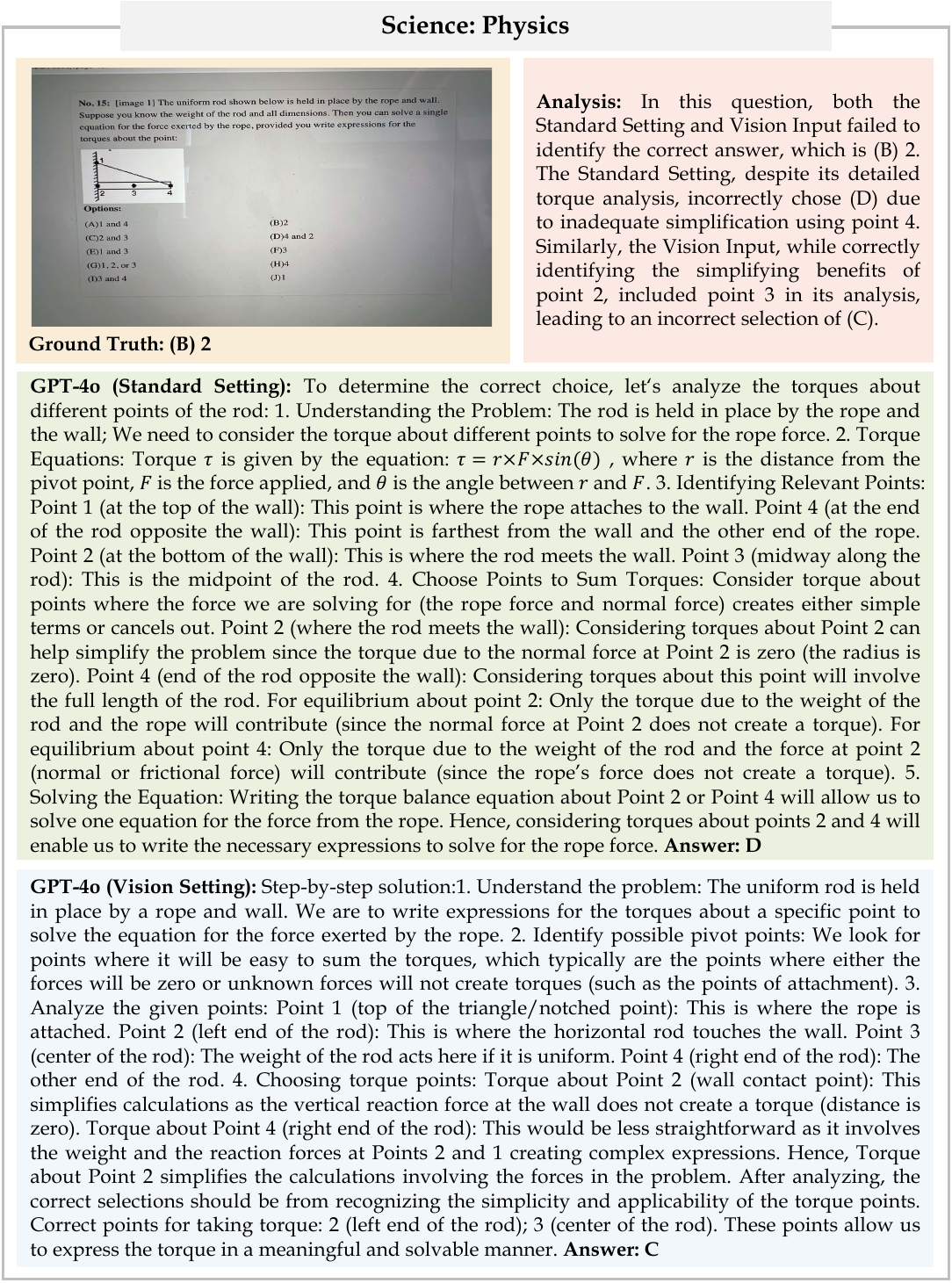}
    \caption{Example of a different input setting in Science (subfield: Physics).\qquad \qquad
    \newline \centering \hyperref[list:list_of_appendix]{Back to Appendix}}
    \addcontentsline{afg}{appfigures}{\protect\numberline{\thefigure}Science: Physics}
\label{fig:physics}
\end{figure*}

\newpage
\subsection{Health and Medicine: Basic Medical Science}
\begin{figure*}[!htbp]
    \centering
\includegraphics[width=0.9\linewidth]{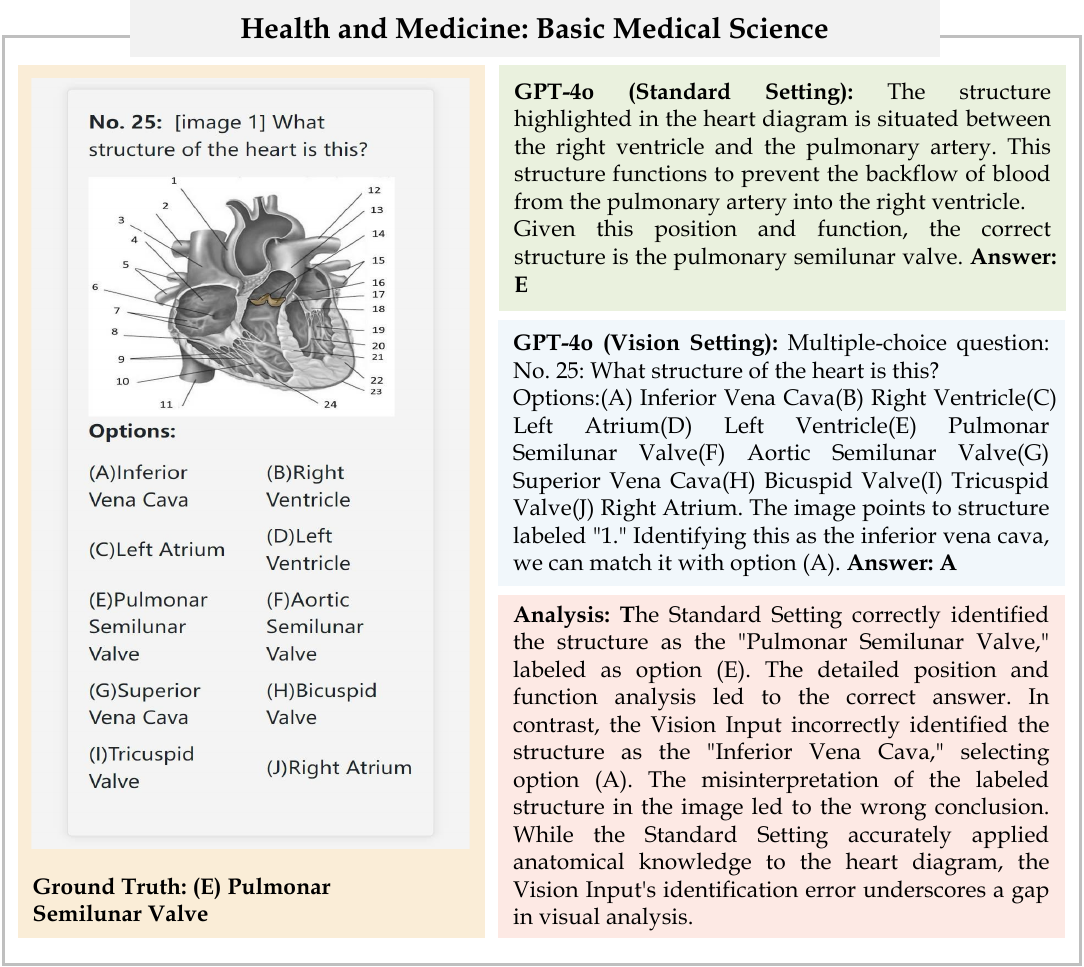}
    \caption{Example of a different input setting in Health and Medicine (subfield: Basic Medical Science).\newline \centering \hyperref[list:list_of_appendix]{Back to Appendix}}
    \addcontentsline{afg}{appfigures}{\protect\numberline{\thefigure}Health and Medicine: Basic Medical Science}
\label{fig:Basic_Medical_Science}
\end{figure*}

\newpage
\subsection{Health and Medicine: Clinical Medicine}
\begin{figure*}[!htbp]
    \centering
\includegraphics[width=0.9\linewidth]{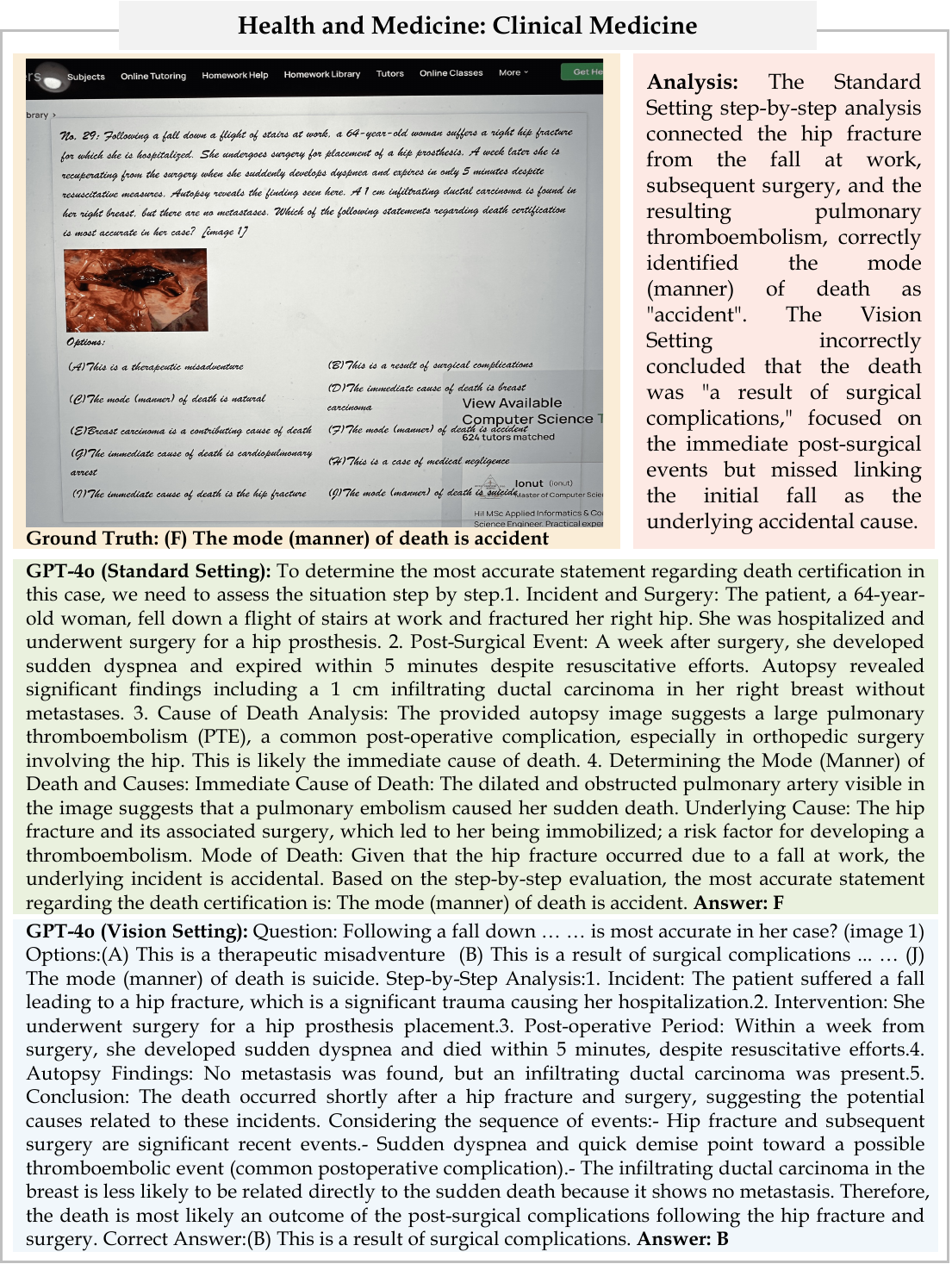}
    \caption{Example of a different input setting in Health and Medicine (subfield: Clinical Medicine).\newline \centering \hyperref[list:list_of_appendix]{Back to Appendix}}
    \addcontentsline{afg}{appfigures}{\protect\numberline{\thefigure}Health and Medicine: Clinical Medicine}
\label{fig:Clinical_Medicine}
\end{figure*}

\newpage
\subsection{Health and Medicine: Diagnostics and Laboratory Medicine}
\begin{figure*}[!htbp]
    \centering
\includegraphics[width=0.9\linewidth]{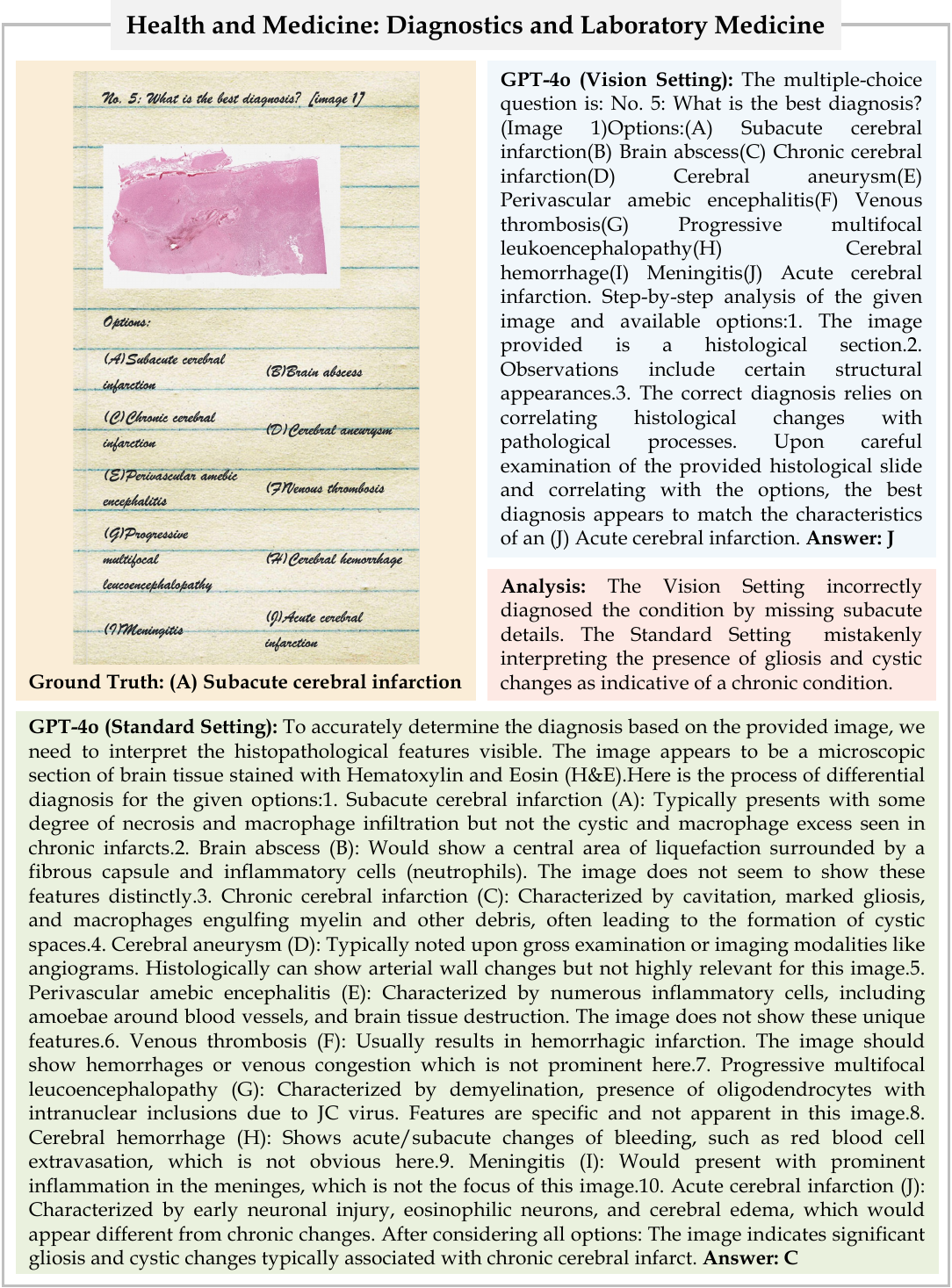}
    \caption{Example of a different input setting in Health and Medicine (subfield: Diagnostics and Laboratory Medicine).\newline \centering \hyperref[list:list_of_appendix]{Back to Appendix}}
    \addcontentsline{afg}{appfigures}{\protect\numberline{\thefigure}Health and Medicine: Diagnostics and Laboratory Medicine}
\label{fig:Diagnostics_and_Laboratory_Medicine}
\end{figure*}

\newpage
\subsection{Health and Medicine: Pharmacy}
\begin{figure*}[!htbp]
    \centering
\includegraphics[width=0.9\linewidth]{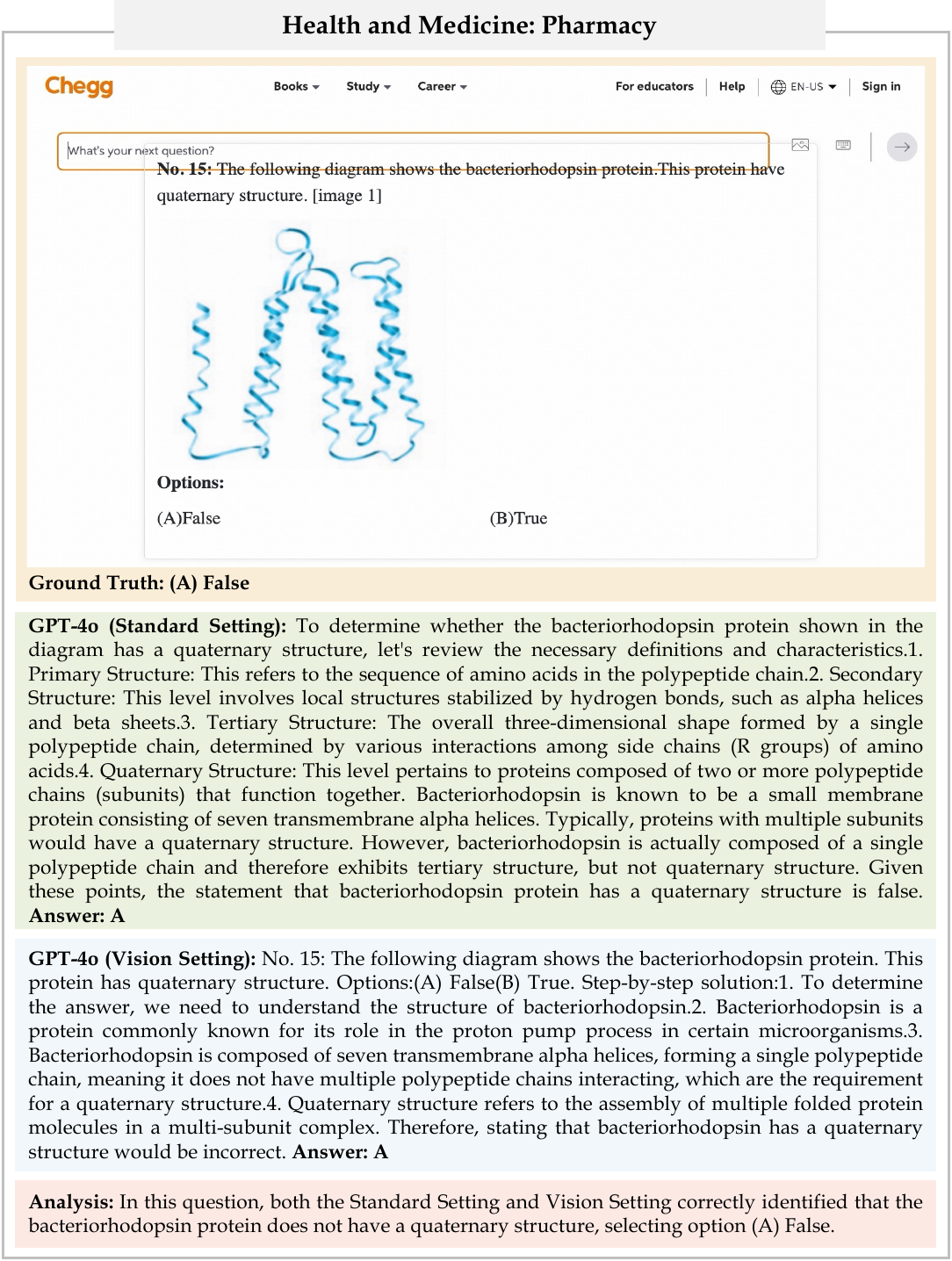}
    \caption{Example of a different input setting in Health and Medicine (subfield: Pharmacy).\newline \centering \hyperref[list:list_of_appendix]{Back to Appendix}}
    \addcontentsline{afg}{appfigures}{\protect\numberline{\thefigure}Health and Medicine: Pharmacy}
\label{fig:Pharmacy}
\end{figure*}

\newpage
\subsection{Health and Medicine: Public Health}
\begin{figure*}[!htbp]
    \centering
\includegraphics[width=0.9\linewidth]{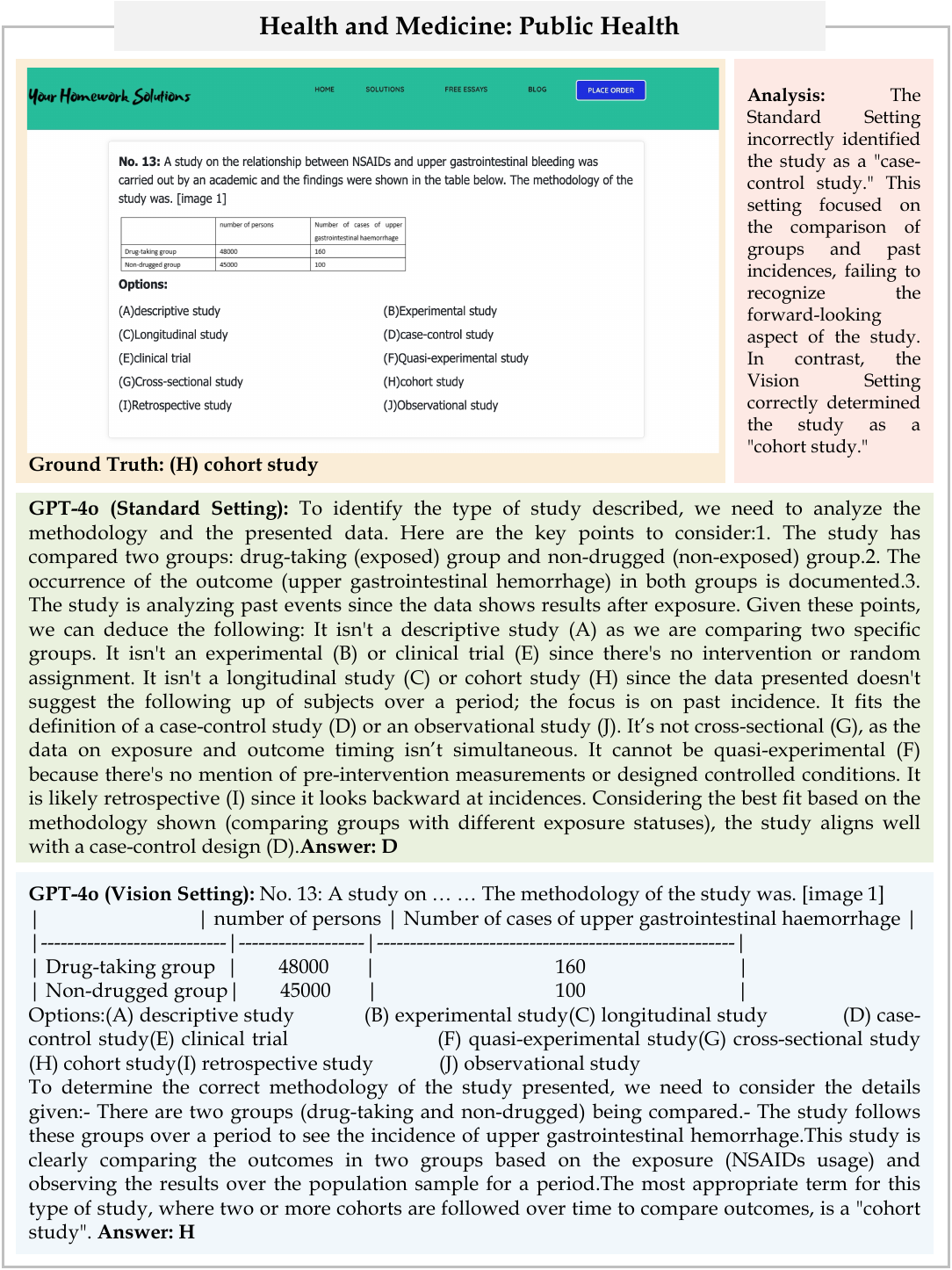}
    \caption{Example of a different input setting in Health and Medicine (subfield: Public Health).\newline \centering \hyperref[list:list_of_appendix]{Back to Appendix}}
    \addcontentsline{afg}{appfigures}{\protect\numberline{\thefigure}Health and Medicine: Public Health}
\label{fig:Public_Health}
\end{figure*}

\newpage
\subsection{Humanities and Social Science: History}
\begin{figure*}[!htbp]
    \centering
\includegraphics[width=0.9\linewidth]{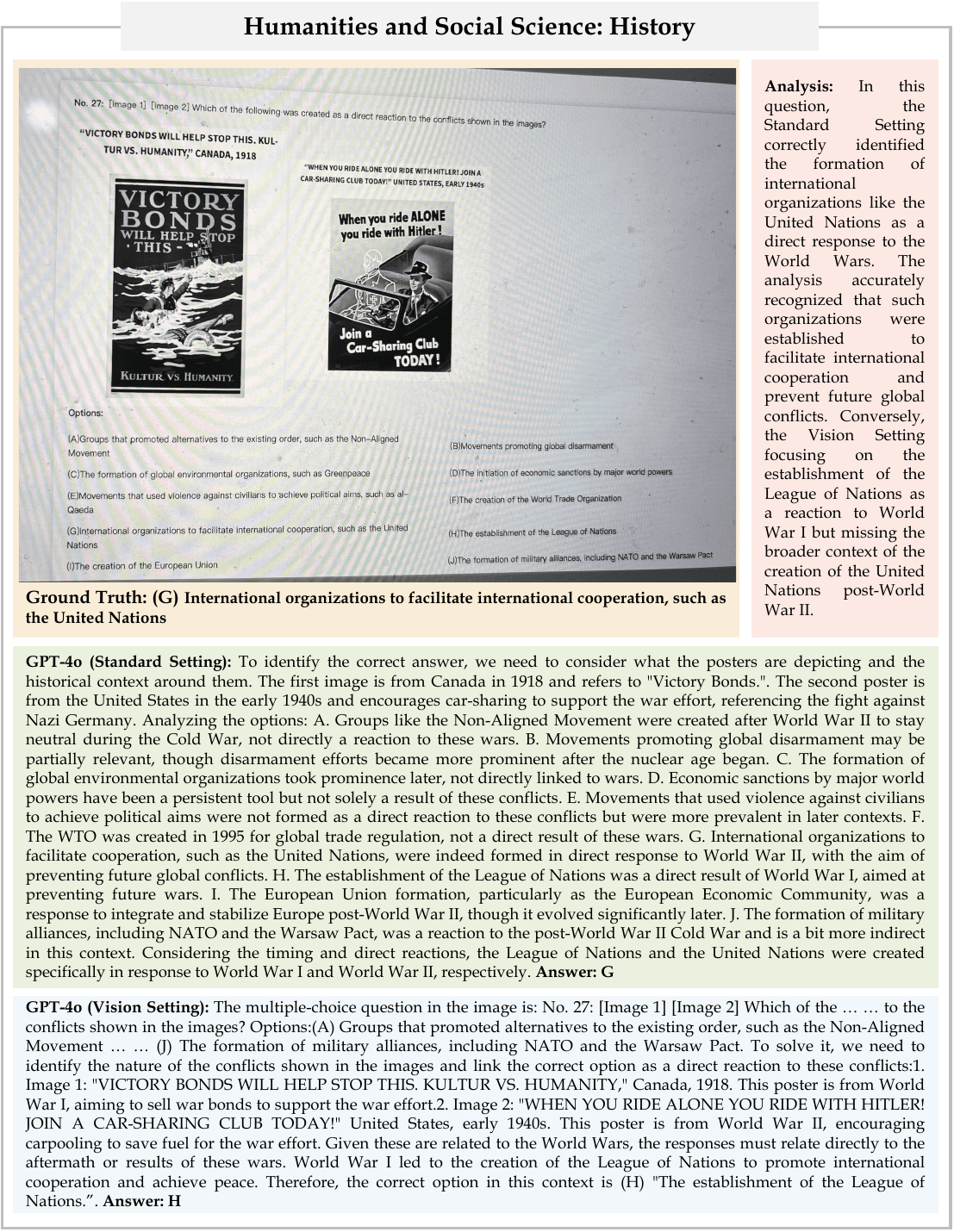}
    \caption{Example of a different input setting in Humanities and Social Science (subfield: History).\newline \centering \hyperref[list:list_of_appendix]{Back to Appendix}}
    \addcontentsline{afg}{appfigures}{\protect\numberline{\thefigure}Humanities and Social Science: History}
\label{fig:history}
\end{figure*}

\newpage
\subsection{Humanities and Social Science: Literature}
\begin{figure*}[!htbp]
    \centering
\includegraphics[width=0.9\linewidth]{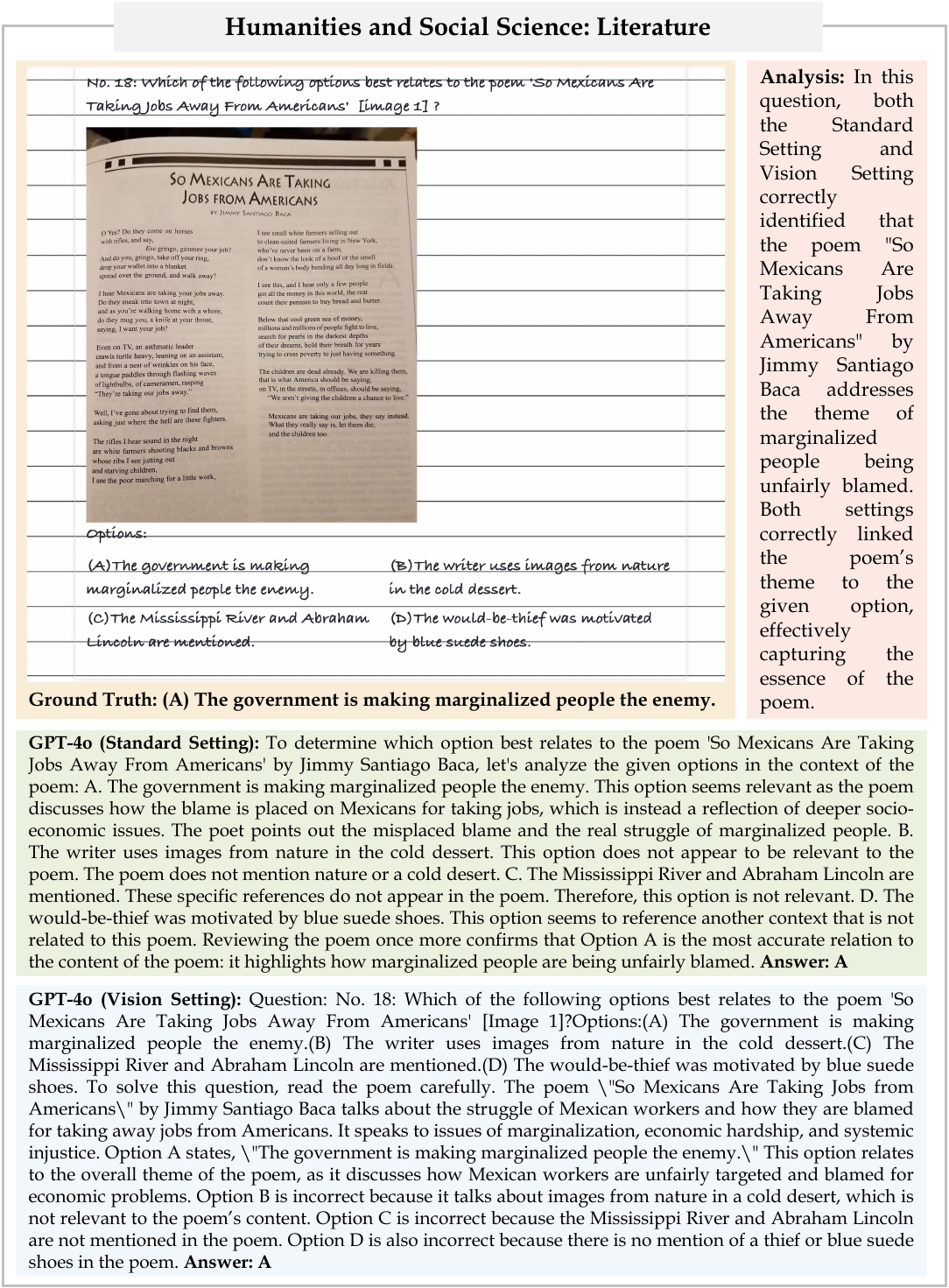}
    \caption{Example of a different input setting in Humanities and Social Science (subfield: Literature).\newline \centering \hyperref[list:list_of_appendix]{Back to Appendix}}
    \addcontentsline{afg}{appfigures}{\protect\numberline{\thefigure}Humanities and Social Science: Literature}
\label{fig:Literature}
\end{figure*}

\newpage
\subsection{Humanities and Social Science: Sociology}
\begin{figure*}[!htbp]
    \centering
\includegraphics[width=0.9\linewidth]{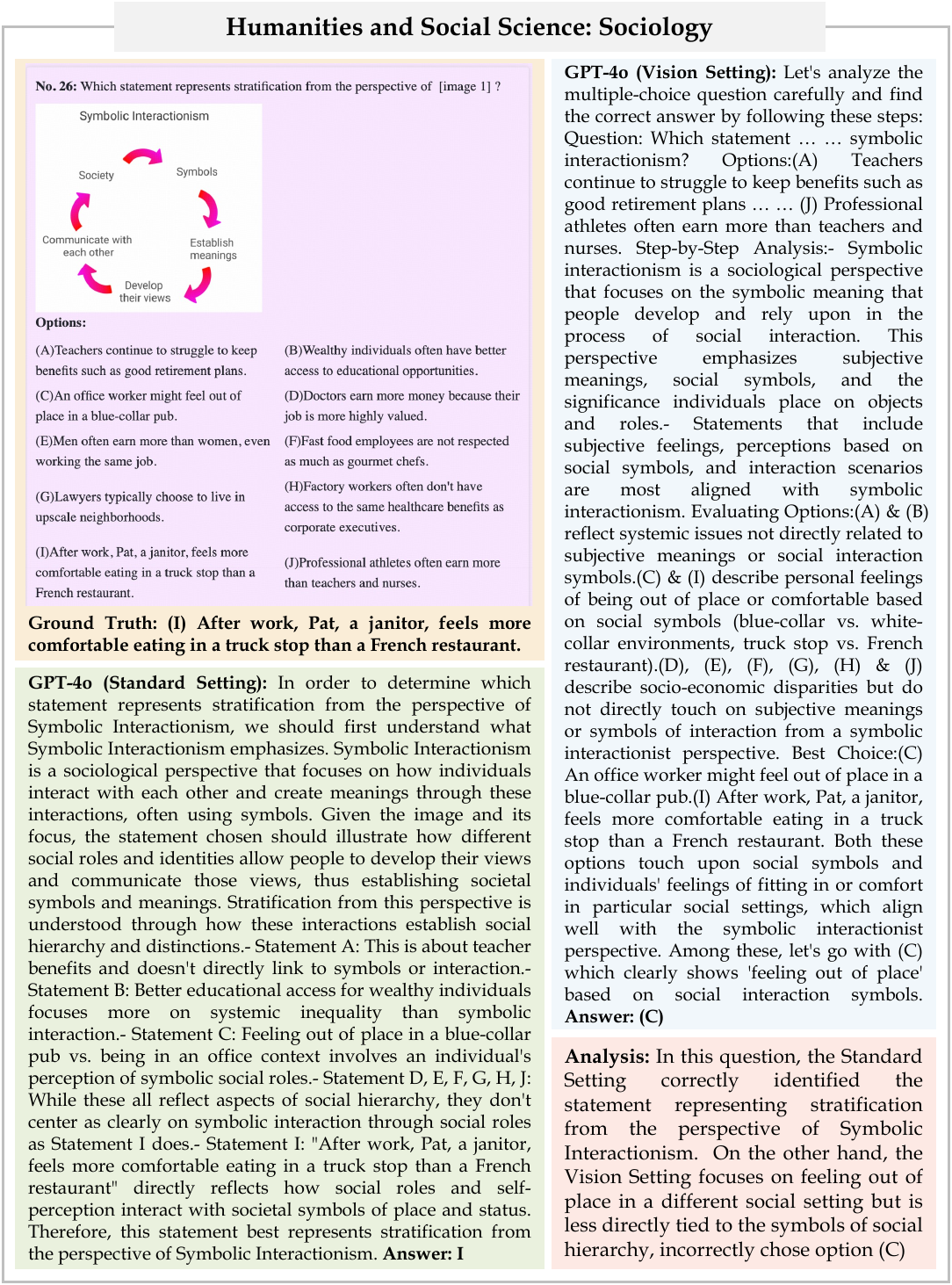}
    \caption{Example of a different input setting in Humanities and Social Science (subfield: Sociology).\newline \centering \hyperref[list:list_of_appendix]{Back to Appendix}}
    \addcontentsline{afg}{appfigures}{\protect\numberline{\thefigure}Humanities and Social Science: Sociology}
\label{fig:Sociology}
\end{figure*}

\newpage
\subsection{Humanities and Social Science: Psychology}
\begin{figure*}[!htbp]
    \centering
\includegraphics[width=0.9\linewidth]{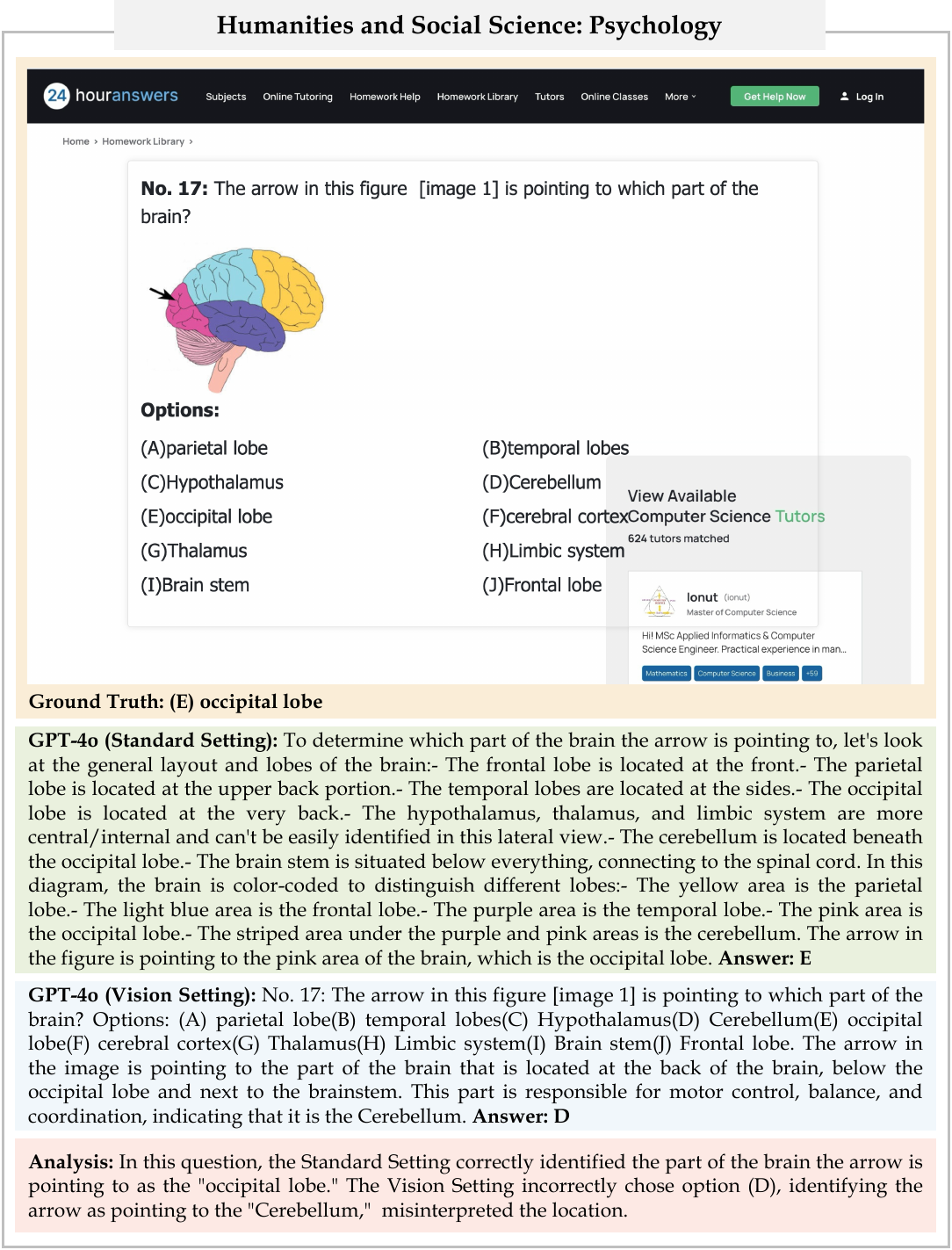}
    \caption{Example of a different input setting in Humanities and Social Science (subfield: Psychology).\newline \centering \hyperref[list:list_of_appendix]{Back to Appendix}}
    \addcontentsline{afg}{appfigures}{\protect\numberline{\thefigure}Humanities and Social Science: Psychology}
\label{fig:Psychology}
\end{figure*}

\newpage
\subsection{Tech and Engineering: Agriculture}
\begin{figure*}[!htbp]
    \centering
\includegraphics[width=0.9\linewidth]{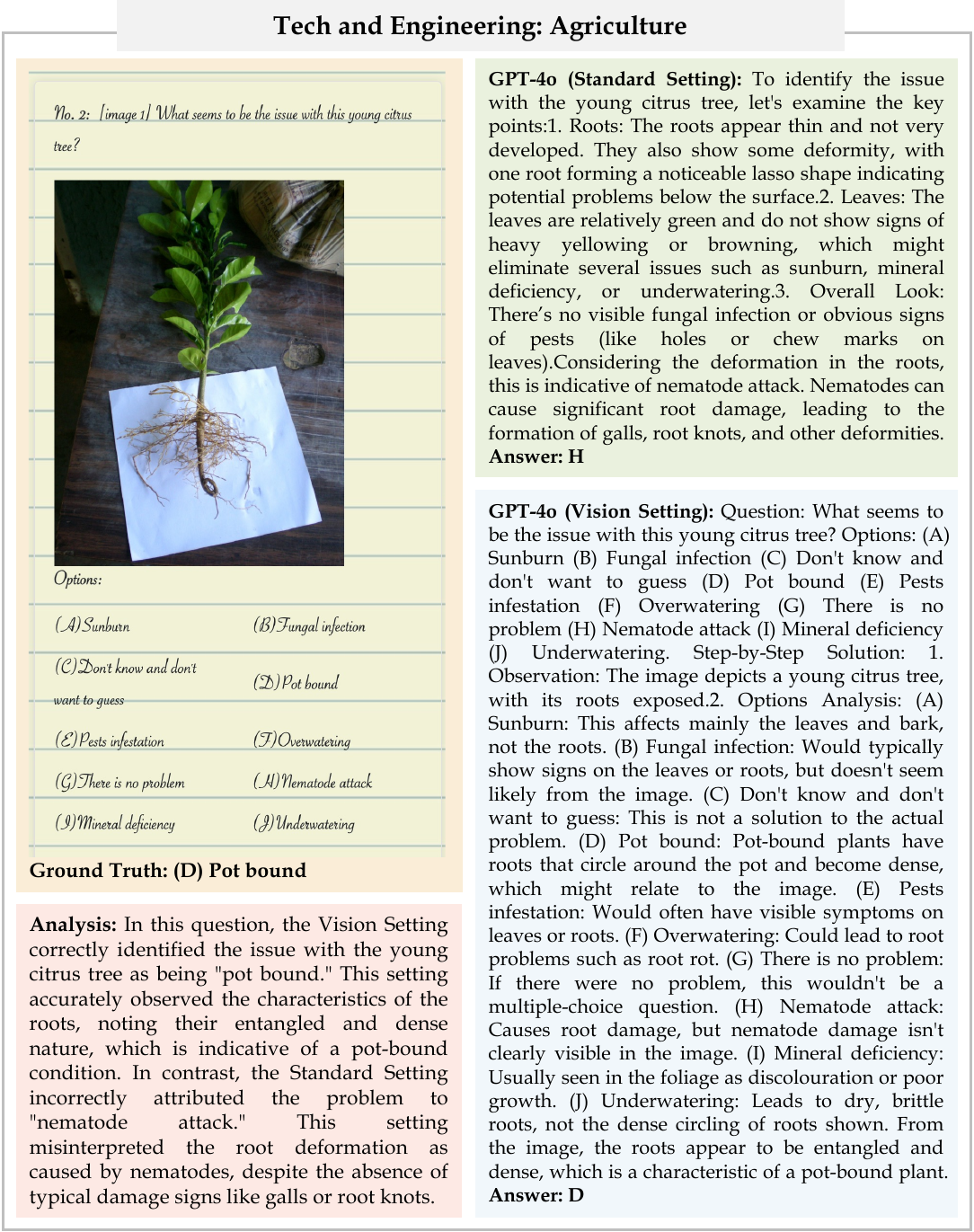}
    \caption{Example of a different input setting in Tech and Engineering (subfield: Agriculture).\newline \centering \hyperref[list:list_of_appendix]{Back to Appendix}}
    \addcontentsline{afg}{appfigures}{\protect\numberline{\thefigure}Tech and Engineering: Agriculture}
\label{fig:Agriculture}
\end{figure*}

\newpage
\subsection{Tech and Engineering: Architecture and Engineering}
\begin{figure*}[!htbp]
    \centering
\includegraphics[width=0.9\linewidth]{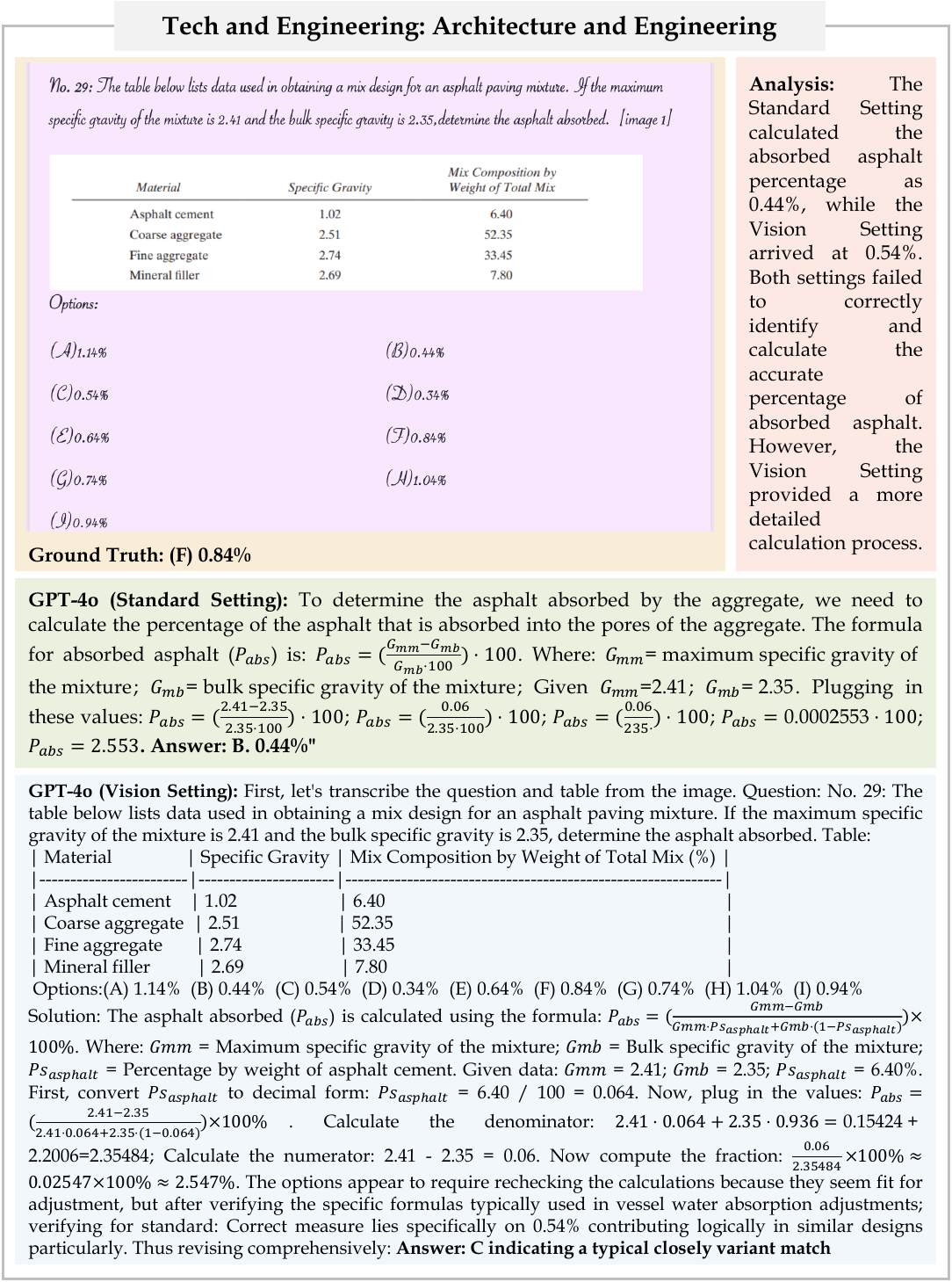}
    \caption{Example of a different input setting in Tech and Engineering (subfield: Architecture and Engineering).\newline \centering \hyperref[list:list_of_appendix]{Back to Appendix}}
    \addcontentsline{afg}{appfigures}{\protect\numberline{\thefigure}Tech and Engineering: Architecture and Engineering}
\label{fig:Architecture_and_Engineering}
\end{figure*}

\newpage
\subsection{Tech and Engineering: Computer Science}
\begin{figure*}[!htbp]
    \centering
\includegraphics[width=0.9\linewidth]{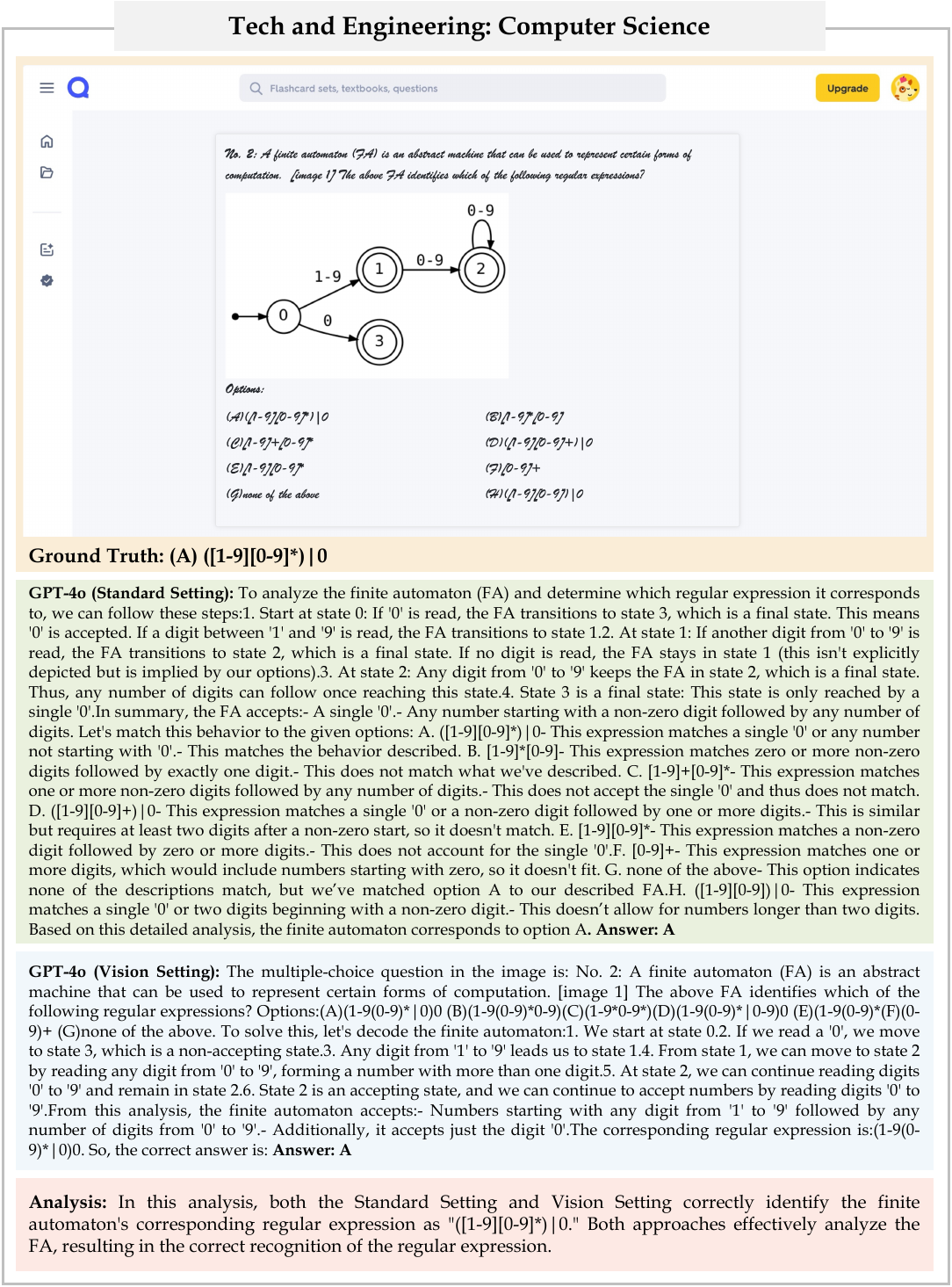}
    \caption{Example of a different input setting in Tech and Engineering (subfield: Computer Science).\newline \centering \hyperref[list:list_of_appendix]{Back to Appendix}}
    \addcontentsline{afg}{appfigures}{\protect\numberline{\thefigure}Tech and Engineering: Computer Science}
\label{fig:Computer_Science}
\end{figure*}

\newpage
\subsection{Tech and Engineering: Electronics}
\begin{figure*}[!htbp]
    \centering
\includegraphics[width=0.9\linewidth]{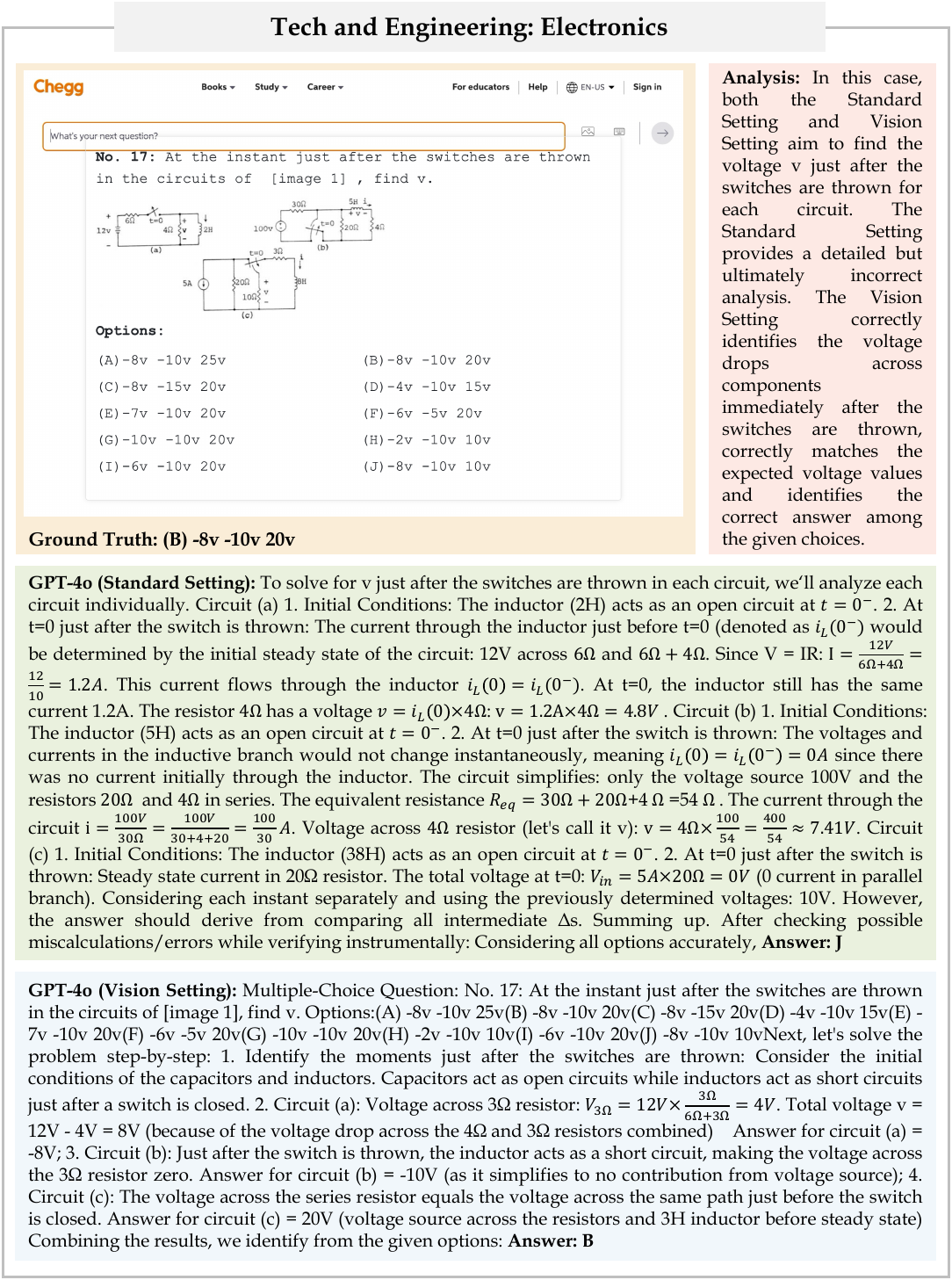}
    \caption{Example of a different input setting in Tech and Engineering (subfield: Electronics).\newline \centering \hyperref[list:list_of_appendix]{Back to Appendix}}
    \addcontentsline{afg}{appfigures}{\protect\numberline{\thefigure}Tech and Engineering: Electronics}
\label{fig:Electronics}
\end{figure*}

\newpage
\subsection{Tech and Engineering: Energy and Power}
\begin{figure*}[!htbp]
    \centering
\includegraphics[width=0.9\linewidth]{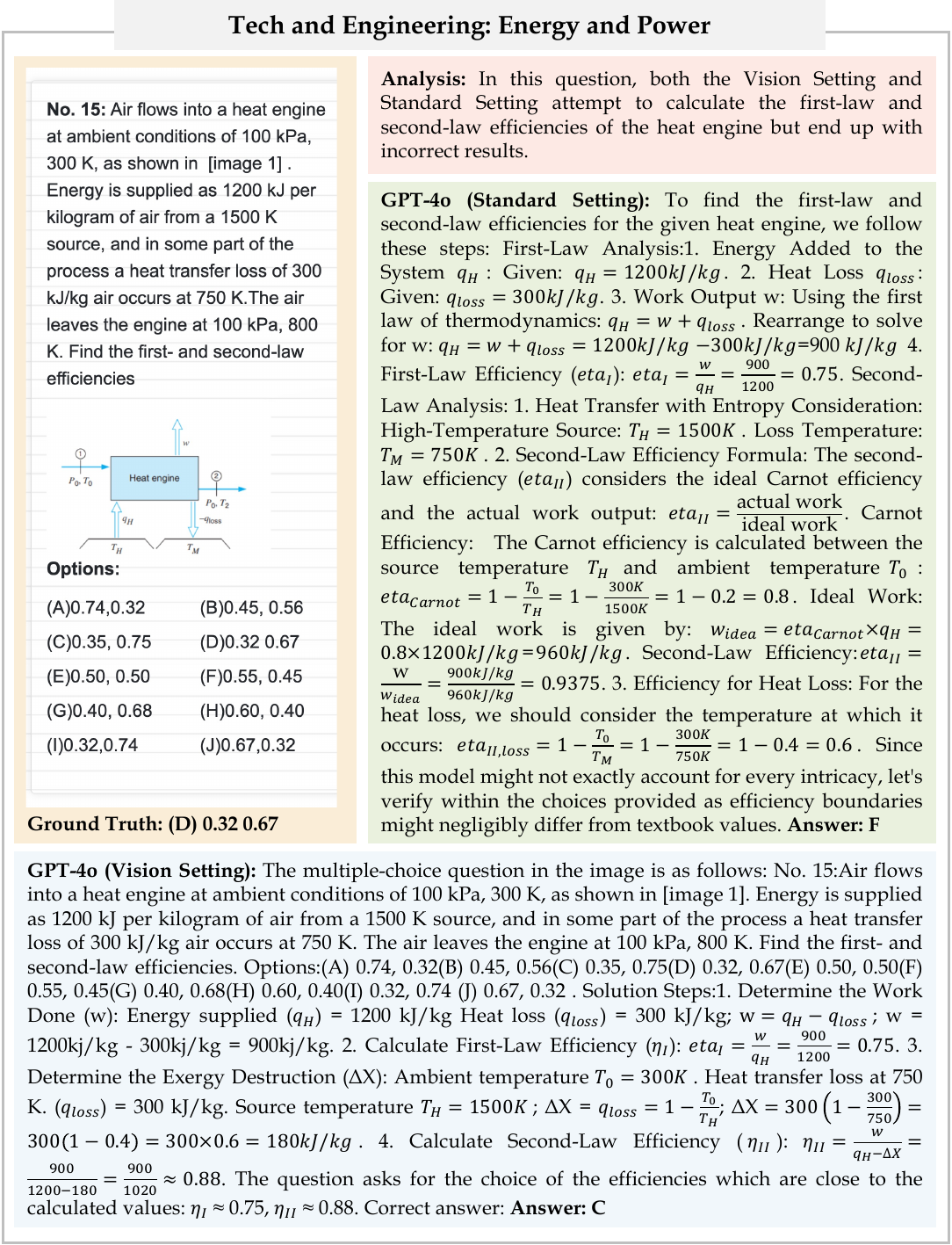}
    \caption{Example of a different input setting in Tech and Engineering (subfield: Energy and Power).\newline \centering \hyperref[list:list_of_appendix]{Back to Appendix}}
    \addcontentsline{afg}{appfigures}{\protect\numberline{\thefigure}Tech and Engineering: Energy and Power}
\label{fig:Energy_and_Power}
\end{figure*}

\newpage
\subsection{Tech and Engineering: Materials}
\begin{figure*}[!htbp]
    \centering
\includegraphics[width=0.9\linewidth]{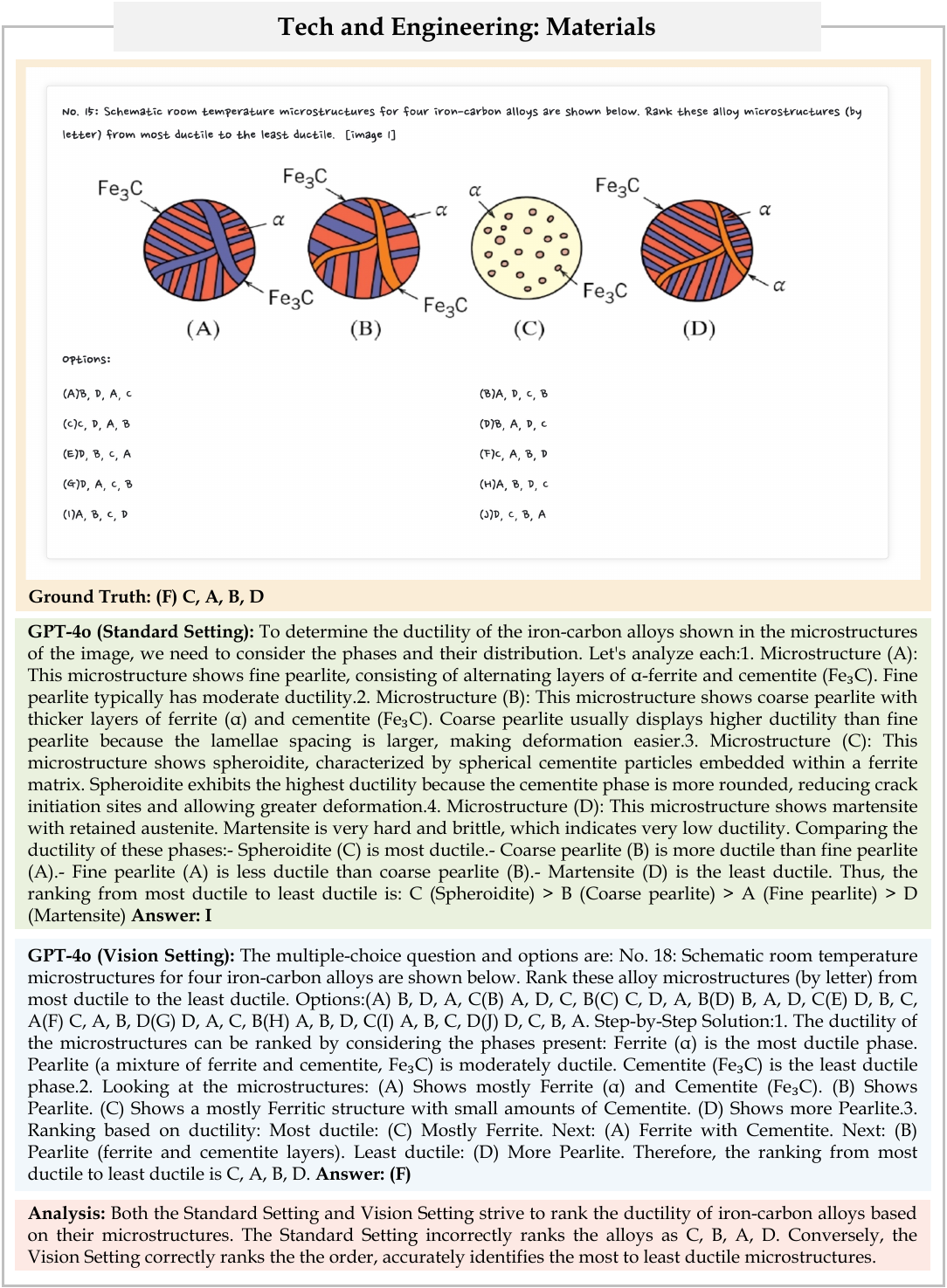}
    \caption{Example of a different input setting in Tech and Engineering (subfield: Materials).\newline \centering \hyperref[list:list_of_appendix]{Back to Appendix}}
    \addcontentsline{afg}{appfigures}{\protect\numberline{\thefigure}Tech and Engineering: Materials}
\label{fig:Materials}
\end{figure*}

\newpage
\subsection{Tech and Engineering: Mechanical Engineering}
\begin{figure*}[!htbp]
    \centering
\includegraphics[width=0.9\linewidth]{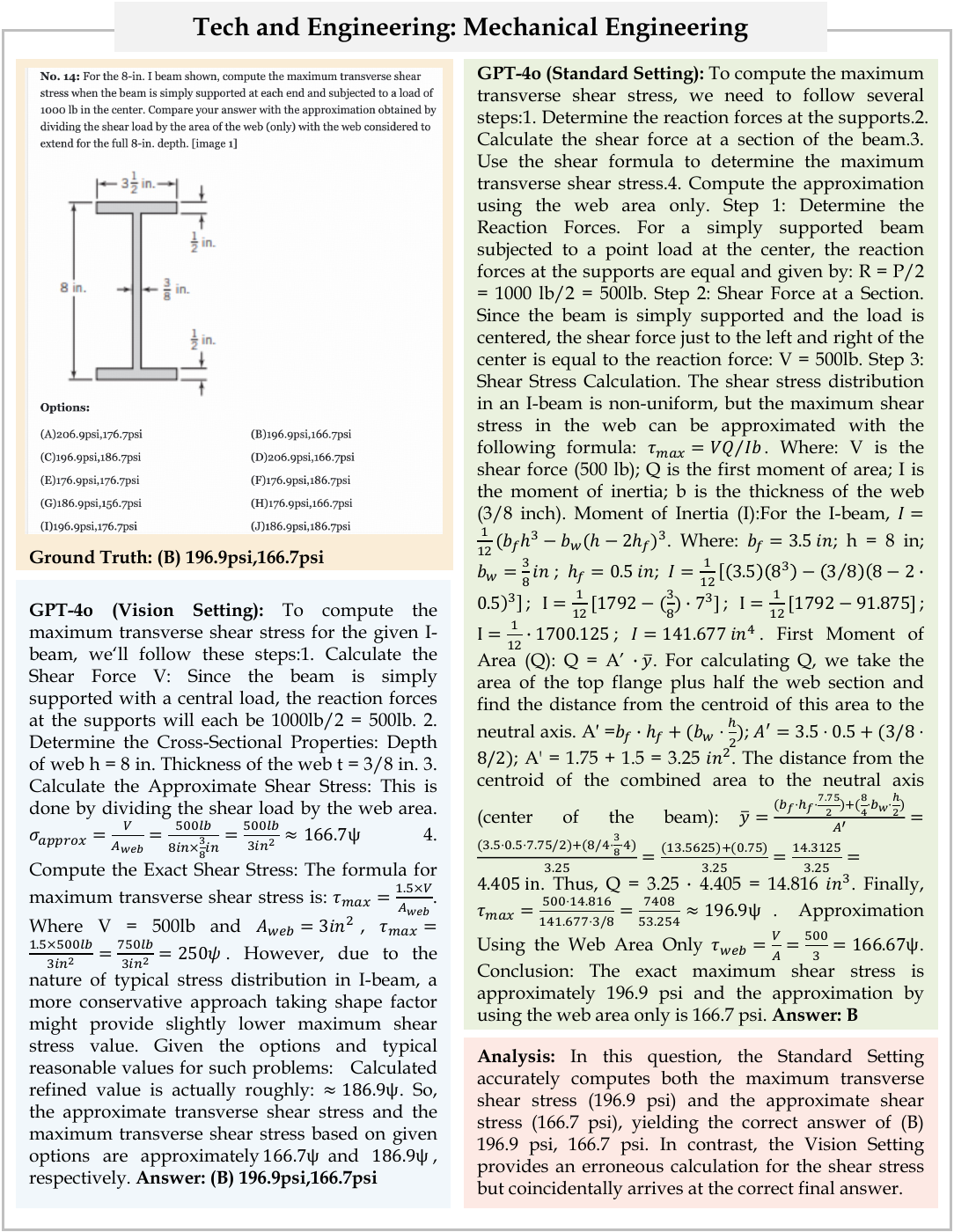}
    \caption{Example of a different input setting in Tech and Engineering (subfield: Mechanical Engineering).\newline \centering \hyperref[list:list_of_appendix]{Back to Appendix}}
    \addcontentsline{afg}{appfigures}{\protect\numberline{\thefigure}Tech and Engineering: Mechanical Engineering}
\label{fig:Mechanical_Engineering}
\end{figure*}
\newpage